\documentclass[runningheads]{llncs}
\usepackage{graphicx}

\usepackage{tikz}
\usepackage{comment}
\usepackage{amsmath,amssymb}
\usepackage{color}
\usepackage{svg}
\usepackage{booktabs}
\usepackage{lmodern}
\usepackage{subcaption}
\usepackage[accsupp]{axessibility}

\usepackage[width=122mm,left=12mm,paperwidth=146mm,height=193mm,top=12mm,paperheight=217mm]{geometry}
\usepackage{microtype}

\begin{document}
\pagestyle{headings}
\mainmatter
\def\ECCVSubNumber{4426}

\title{Structure-aware Editable Morphable Model for 3D Facial Detail Animation and Manipulation}

\titlerunning{Structure-aware Editable Morphable Model}
\author{Jingwang Ling\inst{1} \and
Zhibo Wang\inst{1} \and
Ming Lu\inst{2} \and
Quan Wang\inst{3} \and
Chen Qian\inst{3} \and
Feng Xu\inst{1}}
\authorrunning{J. Ling et al.}
\institute{BNRist and school of software, Tsinghua University \and
Intel Labs China \and
Sensetime Research, China}
\maketitle

\begin{abstract}
Morphable models are essential for the statistical modeling of 3D faces. Previous works on morphable models mostly focus on large-scale facial geometry but ignore facial details. This paper augments morphable models in representing facial details by learning a Structure-aware Editable Morphable Model (SEMM). SEMM introduces a detail structure representation based on the distance field of wrinkle lines, jointly modeled with detail displacements to establish better correspondences and enable intuitive manipulation of wrinkle structure. Besides, SEMM introduces two transformation modules to translate expression blendshape weights and age values into changes in latent space, allowing effective semantic detail editing while maintaining identity. Extensive experiments demonstrate that the proposed model compactly represents facial details, outperforms previous methods in expression animation qualitatively and quantitatively, and achieves effective age editing and wrinkle line editing of facial details. Code and model are available at \url{https://github.com/gerwang/facial-detail-manipulation}.
\end{abstract}

\section{Introduction}
\label{sec:intro}

\begin{figure}[t]
  \centering
  \begin{minipage}[t]{.492\textwidth}
    \centering
    \includeinkscape[width=.98\linewidth]{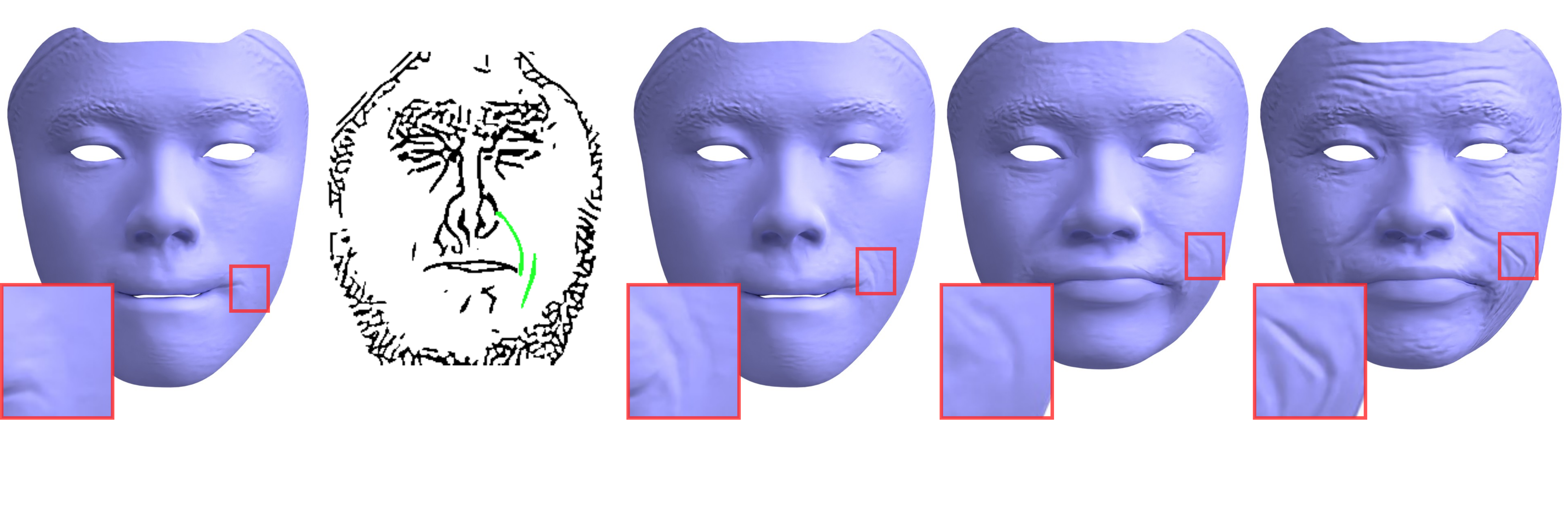}
  \end{minipage}
  \begin{minipage}[t]{.492\textwidth}
    \centering
    \includeinkscape[width=.98\linewidth]{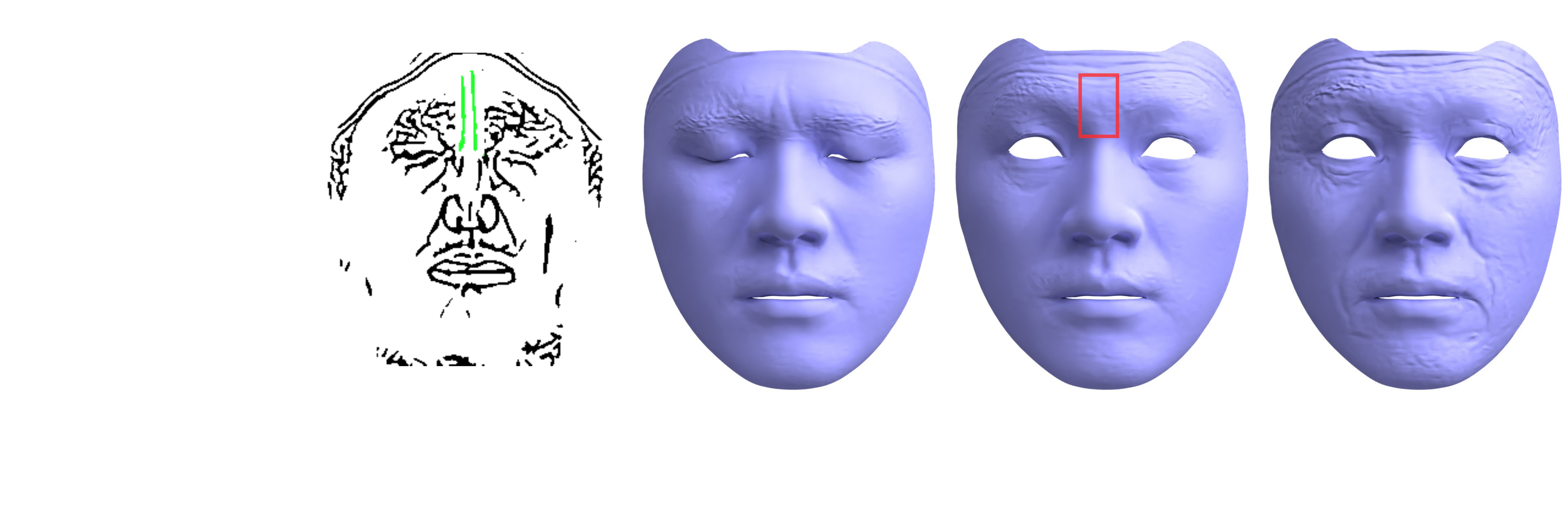}
  \end{minipage}
  \caption{
    SEMM allows the user to manipulate facial details by (i) drawing/erasing wrinkle lines and changing (ii) expression and (iii) age.
    Realistic wrinkles can be generated by drawing lines (shown in green) on the face.
    Generated details can be animated when changing expression and transform properly during aging.
  }
  \label{fig:teaser}
\end{figure}

Morphable face models\cite{egger20203d} capture the statistical distribution of human faces, which provides them with capabilities to generate and edit 3D faces.
Therefore, they are widely used in face reconstruction\cite{blanz1999morphable}, expression animation\cite{cao2014displaced}, and interactive editing\cite{lau2009face}.
In these applications, facial details play a vital role in conveying the perception of expression and age and enhancing the realism of the generated face.
For example, during animation, facial details appear or disappear as muscles contract or relax, vividly reflecting the subtle expression.
During interactive editing, animators may wish to manipulate wrinkles at specific positions.
However, current morphable models usually only represent the large-scale facial geometry, making the results of the above applications over-smooth and unrealistic, with details absent.
A morphable model for facial details is still missing to the best of our knowledge.

In this work, we augment classic 3D morphable models (3DMMs) in representing facial details by proposing a Structure-aware Editable Morphable Model (SEMM).
Specifically, a detail model synthesizes a displacement map that encodes detail geometric information, which is then applied to a mesh generated by a 3DMM to get a high-fidelity 3D face.
We design a separate detail model because mixing low and high frequencies hinders the learning of high frequencies \cite{huynh2018mesoscopic}, and the effectiveness of learning high-frequency facial details separated from large-scale geometry is verified in \cite{golovinskiy2006statistical,cao2015real,chen2019photo}.
With careful design choices, our model is compatible with widely used large-scale face models, and can be seamlessly integrated into the animation pipeline to produce detail animation consistent with the large scale, despite modeling details separately.

Morphable models often assume that faces can be aligned to a fixed template, which holds for the large-scale shape but cannot account for wrinkle details.
For example, forehead wrinkles exhibit thin line structure and may vary in branch numbers on different subjects, making it challenging to define and compute the alignment.
As observed in \cite{paysan2010statistical}, wrinkle lines are almost always imperfectly aligned in the training data, resulting in averaged details and over-smooth generated shape.
We also find missing wrinkles when directly modeling displacement maps in training an autoencoder, as the used reconstruction loss abruptly increases when wrinkles misalign even slightly.
To tackle this issue, we first extract wrinkle lines that encode the structure of details on the facial surface.
Inspired by \cite{Park_2019_CVPR,chibane2020ndf,yenamandra2021i3dmm} that use distance functions in implicit shape modeling, we develop a distance field representation of the wrinkle lines.
The reconstruction loss on distance fields gradually descends as wrinkles start to align, providing meaningful gradients for autoencoder learning.
Therefore, we propose to combine displacement maps and distance fields to construct implicit correspondences of facial details and more accurately model the wrinkle structure.

We instruct the model to generate accurate wrinkle structure on the displacement map, by first ensuring the consistency between the generated displacements and distance field, and then supervising the distance field to preserve the structure of wrinkle lines.
Specifically, we propose an autoencoder that reconstructs a pair of displacement map and distance field map from a latent code, and train the autoencoder adversarially with a discriminator to enforce consistency.
The consistency can lead to more precise wrinkle structure during autoencoding and enable a user to manipulate the details by editing the wrinkle lines.

The latent space of morphable models is often divided into identity and expression.
Additionally, our detail model allows age control, another key semantic factor that provides finer granularity of facial detail editing.
To allow effective expression and age control while preserving identity, we propose two transformation modules to regress direction vectors of changes in the latent space, and supervise them with expression- and age-specific discriminator outputs. 
The direction vectors, as suggested by \cite{gansteerability,zhuang2020enjoy}, permit semantic editing while better maintaining identity.
To meaningfully control the latent space, we adopt expression blendshape weights and age values as control parameters, which are intuitive and compatible with the facial animation pipeline. 
Our method achieves qualitatively and quantitatively better expression editing than previous methods and enables the effective control of detail aging, which is not shown in previous methods.

To summarize, our contributions are: (i) the first attempt to propose an editable morphable model that can animate details by editing wrinkle lines, changing expression weights and age values, (ii) a distance field-based autoencoder network to better model detail structure and give intuitive control over the wrinkles, and (iii) two transformation modules to model expression and age changes, which achieve both accurate representation and effective editing of these two semantic factors.
Code and model will be released.

\section{Related Work}

{\bf Morphable Face Models.} 
Since the pioneering work of \cite{blanz1999morphable}, morphable face models have been widely investigated in the literature.
\cite{blanz1999morphable,booth20163d,gerig2018morphable} analyze the identity variation in neutral facial shape by principal component analysis.
\cite{parke1974parametric,lewis2014practice,neumann2013sparse,wu2016anatomically} construct person-specific linear models to describe expression variations.
Furthermore, \cite{vlasic2006face,cao2013facewarehouse,brunton2014multilinear,yang2020facescape} model the joint distribution of identity and expression by constructing a multilinear model.
After statistical analysis of the facial geometry, morphable models can generate 3D faces from a compact latent space and perform expression editing\cite{amberg2009weight,lau2009face,jiang2019disentangled,chandran2020semantic} through latent space editing.

To improve the representation ability of morphable models, which are often linear, several extensions are made to add nonlinearity.
Some methods combine linear models with nonlinear jaw and neck articulation\cite{li2017learning} or a physical model \cite{ichim2017phace}.
\cite{ranjan2018generating,bagautdinov2018modeling,abrevaya2018multilinear,bouritsas2019neural} propose to learn nonlinear morphable models using autoencoder architectures.
Generative adversarial networks are also explored to perform 3D face modelling\cite{slossberg2018high,abrevaya2019decoupled,shamai2019synthesizing,cheng2019meshgan}. 
Please refer to \cite{egger20203d} for a comprehensive survey.
All the above models focus on large-scale geometry only.
Our nonlinear detail model can be used on top of those morphable models to jointly fit the distribution in a complete frequency domain of both large-scale and details.

It is challenging to model details by current morphable models, as facial details are difficult to align in registration, which is a required process to establish dense scan correspondences.
In practice, \cite{paysan2010statistical} observes that details are often averaged out during morphable model construction. 
To mitigate this issue, \cite{bolkart2015groupwise} proposes a method to iteratively construct a morphable model and improve registration correspondences.
\cite{yenamandra2021i3dmm} represents facial geometry as a deep implicit function and builds a morphable model which automatically establishes correspondences between scans.
Our model builds better semantic correspondences of facial details by a novel method to model the distance field of wrinkle lines.

{\bf 3D Facial Detail Animation and Manipulation.}
Wrinkle formation is strongly correlated with physical facial layers such as elastic fibers and muscles\cite{serup2006handbook,igarashi2007appearance}, which share the same topology among people\cite{radlanski2012face}.
This fact enables modeling details either physically or in a data-driven manner.
Several methods model the facial skin through physics to simulate wrinkle effects\cite{wu1999simulating,boissieux2000simulation,li2007modeling}.
However, those methods require a lot of computation time and hand-tuned physical parameters, therefore we focus on data-driven modeling.
To acquire details from the image data, plenty of methods use either shape from shading\cite{garrido2013reconstructing,shi2014automatic,suwajanakorn2014total,ma2019real} or deep neural networks\cite{richardson2017learning,sela2017unrestricted,sengupta2018sfsnet,jiang20183d,li2018feature,guo2018cnn,zeng2019df2net,tran2019towards,chen2020self,feng2021learning}, but they cannot animate the reconstructed details.
\cite{feng2021learning} uses an encoder-decoder to reconstruct an animatable detailed face from a single image. 
However, it solely relies on shading constraints from images and generates less realistic details than captured 3D scans.
To leverage the scan data, \cite{bickel2008pose,xu2014controllable,fyffe2014driving,bermano2014facial,shin2014extraction} transfer details from a source face to a target face.
They can obtain realistic details, but the details are not specific to the target face.
Multi-identity local models\cite{golovinskiy2006statistical,li2015lightweight,cao2015real,chen2019photo} can be built from high-resolution scan datasets. However, they only model patch-based local detail displacements, effective for detail reconstruction but cannot synthesize detail animation.
\cite{li2020learning,deng2021plausible} can synthesize plausible details from large-scale shape or texture, but aging wrinkles are hard to synthesize as they cannot be fully reflected from large-scale.
\cite{yang2020facescape,li2020dynamic} can synthesize high-fidelity animatable faces given an image or scan. However, they assume that all the details in input faces are static, thus cannot handle dynamic details in inputs.
Several methods allow users to intuitively create new wrinkles using sketches\cite{bando2002simple,li2007wrinkle,xu2014controllable,kim2015interactive}, but they cannot animate the generated wrinkles.
Our model first represents facial details and their changes caused by structure, expression, and age in a unified latent space. Thus, it provides easier ways to animate and manipulate facial details.

\section{Overview}

\begin{figure*}[t]
  \centering
  \includegraphics[width=\linewidth]{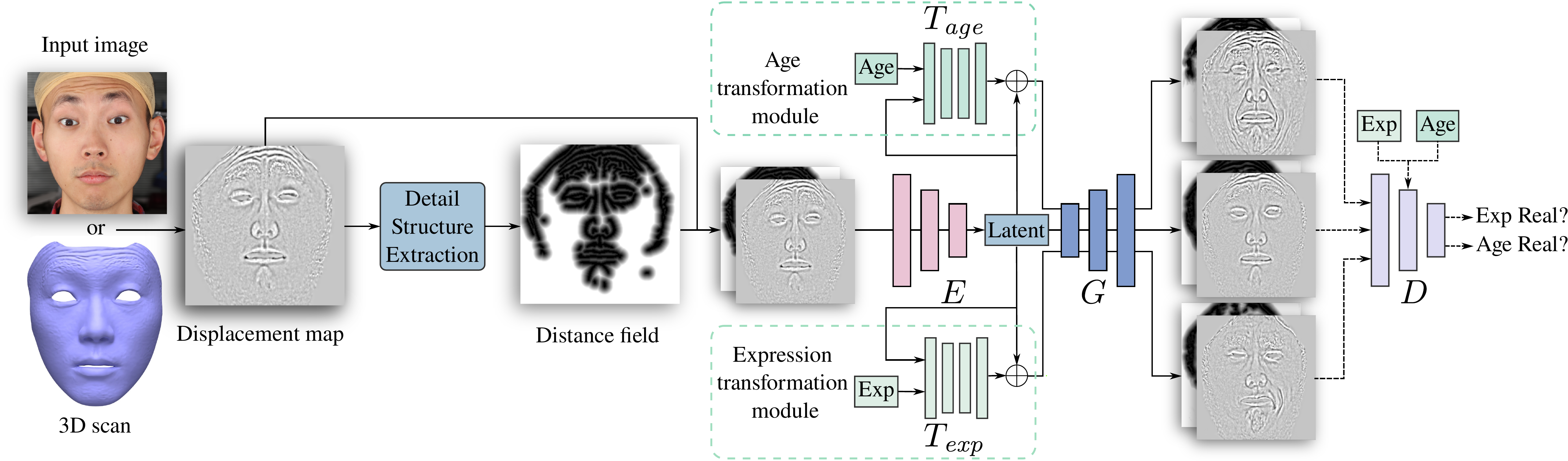}
  \caption{Overview of our method.
}
  \label{fig:pipeline}
\end{figure*}

SEMM serves as a detail model that extends a large-scale 3D morphable model in \cite{yang2020facescape} to represent shape details.
Recall that the large-scale model generates a face mesh, whose expression is controlled by blendshape weights\cite{lewis2014practice}.
The detail model is designed to be compatible with the large-scale model both in shape representation and control parameters.
For shape representation, the detail model synthesizes a displacement map that encodes surface details.
It is then combined with the mesh to get a high-fidelity 3D face.
The process is natively supported by modern graphics hardware\cite{cao2015real}.
The same blendshape weights are used to control the latent code of the detail model to generate expression animation.
Therefore, the generated facial animation is consistent with large scale and details.

Fig. \ref{fig:pipeline} illustrates the pipeline of our method.
To manipulate facial details, we first get original details from an input image or scan and represent them on a displacement map, from which we then extract a distance field encoding the detail structure (Sec. \ref{sec:detail_process}).
The original details only represent a static input, thus we find their latent code for manipulation.
$E$ and $G$ are used to encode and generate (middle path through $G$) both the displacement map and distance field, and are adversarially trained with discriminator $D$ to model the joint displacement-distance distribution (Sec. \ref{sec:dpmap_df_model}).
We add $T_{\mathit{exp}}$ and $T_{\mathit{age}}$ to transform the latent code to enable expression and age editing (upper and lower paths through $G$), and modify $D$ to supervise expression and age (Sec. \ref{sec:exp_age_edit}). Finally, Sec. \ref{sec:experiments} shows experiments of expression, age and structure editing of facial details.

\section{Detail shape processing}
\label{sec:detail_process}

\begin{figure}[t]
  \centering
  \includeinkscape[width=.98\linewidth]{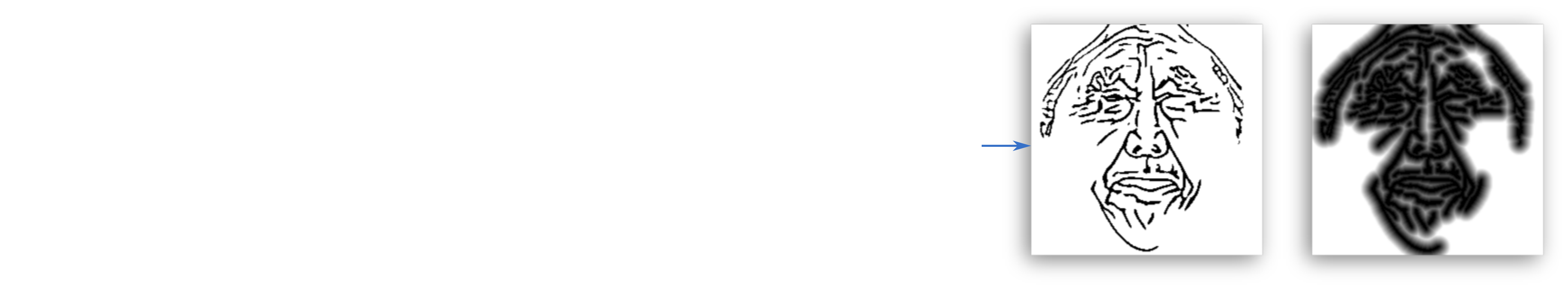}
  \caption{We extract a distance field from a displacement map by (a) removing low-frequency components, (b) extracting detail lines, and (c) applying distance transform.}
  \label{fig:data_process}
\end{figure}

We represent facial details as a displacement map for its efficiency and common use in 3D mesh animation pipelines.
Each pixel is parameterized in the mesh's UV space and encodes a signed displacement from the mesh along the surface normal direction.

Either a face image or a 3D scan can be used to obtain a displacement map of input details, which enables further manipulation.
Specifically, we first fit the large-scale morphable model to the image or scan.
For an image input, we then extract a texture map from it and use a Pix2PixHD\cite{wang2018high} network trained on \cite{yang2020facescape} to reconstruct a displacement map.
For a scan input, it is smoothed via Laplacian smoothing\cite{taubin95}, and the difference between the original and smoothed scan is baked into UV space as a displacement map. We use the scan dataset from \cite{yang2020facescape} for training, and the scans are similarly processed into displacement maps. 
Before modeling, we filter out low-frequency components in the displacement maps, which do not affect detail rendering (see Fig. \ref{fig:data_process} (a)), but may hinder the learning of high frequencies\cite{huynh2018mesoscopic}.

We propose to extract the spatial location of wrinkle lines to describe detail structure.
Inspired by, but different from \cite{cao2015real} which extract wrinkle patches on the lines, we incorporate the lines into the model to supervise it to generate accurate structure.
On the filtered displacement maps, wrinkles are shaped as lines because of how they are formed.
We perform denoising and use a sketch simplification network\cite{simo2016learning} to extract the lines (Fig. \ref{fig:data_process} (b)). 
The extracted lines depict the occupancy of facial details in the UV space.

However, the wrinkle lines are imperfectly aligned, which we find unsuitable for modeling.
As wrinkle lines can be viewed as 2D shapes, we seek a solution from recent progress in shape modeling\cite{Park_2019_CVPR,yenamandra2021i3dmm}, which has shown improved correspondences using a signed distance function.
We adopt the unsigned distance \cite{chibane2020ndf} to model thin lines.
Specifically, we convert the line map to a distance field by Euclid distance transform\cite{li2019linestofacephoto} to obtain each pixel's distance to the nearest detail line. 
Following \cite{tsdf,chibane2020ndf}, we truncate the distance value to $5\%$ of the map width to concentrate on the neighborhood of lines, producing the final distance field maps (Fig. \ref{fig:data_process} (c)).

\section{Modeling displacements and distance fields}
\label{sec:dpmap_df_model}

{\bf Autoencoding the joint distribution.} 
For the displacement map $\mathbf{x}^d$ and its distance field map $\mathbf{x}^s$, we model their {\it joint} distribution $p(\mathbf{x}^d,\mathbf{x}^s)$.
The joint distribution is obtained by mapping a latent code $\mathbf{z} \sim p(\mathbf{z})$ to a pair $(\widehat{\mathbf{x}}^d,\widehat{\mathbf{x}}^s)$ via a generator network, for which we use StyleGAN2\cite{karras2020analyzing} to leverage its high synthesis quality.
As the latent code explains variations both in the displacement map and distance field, we input $(\mathbf{x}^d,\mathbf{x}^s)$ to an encoder following the design in \cite{park2020swapping} for fast inference.
The autoencoding process can be formulated as:
\begin{equation}
\label{eq:autoencode}
\mathbf{z} = E(\mathbf{x}) = E(\mathbf{x}^d, \mathbf{x}^s),\qquad \widehat{\mathbf{x}}_{\mathit{rec}} = G(\mathbf{z})
\end{equation}
where $\mathbf{z}$ is a compact latent code of 576 dimensions, $E$ is the encoder, $G$ is the generator, $\mathbf{x} = (\mathbf{x}^d, \mathbf{x}^s)$ and $\widehat{\mathbf{x}}_{\mathit{rec}}=(\widehat{\mathbf{x}}_{\mathit{rec}}^d,\widehat{\mathbf{x}}_{\mathit{rec}}^s)$ are the input and reconstructed samples respectively.

\begin{figure}[t]
  \centering
  \begin{minipage}[t]{.48\textwidth}
    \centering
    \includeinkscape[width=.95\linewidth]{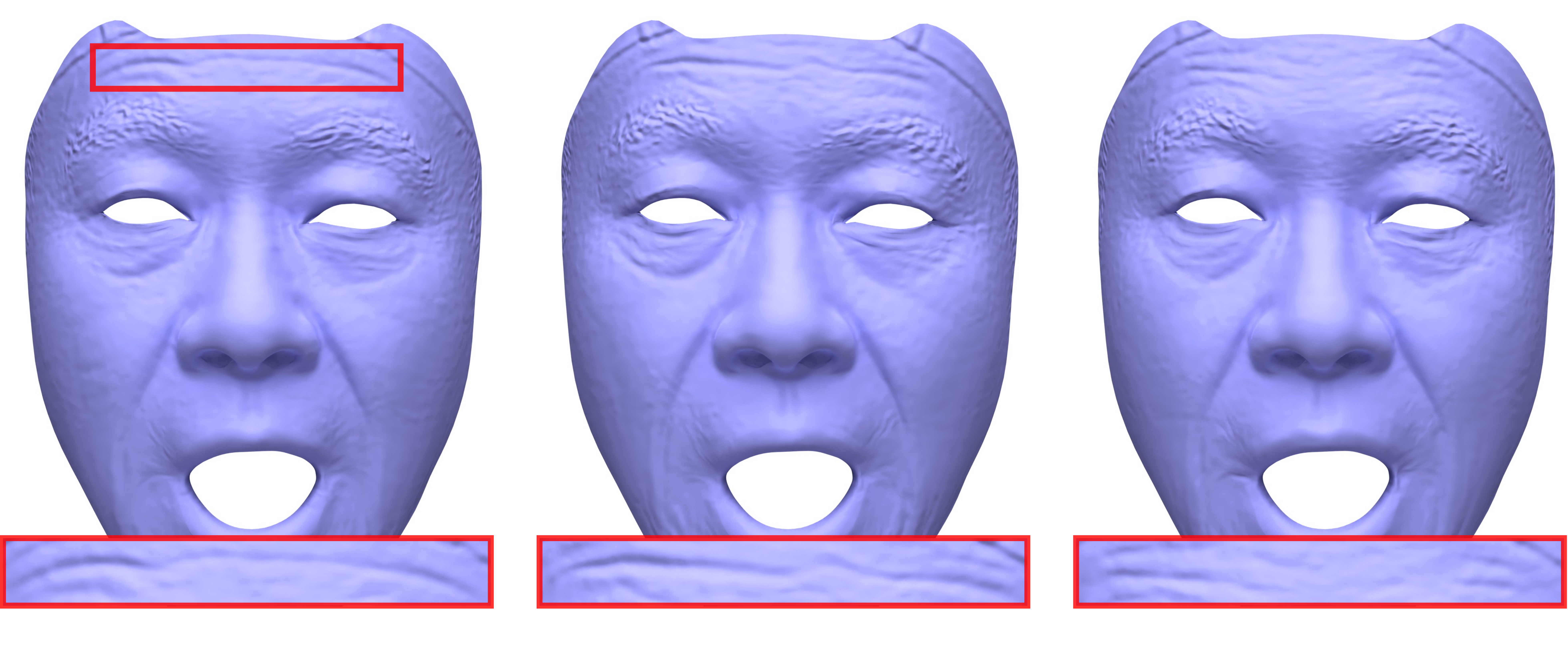}
  \end{minipage}
  \begin{minipage}[t]{.48\textwidth}
    \centering
    \includeinkscape[width=.95\linewidth]{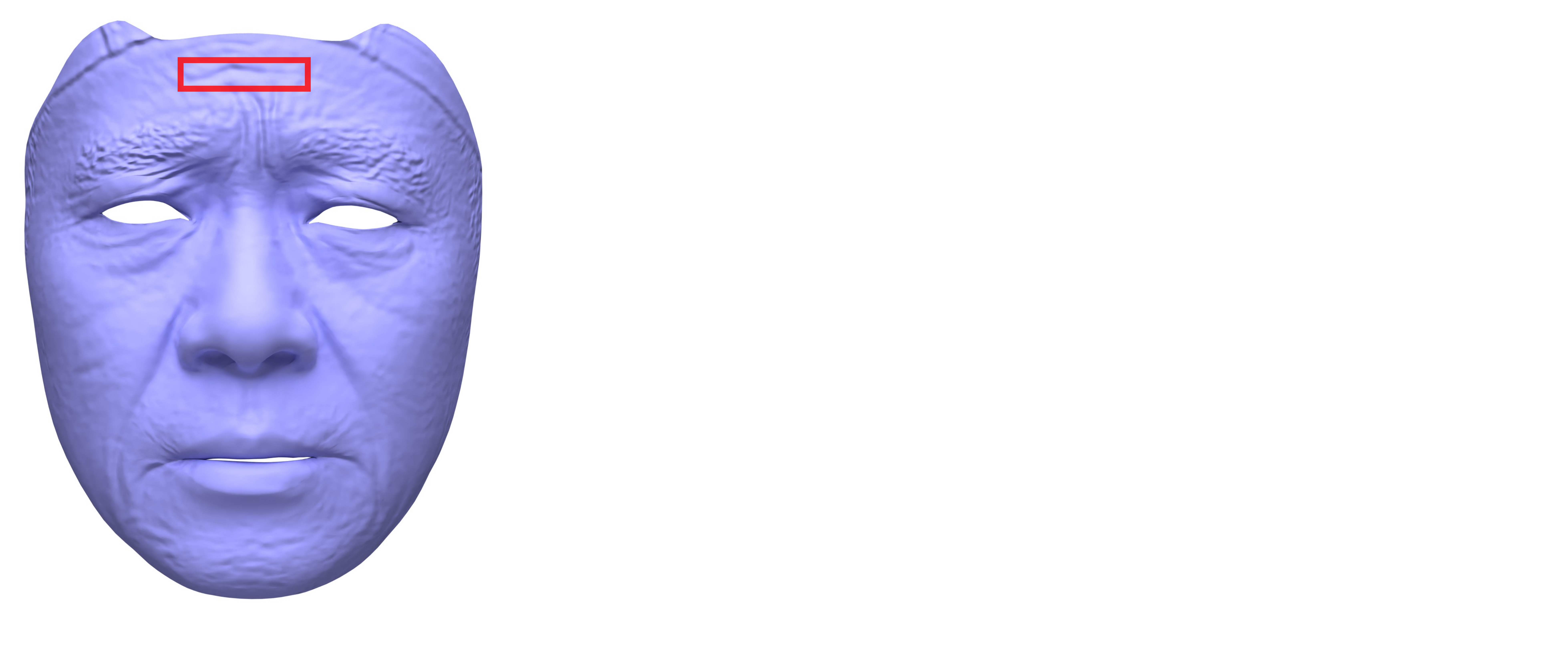}
  \end{minipage}
  \caption{Autoencoding results of jointly modeling displacements and distance fields (middle) and modeling displacements only (right).}
  \label{fig:ablate_no_seg}
\end{figure}

{\bf Consistency via the discriminator.} To both enable realistic synthesis and ensure the consistency between synthesized $\widehat{\mathbf{x}}^d$ and $\widehat{\mathbf{x}}^s$, we add a discriminator which inputs $(\widehat{\mathbf{x}}^d,\widehat{\mathbf{x}}^s)$ in training.
The consistency, as a benefit from joint distribution modeling, was previously exploited in \cite{li2021semantic,zhang2021datasetgan} to generate segmentation labels for unannotated images.
While they model a joint image-label distribution for semi-supervised learning, here we consider the consistency between displacement maps and distance fields for structure-aware generation and editing.
The consistency can guide the detail structure in the generated displacements to follow the distance field.
This fact allows us to add supervision on the distance field during training to more accurately reconstruct the detail structure.
Compared with modeling displacements only, modeling the joint distribution can better preserve wrinkles during autoencoding, as shown in Fig. \ref{fig:ablate_no_seg}.
Additionally, the consistency allows a user to manipulate the detail structure intuitively by editing the input distance field, as shown in Sec. \ref{sec:app}.

{\bf Structure-aware reconstruction and editing.}
For a generated output $\widehat{\mathbf{x}}_{*} = (\widehat{\mathbf{x}}_{*}^d, \widehat{\mathbf{x}}_{*}^s)$, the above discriminator ensures consistency between $\widehat{\mathbf{x}}_{*}^d$ and $\widehat{\mathbf{x}}_{*}^s$. 
Additionally, to enforce $\widehat{\mathbf{x}}_{*}$ is consistent with a target $\mathbf{x}_{*} = (\mathbf{x}_{*}^d, \mathbf{x}_{*}^s)$ in Eqn. \ref{eq:rec_obj}, we design a reconstruction loss $\ell_{\mathit{rec}}$ that can be decomposed into two terms:
\begin{equation}
\label{eq:rec}
\ell_{\mathit{rec}}(\widehat{\mathbf{x}}_{*}, \mathbf{x}_{*}) = \ell_{\mathit{FM}}(\widehat{\mathbf{x}}_{*}, \mathbf{x}_{*}) + \lambda_{\mathit{df}} \ell_{\mathit{df}}(\widehat{\mathbf{x}}_{*}, \mathbf{x}_{*})
\end{equation}
First, we use the feature matching loss $ \ell_{\mathit{FM}}$ to formulate a multi-scale reconstruction loss:
\begin{equation}
\label{eq:LFM}
\ell_{\mathit{FM}}(\widehat{\mathbf{x}}_{*}, \mathbf{x}_{*}) = \sum_{i=1}^{T}\frac{1}{N_{i}}\left\Vert D^{(i)}(\widehat{\mathbf{x}}_{*})-D^{(i)}(\mathbf{x}_{*}) \right\Vert_1
\end{equation}
This loss follows \cite{wang2018high}, but on a StyleGAN2 discriminator $D$ with $T$ layers and $N_{i}$ components in the $i$th layer.
Second, we use a distance field loss $\ell_{\mathit{df}}$ to help reconstruct the detail structure:
\begin{equation}
\ell_{\mathit{df}}(\widehat{\mathbf{x}}_{*}, \mathbf{x}_{*}) = \left\Vert \min(\mathbf{\widehat{x}}_{*}^{s},\delta)-\min(\mathbf{x}_{*}^{s},\delta)\right\Vert_1
\end{equation}
where $\mathbf{x}_{*}=(\mathbf{x}_{*}^{d},\mathbf{x}_{*}^{s})$ corresponds to the target displacement map and its distance field, $\widehat{\mathbf{x}}_{*}=(\widehat{\mathbf{x}}_{*}^{d},\widehat{\mathbf{x}}_{*}^{s})$ corresponds to the generated pair, and $\delta$ is set to $5\%$ of the map width as a threshold to concentrate the loss to areas near details.
Note that $\ell_{\mathit{df}}$ is not related to displacement maps.

To explain our design of $\ell_{\mathit{rec}}$: First, we use feature matching loss (Eqn. \ref{eq:LFM}) instead of L1 loss because L1 focuses on per pixel reconstruction and tends to ignore structure similarity, as it strongly penalizes structure-similar details with merely slight misalignment. 
Second, by supervising the distance field instead of wrinkle lines (Fig. \ref{fig:data_process} (c)), misaligned wrinkles start to have some overlap, which leads to a gradually descending loss value as wrinkles start to align.
Both of the two tend to encourage structure similarity and tolerate slight misalignment of wrinkles, which we believe can lead to a more compact and well-behaved detail representation. 

Supervising the distance field also enables a user to perform wrinkle line editing.
Specifically, after the wrinkle line map is extracted from the displacement map (Fig. \ref{fig:data_process} (b)), the user draws or erases lines on the line map, before it is converted to a distance field.
The original displacement map and the edited distance field are then passed through $E$ and $G$ to get the edited displacement map that is consistent with user edits.
This behavior is achieved via a training objective $\mathcal{L}_{\mathit{struct}}$. 
In training, we sample a displacement map $\mathbf{x}_{\mathit{sty}}^d$ and a distance field $\mathbf{x}^s$ from different faces to simulate user editing, and encode them jointly to synthesize the result as
\begin{equation}
\label{eq:mix}
    \widehat{\mathbf{x}}_{\mathit{mix}} = G(E(\mathbf{x}_{\mathit{sty}}^d,\mathbf{x}^s)).
\end{equation}
We use the distance field loss to enforce the model to preserve the input distance field $\mathbf{x}^s$
\begin{equation}
 \mathcal{L}_{\mathit{struct}} = \mathop{\mathbb{E}_{\mathbf{x},\mathbf{x}_{\mathit{sty}}}} [\lambda_{\mathit{df}} \ell_{\mathit{df}}(\widehat{\mathbf{x}}_{\mathit{mix}}, \mathbf{x})]
\end{equation}
Notice that because the ground truth edited displacement map is unknown, here we can not use $\ell_{\mathit{FM}}$ as Eqn. \ref{eq:rec} did. 
However, an adversarial objective in Eqn. \ref{eq:adv_all} supervises the generated displacement map to be realistic. 
In the supplementary material, we provide more discussion on wrinkle line editing.

\section{Expression and age editing}
\label{sec:exp_age_edit}

As a deep generative model, the detail model acquires a semantic understanding of the modeled data in its latent space\cite{gansteerability}, thus enabling expression and age semantic editing. 
To be compatible with the facial animation pipeline and intuitive to control, instead of exposing the latent space to the user, we adopt blendshape weights and age values as control parameters.
Blendshape weights \cite{lewis2014practice} are commonly used in expression animation, where each dimension corresponds to the activation strength of a predefined basic expression. 
This definition makes it both compatible with the large-scale mesh animation and easy to tune by hand.
When a user edits the expression or age of facial details, first, she specifies the target blendshape weights or age. 
Then, the detail model generates edited details from transformed latent codes.
For training, we get each sample's blendshape weights and age, which are usually available in a scan dataset.
Note that these annotations are {\it not} used in testing.
As detailed below, we design the training to achieve effective expression and age editing while maintaining the identity.

{\bf Edit-guided transformation modules.}
We add two networks that output latent direction vectors, which permit semantic editing while better maintaining identity\cite{gansteerability,zhuang2020enjoy}.
Specifically, given a latent code $\mathbf{z}$ of original details and target blendshape weights $\widetilde{\mathbf{e}}$, the expression transformation module regresses a direction vector.
It is then added to the latent code to decode a sample that exhibits desired expression changes. Age editing is done similarly, and they can be formulated as
\begin{equation}
\label{eqn:exp_edit}
\widehat{\mathbf{x}}_{\mathit{exp}}=G\big(\mathbf{z} + T_{\mathit{exp}}(\mathbf{z}, \widetilde{\mathbf{e}})\big)
\end{equation}
\begin{equation}
\label{eqn:age_edit}
\widehat{\mathbf{x}}_{\mathit{age}}=G\big(\mathbf{z} + T_{\mathit{age}}(\mathbf{z}, \widetilde{a})\big)
\end{equation}
where $\widetilde{\mathbf{e}}$ and $\widetilde{a}$ are the target expression and age, $T_{\mathit{exp}}$ and $T_{\mathit{age}}$ are the two transformation modules to model the difference between the target and the current latent codes, $\widehat{\mathbf{x}}_{\mathit{exp}}$ and $\widehat{\mathbf{x}}_{\mathit{age}}$ are the decoded samples that exhibit specified expression and age changes. 
Experiments in Sec. \ref{sec:ablation} demonstrate that by using transformation modules, we can achieve both accurate representation and effective editing of details.

{\bf Expression- and age-specific discriminator outputs.} To supervise the editing using expression and age annotations, we modify the discriminator to separately output expression and age information in a multi-task manner\cite{liu2019few,choi2020stargan}.
It assumes the dataset is divided into $n_{\mathit{exp}}$ key expressions and $n_{\mathit{age}}$ evenly spaced age groups, and outputs a vector of length $n_{\mathit{exp}} + n_{\mathit{age}}$.
Each element in the vector describes whether the sample exhibits its corresponding expression or stays in its age group.
Specifically, if we want the sample to exhibit the $i$th expression $\mathbf{e}_{*}$ and in the $j$th age group $a_{*}$, the $i$th and $n_{\mathit{exp}}+j$th output $D_{\mathbf{e}_{*}}$ and $D_{a_{*}}$ will be used to formulate the adversarial loss as
\begin{equation}
\label{eqn:fake}
\ell_{\mathit{GAN}}^{\mathit{Fake}}(\mathbf{x}_{*}, \mathbf{e}_{*}, a_{*}) = \log (1-D_{\mathbf{e}_{*}}(\mathbf{x}_{*})) +  \log (1-D_{a_{*}}(\mathbf{x}_{*}))
\end{equation}
\begin{equation}
\label{eqn:real}
\ell_{\mathit{GAN}}^{\mathit{Real}}(\mathbf{x}_{*}, \mathbf{e}_{*}, a_{*}) = \log (D_{\mathbf{e}_{*}}(\mathbf{x}_{*})) +  \log (D_{a_{*}}(\mathbf{x}_{*}))
\end{equation}
where the first term of both Eqn. \ref{eqn:fake} and \ref{eqn:real} enforces sample $\mathbf{x}_{*}$ to exhibit expression $\mathbf{e}_{*}$, and the second is to constrain age $a_{*}$.

We want the reconstructed sample $\widehat{\mathbf{x}}_{\mathit{rec}}$ (Eqn. \ref{eq:autoencode}) and the mix-generated sample $\widehat{\mathbf{x}}_{\mathit{mix}}$ (Eqn. \ref{eq:mix}) to preserve the original expression $\mathbf{e}$ and age $a$, and the edited samples (Eqn. \ref{eqn:exp_edit} and \ref{eqn:age_edit}) to reach target expression or age while keeping the other attribute fixed.
The total adversarial objective can be formulated as
\begin{equation}
\label{eq:adv_all}
\begin{aligned}
\mathcal{L}_{\mathit{GAN}} = \mathop{\mathbb{E}_{\mathbf{x},\mathbf{x}_{\mathit{exp}},\widetilde{a},\mathbf{x}_{\mathit{sty}}}}[\ell_{\mathit{GAN}}^{\mathit{Real}}(\mathbf{x}, \mathbf{e}, a) + \ell_{\mathit{GAN}}^{\mathit{Fake}}(\widehat{\mathbf{x}}_{\mathit{rec}},\mathbf{e},a) 
\\ +\ell_{\mathit{GAN}}^{\mathit{Fake}}(\widehat{\mathbf{x}}_{\mathit{exp}},\widetilde{\mathbf{e}},a)  + \ell_{\mathit{GAN}}^{\mathit{Fake}}(\widehat{\mathbf{x}}_{\mathit{age}},\mathbf{e},\widetilde{a})+ \ell_{\mathit{GAN}}^{\mathit{Fake}}(\widehat{\mathbf{x}}_{\mathit{mix}},\mathbf{e},a)]
\end{aligned}
\end{equation}

{\bf Maintaining the identity.} 
To learn a meaningful latent space and maintain the identity when no editing or expression editing is performed, during training, we sample a pair of samples $\mathbf{x}, \mathbf{x}_{\mathit{exp}} \sim \mathcal{X}$ of the same person, with different expressions $\mathbf{e} = e(\mathbf{x})$ and $\mathbf{\widetilde{e}} = e(\mathbf{x}_{\mathit{exp}})$ respectively.
We enforce the encoded latent code can reconstruct the input $\mathbf{x}$, and the expression-edited latent code obtained in Eqn. \ref{eqn:exp_edit} can reconstruct  $\mathbf{x}_{\mathit{exp}}$ using the reconstruction loss $\ell_{\mathit{rec}}$ defined in Eqn. \ref{eq:rec}:
\begin{equation}
\label{eq:rec_obj}
\begin{aligned}
\mathcal{L}_{\mathit{rec}} = \mathop{\mathbb{E}_{\mathbf{x},\mathbf{x}_{\mathit{exp}}}}[\ell_{\mathit{rec}} (\widehat{\mathbf{x}}_{\mathit{rec}}, \mathbf{x}) 
+\ell_{\mathit{rec}} (\widehat{\mathbf{x}}_{\mathit{exp}},\mathbf{x}_{\mathit{exp}})
].
\end{aligned}
\end{equation}

Notice that we do not have the data of the same subject at different ages.
Thus, we use a cycle consistency objective to maintain original identity during age editing. 
During training, we randomly sample a target age $\widetilde{a} \sim \mathcal{U}(16, 70)$, and perform age transformation according to Eqn. \ref{eqn:age_edit}. Then we enforce the reconstruction of original details when transforming the edited sample back to the original age $a$ as
\begin{equation}
\mathcal{L}_{\mathit{cyc}} = \mathop{\mathbb{E}_{\mathbf{x},\widetilde{a}}}\Big[\ell_{\mathit{rec}}\Big(G\big(E(\widehat{x}_{\mathit{age}})+T_{\mathit{age}}(E(\widehat{x}_{\mathit{age}}),a)\big), \mathbf{x}\Big)\Big]
\end{equation}

{\bf Full objective.}
The total training objective can be formulated as
\begin{equation}
\begin{aligned}
\min_{E,T_{\mathit{age}},T_{\mathit{exp}},G}\;\max_{D_{*}}\; \mathcal{L}_{\mathit{rec}} &+ \lambda_{\mathit{GAN}} \mathcal{L}_{\mathit{GAN}} \\ + \mathcal{L}_{\mathit{struct}} &+ \lambda_{\mathit{cyc}}\mathcal{L}_{\mathit{cyc}}
\end{aligned}
\end{equation}

\section{Experiments}
\label{sec:experiments}

\textbf{Implementation Details.}
We use the publicly available dataset from \cite{yang2020facescape} for training, which is randomly divided into 14,930 training and 1,623 test samples.
We model the displacement map at 256x256 resolution, which we find is enough to encode wrinkle-level details.
Displacement maps and distance fields are normalized to approximately the same standard deviation to balance the discriminator's attention and improve adversarial training.
From the training dataset, we obtain 51-dimensional blendshape weights from the known expression and one-dimensional age from the demographic information.
The expression and age transformation modules are parameterized by 4-layer MLPs.
We use the network architecture of $G$ and $D$ in \cite{karras2020analyzing} and $E$ in \cite{park2020swapping}.
Following \cite{karras2020analyzing}, our model uses a non-saturating adversarial loss with R1 regularization \cite{mescheder2018training} to stabilize training.
As we use a multi-task discriminator, $\mathbf{x}_{\mathit{sty}}^d$ and $\mathbf{x}^s$ in Eqn. \ref{eq:mix} are drawn from the same expression and the same age group.
We set $\lambda_{\mathit{df}} = 2.5$, $\lambda_{\mathit{GAN}}=0.05$ and $\lambda_{\mathit{cyc}}=1$ to balance the loss terms.
The loss terms can be categorized into groups, and each group shares the same weight, leading to only three above weights.
We set $n_{\mathit{exp}}=20$ and $n_{\mathit{age}}=7$ during the training.
Our model is trained on two RTX 3090s with a batch size of 16 for 17,700k iterations and takes 21 ms to encode-decode a displacement map at test time.

{\bf Detail quantitative metrics.}
Previous methods often conduct qualitative studies alone to evaluate facial details.
In addition, we propose to use LPIPS\cite{zhang2018unreasonable} as a quantitative metric to measure the similarity between two displacement maps.
While LPIPS is originally for measuring natural RGB images, we find it effective to measure the similarity of facial details because it mainly measures the visual similarity and semantic accuracy of wrinkles instead of requiring per-pixel alignment.
In the supplementary material, we also investigate the behavior difference between LPIPS and L1.

\subsection{Comparisons}

We compare our method with FaceScape \cite{yang2020facescape} and DECA \cite{feng2021learning}, which are SOTAs that can both represent facial details and their changes with expressions.
During the comparison, all the methods first obtain a detail representation from an input image, and then use the representation and input target expression parameters to generate details with a different expression.
We use images captured in the dataset from \cite{wang2020single} as test inputs, as it has images of various facial details caused by expressions.
A reference image with the same expression as the target expression parameters is shown for visualization, and is not inputted to any method.
When animating to other expressions, both the large-scale mesh and details deform according to the expression parameters.
We conduct the comparison via (1) qualitative study, (2) user study and (3) quantitative study using LPIPS.
Please see the supplementary material for the user study and quantitative comparison.

\begin{figure*}[ht]
  \centering
  \begin{minipage}[t]{.495\textwidth}
    \centering
    \includeinkscape[width=.98\linewidth]{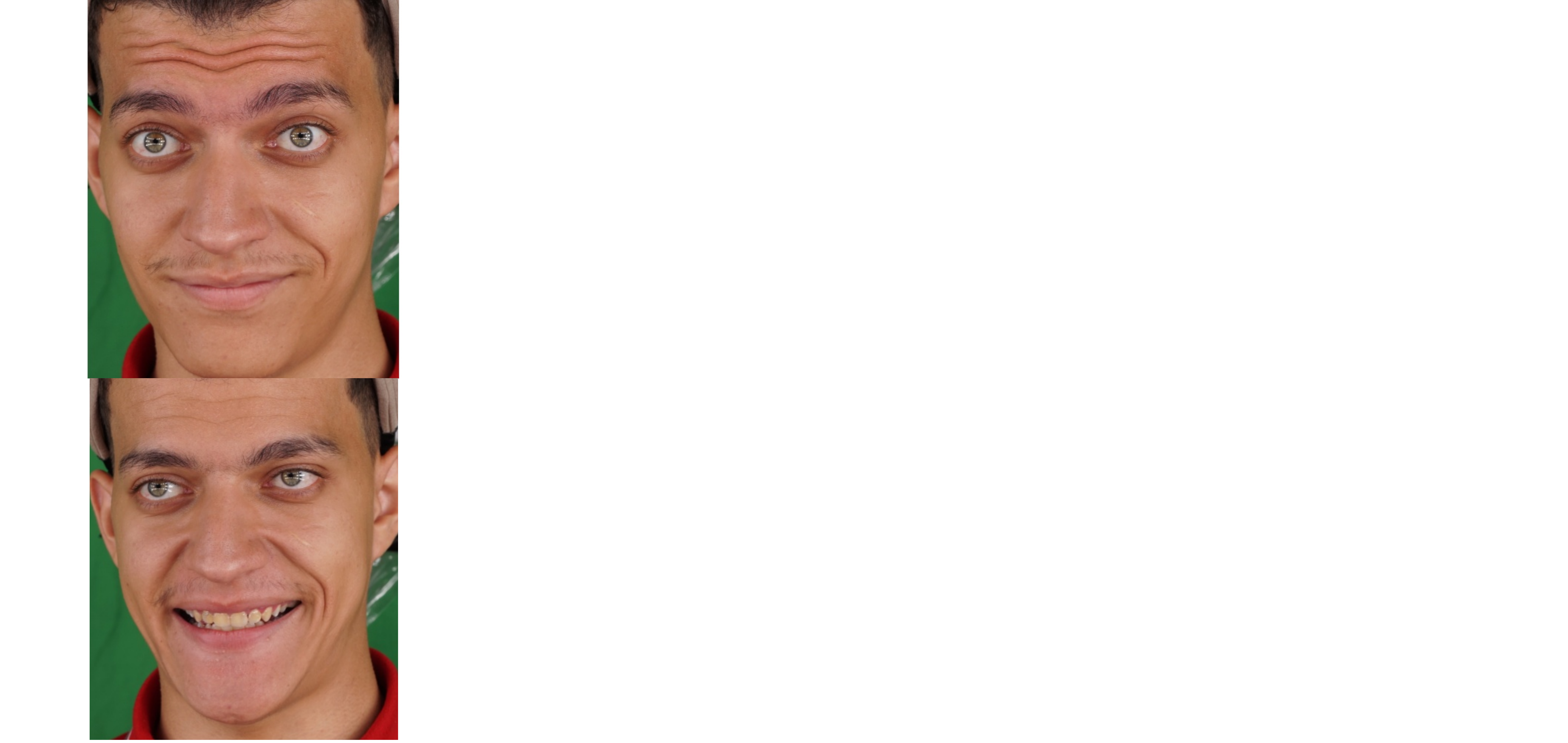}
    \includeinkscape[width=.98\linewidth]{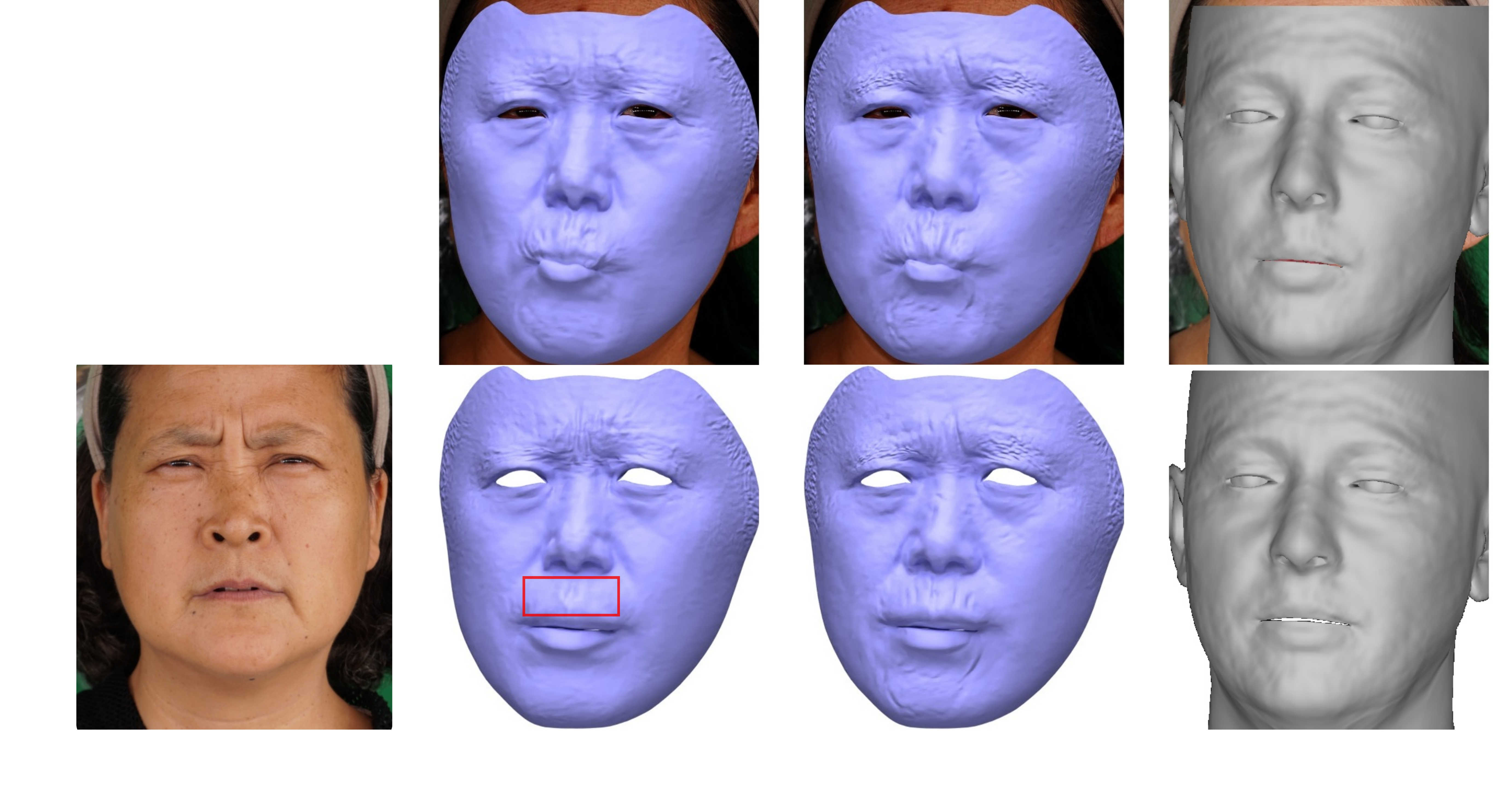}
  \end{minipage}
  \begin{minipage}[t]{.495\textwidth}
    \centering
    \includeinkscape[width=.98\linewidth]{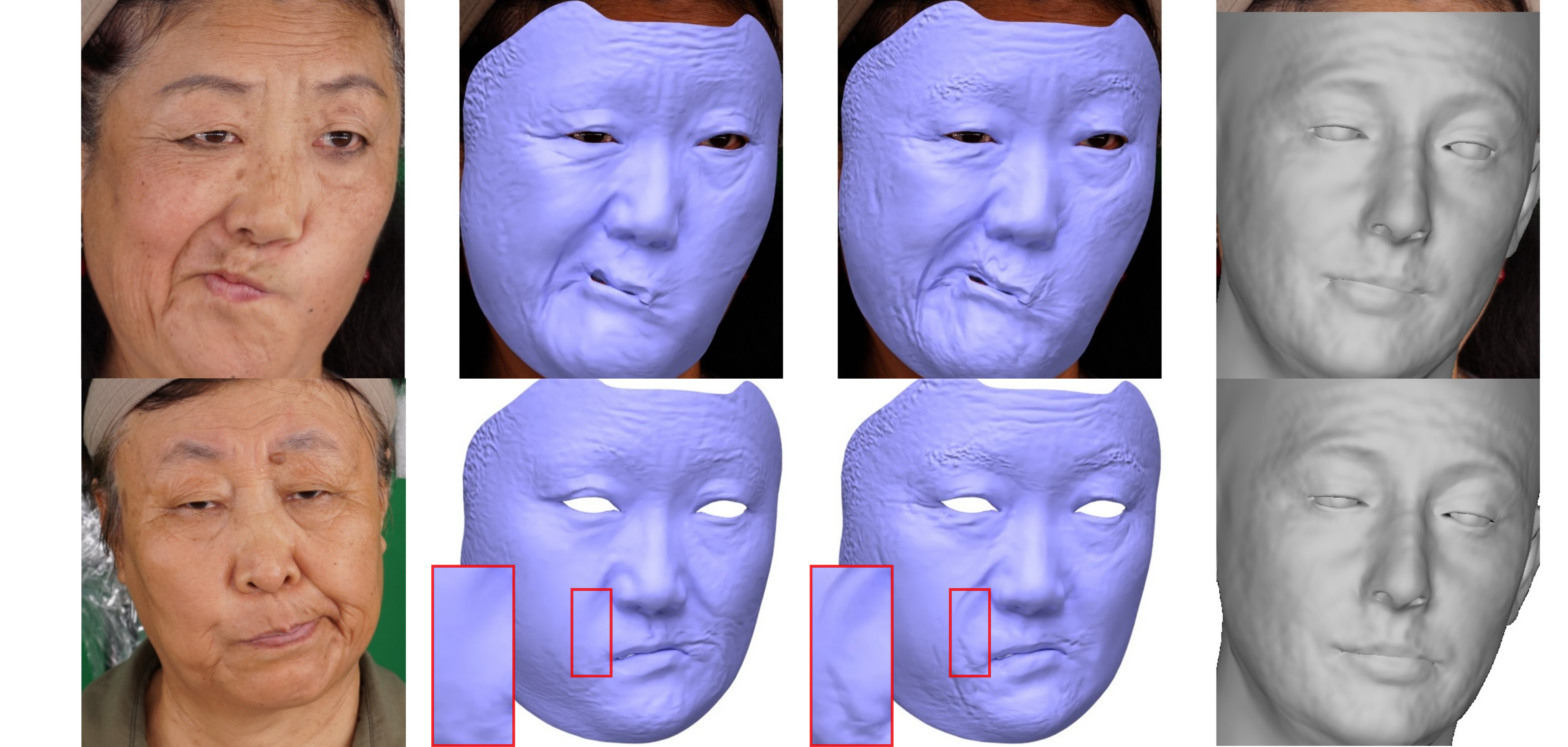}
    \includeinkscape[width=.98\linewidth]{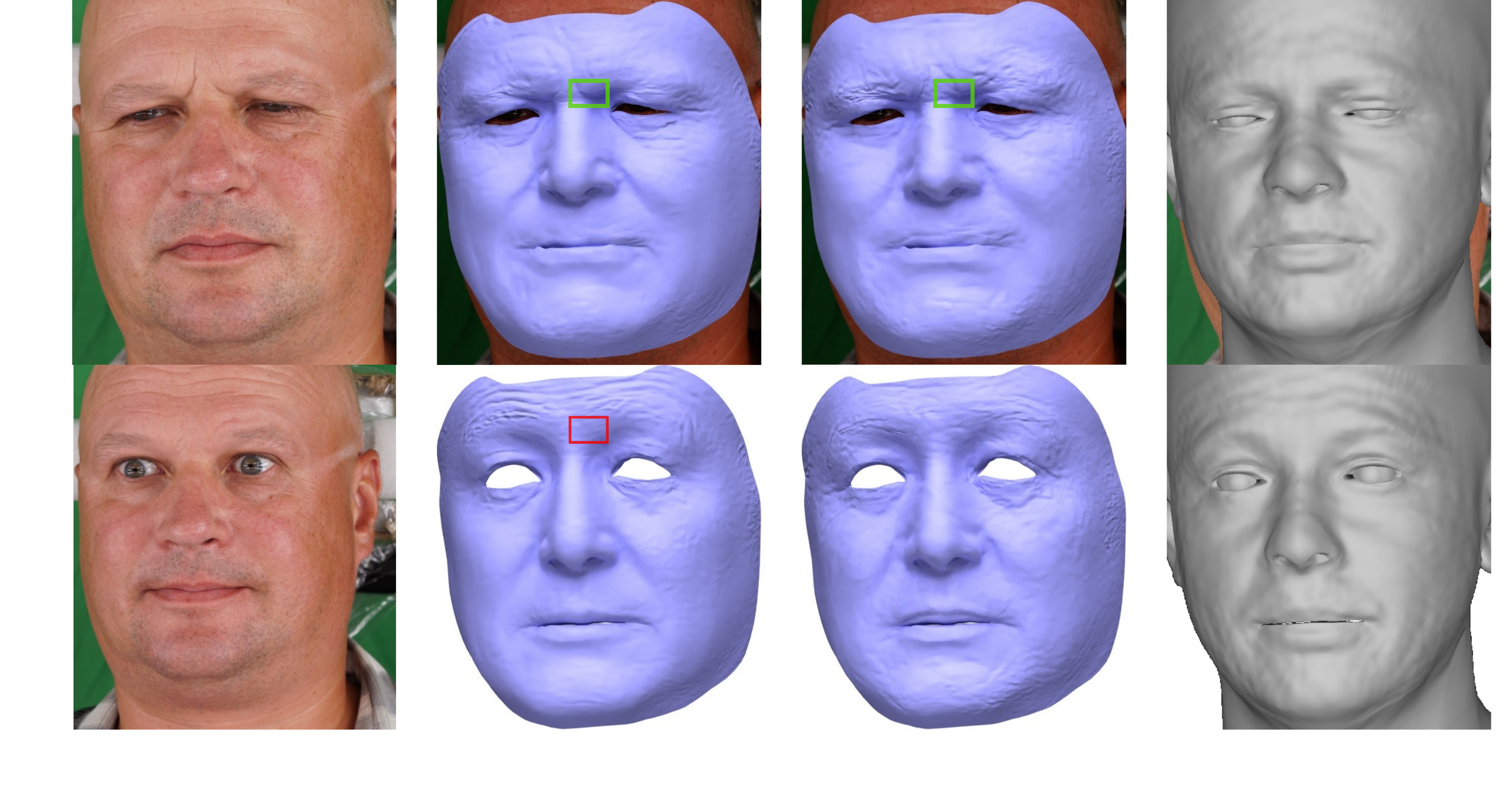}
  \end{minipage}
  \caption{Comparison with previous methods on expression editing.}
  \label{fig:compare_exp}
\end{figure*}

{\bf Qualitative study.} 
For each test case in Figure \ref{fig:compare_exp}, the first row shows the details generated from the original representation, and the second row shows the generated details with the target expression.
FaceScape can generate plausible dynamic details of the target expression, but it assumes all the details in the input images are static.
Here, dynamic details refer to the details that change with expressions, while static details mean the ones invariant to expressions.
Therefore, it cannot correctly animate the dynamic details presented in the input image, causing artifacts (denoted by red boxes).
DECA can properly animate the facial details in arbitrary input expressions.
However, its details are less person-specific, and some age-related wrinkles are absent.
Our method can both generate more diverse dynamic details than other methods (top right, bottom right) and properly animate the input dynamic details.
Specifically, in the bottom-left sample, our model is shown to learn to activate dynamic details according to facial muscles, by deactivating the lip region wrinkles while keeping wrinkles between eyebrows.

\subsection{Ablation studies}
\label{sec:ablation}

\begin{figure}[t]
  \centering
  \begin{minipage}[t]{.495\textwidth}
    \centering
    \includegraphics[width=.98\linewidth]{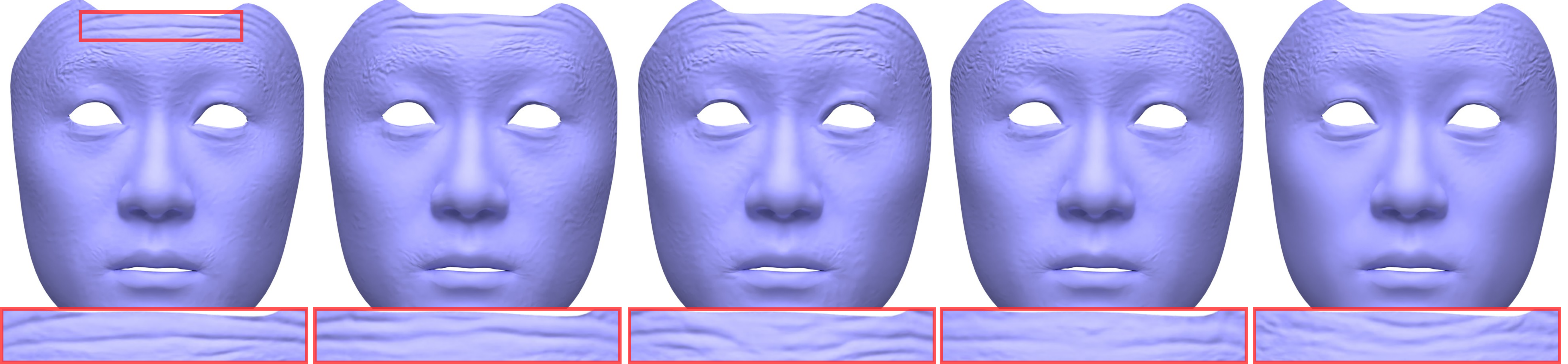}
    \includegraphics[width=.98\linewidth]{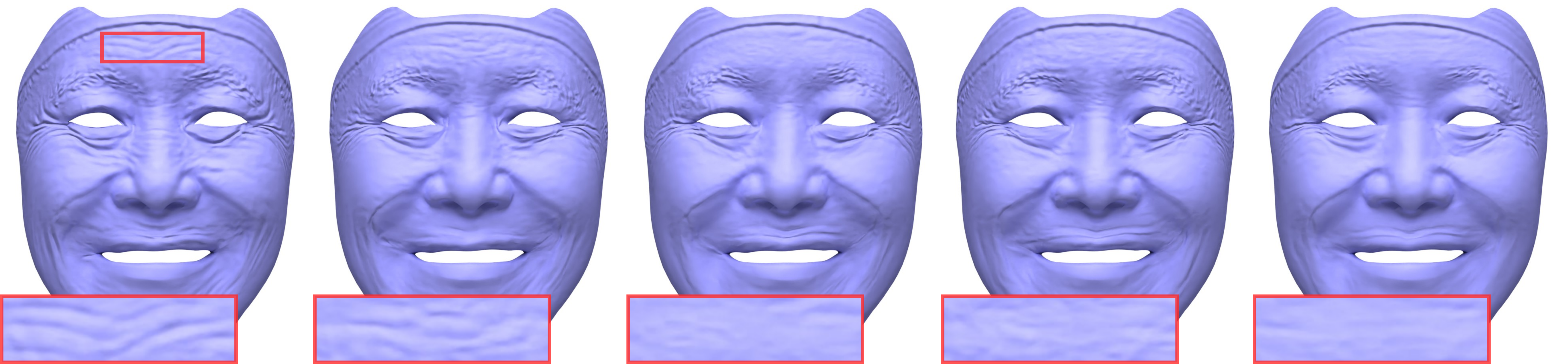}
    \includeinkscape[width=.98\linewidth]{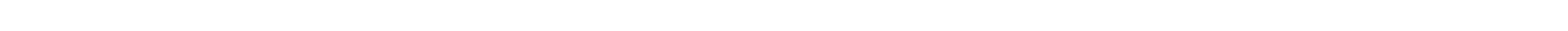}
  \end{minipage}
  \begin{minipage}[t]{.495\textwidth}
    \centering
    \includegraphics[width=.98\linewidth]{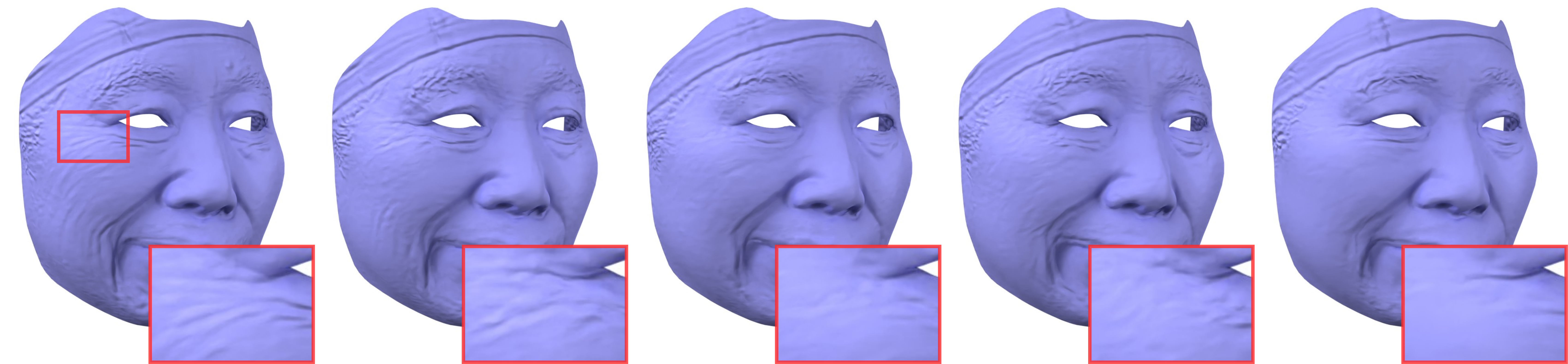}
    \includegraphics[width=.98\linewidth]{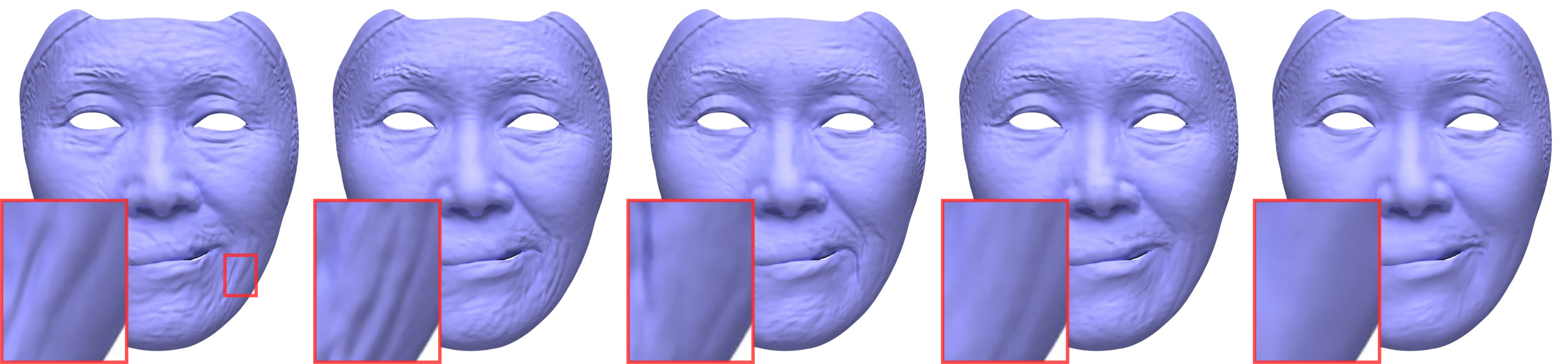}
    \includeinkscape[width=.98\linewidth]{./svg-inkscape/ablate_all/740+8_mouth_left_ablate_all_caption.pdf_tex}
  \end{minipage}
  \caption{Reconstruction results of our model and baselines.}
  \label{fig:ablate_all}
  \end{figure}

\begin{table}[t]
  \centering
  \caption{LPIPS error of our method and baselines.}
  \begin{tabular}{c c c}
    \toprule
Method & reconstruction & editing \\
    \midrule
    w/o $\ell_{\mathit{FM}}$ & 0.1604 & 0.1755 \\
    w/o $\mathit{df}$ &  0.1214 & 0.1481 \\
    w/o $T_{\mathit{exp},\mathit{age}}$ & 0.1263 & 0.1456 \\
Ours & {\bf 0.1134} &  {\bf 0.1455}\\
    \bottomrule
  \end{tabular}
  \label{tab:ablation}
\end{table}

We evaluate three key components of our proposed method by comparing our model with three baselines with ablated components: 
(1) \textbf{Feature matching loss} is evaluated by the w/o $\ell_{\mathit{FM}}$ setting where we replace the feature matching loss with L1 loss.
(2) \textbf{Distance field} is evaluated by the w/o $\mathit{df}$ setting where use detail lines instead of distance fields for joint modeling.
(3) \textbf{Transformation modules} are evaluated by the w/o $T_{\mathit{exp},\mathit{age}}$ setting where we directly concatenate the expression and age parameters to the latent code before decoding.

We use the reconstruction LPIPS and the editing LPIPS to measure the ability to represent input details and make suitable expression changes of facial details.
Specifically, we first encode the input displacement map to a latent code.
We then decode a displacement map using the encoded latent code and calculate the reconstruction error using LPIPS.
We also transform each sample to other expressions and evaluate the LPIPS between the generated details and the ground truth.
The test split of the dataset from \cite{yang2020facescape} is used, because it provides ground truth details of the same person with different expressions.

The average LPIPS errors are shown in Table \ref{tab:ablation}, and some reconstruction samples are shown in Fig. \ref{fig:ablate_all}.
From the results, w/o $\ell_{\mathit{FM}}$ and w/o $\mathit{df}$ give higher errors in both reconstruction and expression editing.
They result in missing or less pronounced wrinkle details than the full model.
This is because in training, they give a stronger penalization on the reconstructed wrinkles that are not pixel-aligned with the input.
W/o $T_{\mathit{exp},\mathit{age}}$ generates some details that are related to the expression, but not in the input (top left, bottom right).
Some individual-specific wrinkles are also absent in its results  (bottom left, top right).
These lead to a larger reconstruction error.

\subsection{Applications}

\begin{figure*}[t]
  \centering
  \begin{minipage}[t]{.485\textwidth}
    \centering
    \includeinkscape[width=.98\linewidth]{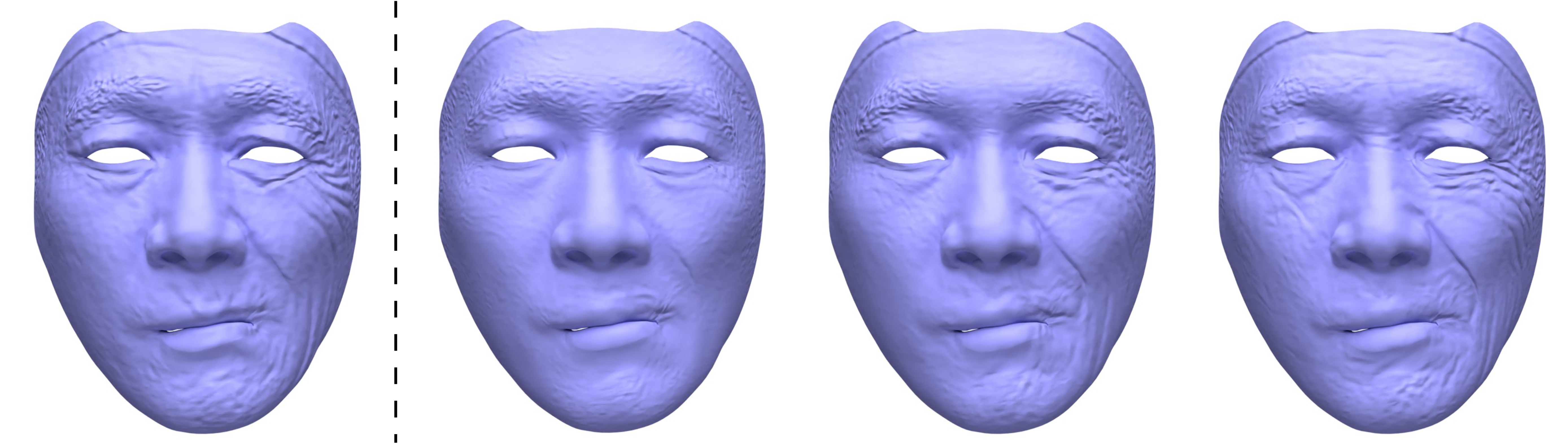}
    \includeinkscape[width=.98\linewidth]{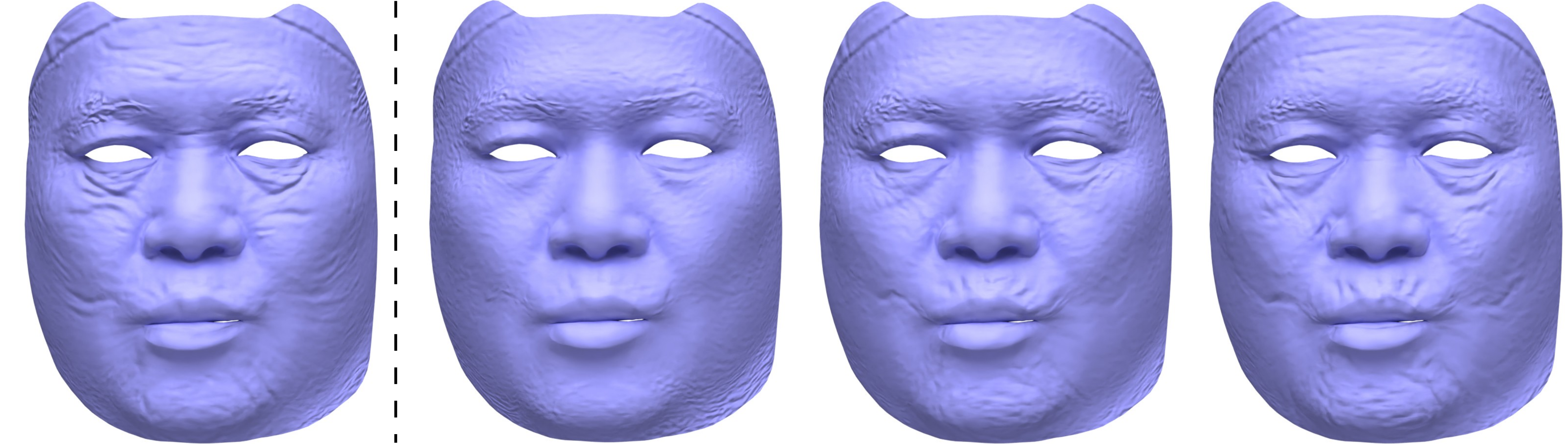}
    \includeinkscape[width=.98\linewidth]{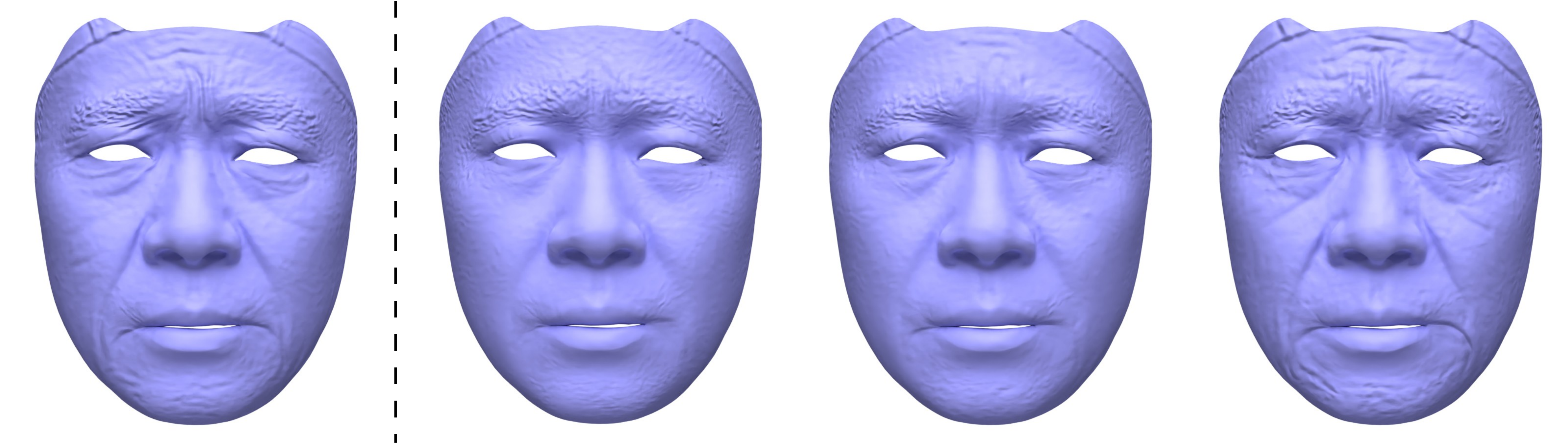}
    \includeinkscape[width=.98\linewidth]{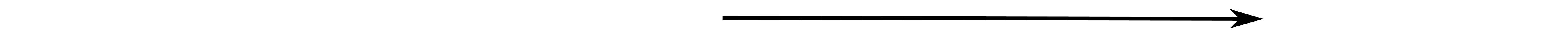}
  \end{minipage}
  \begin{minipage}[t]{.485\textwidth}
    \centering
    \includeinkscape[width=.98\linewidth]{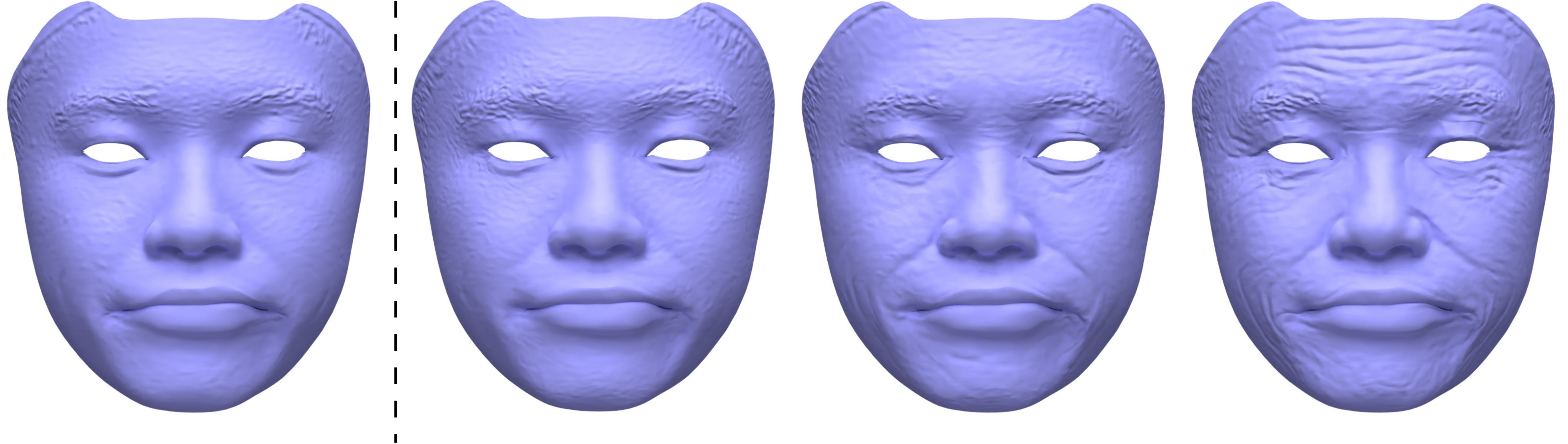}
    \includeinkscape[width=.98\linewidth]{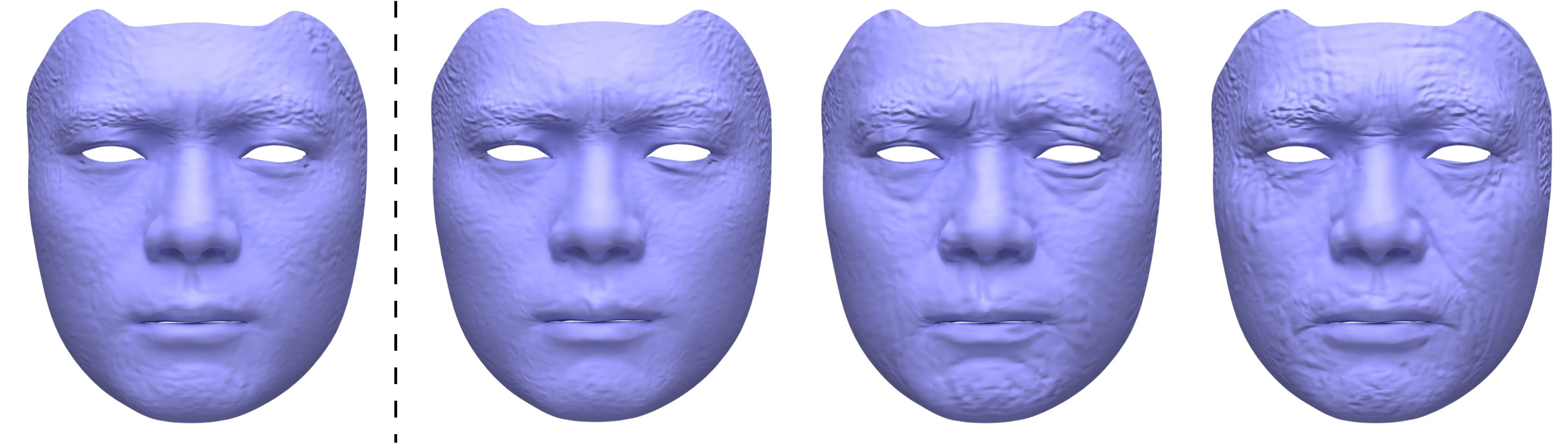}
    \includeinkscape[width=.98\linewidth]{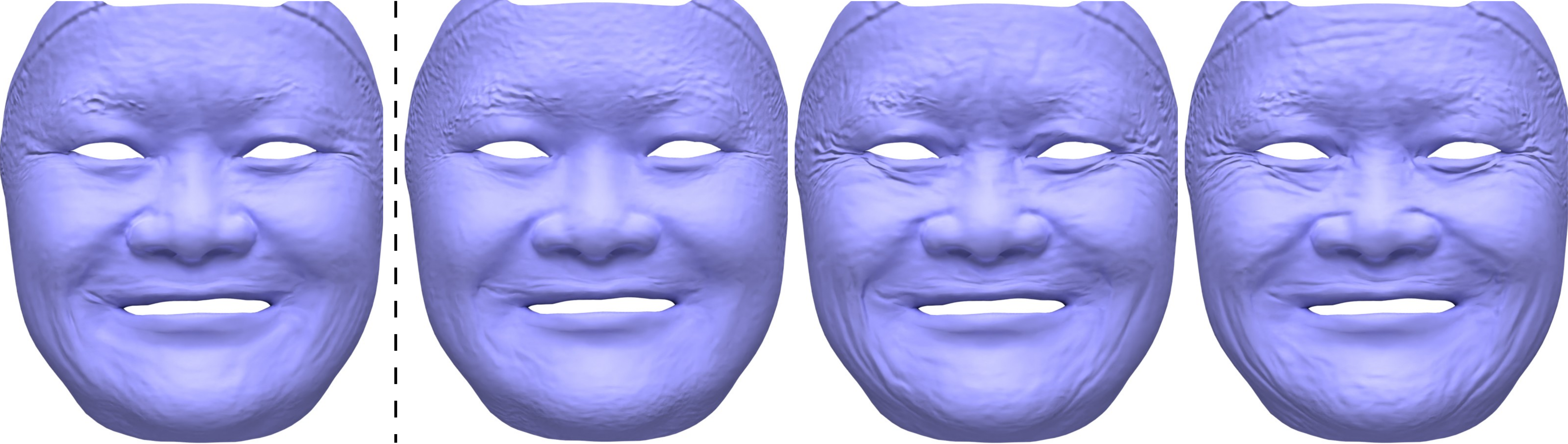}
    \includeinkscape[width=.98\linewidth]{./svg-inkscape/age_progression/age_progression_caption.pdf_tex}
  \end{minipage}
  \caption{Age progression synthesized by our SEMM.}
  \label{fig:age_progression}
\end{figure*}

\begin{figure*}[t]
\centering
\begin{minipage}[t]{.31\textwidth}
  \centering
  \includeinkscape[width=\linewidth]{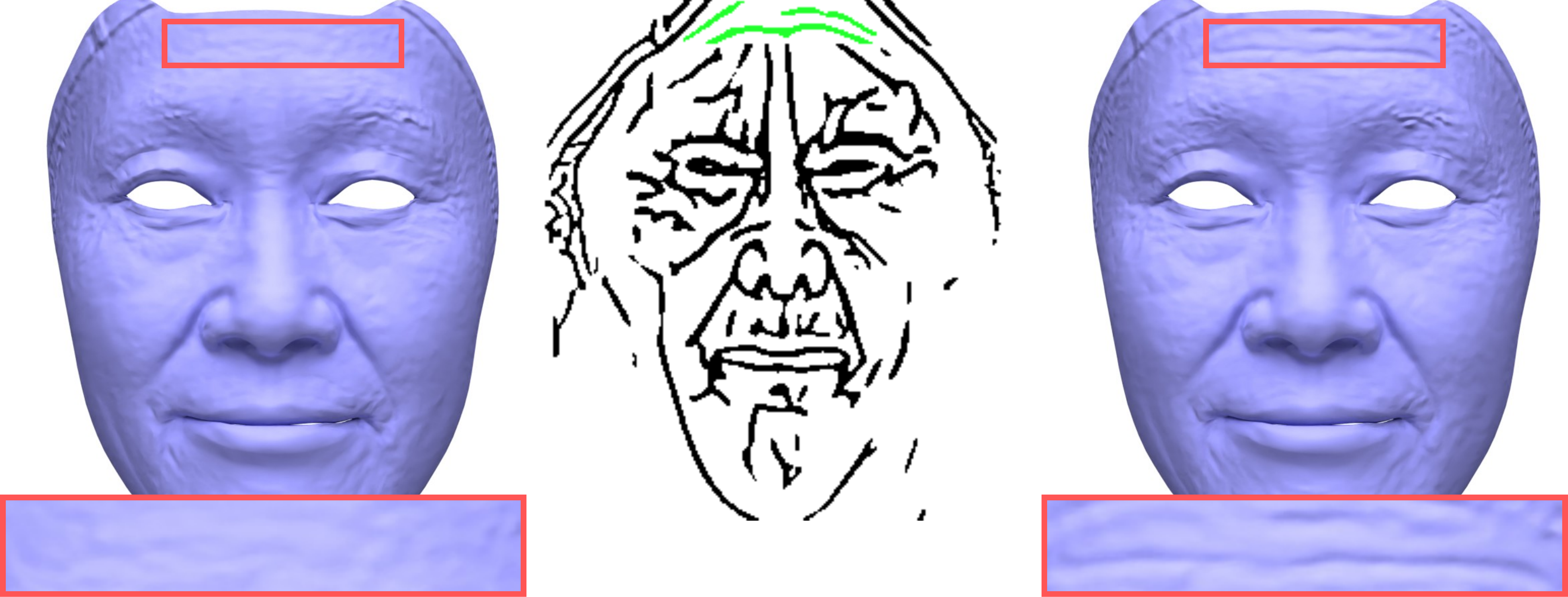}
  \includeinkscape[width=\linewidth]{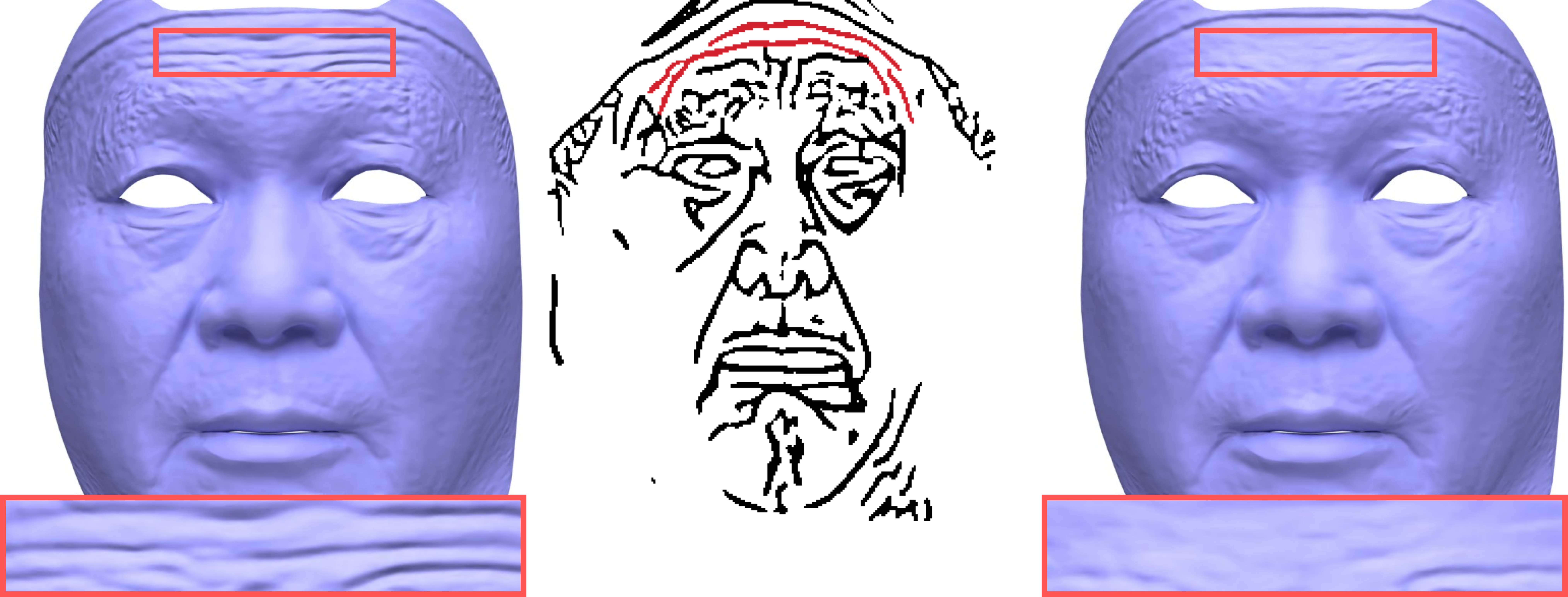}
\end{minipage}
\hspace{.015\textwidth}
\begin{minipage}[t]{.31\textwidth}
  \centering
  \includeinkscape[width=\linewidth]{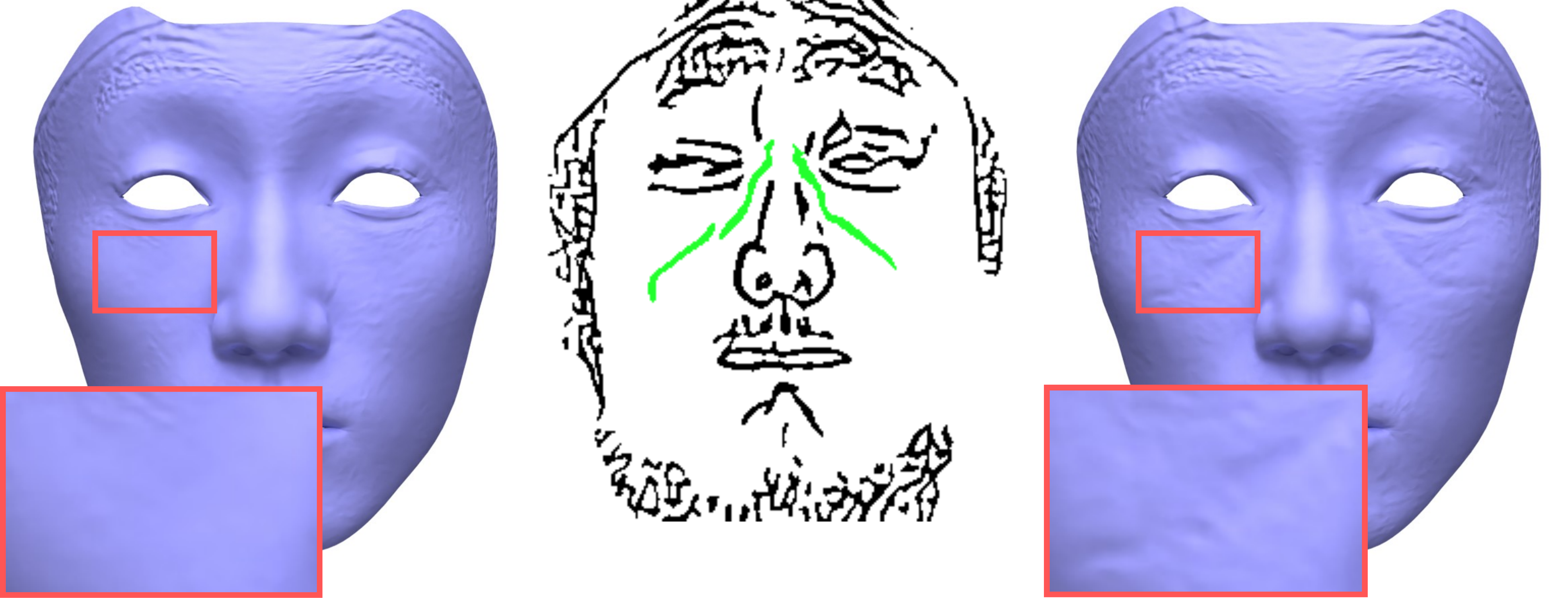}
  \includeinkscape[width=\linewidth]{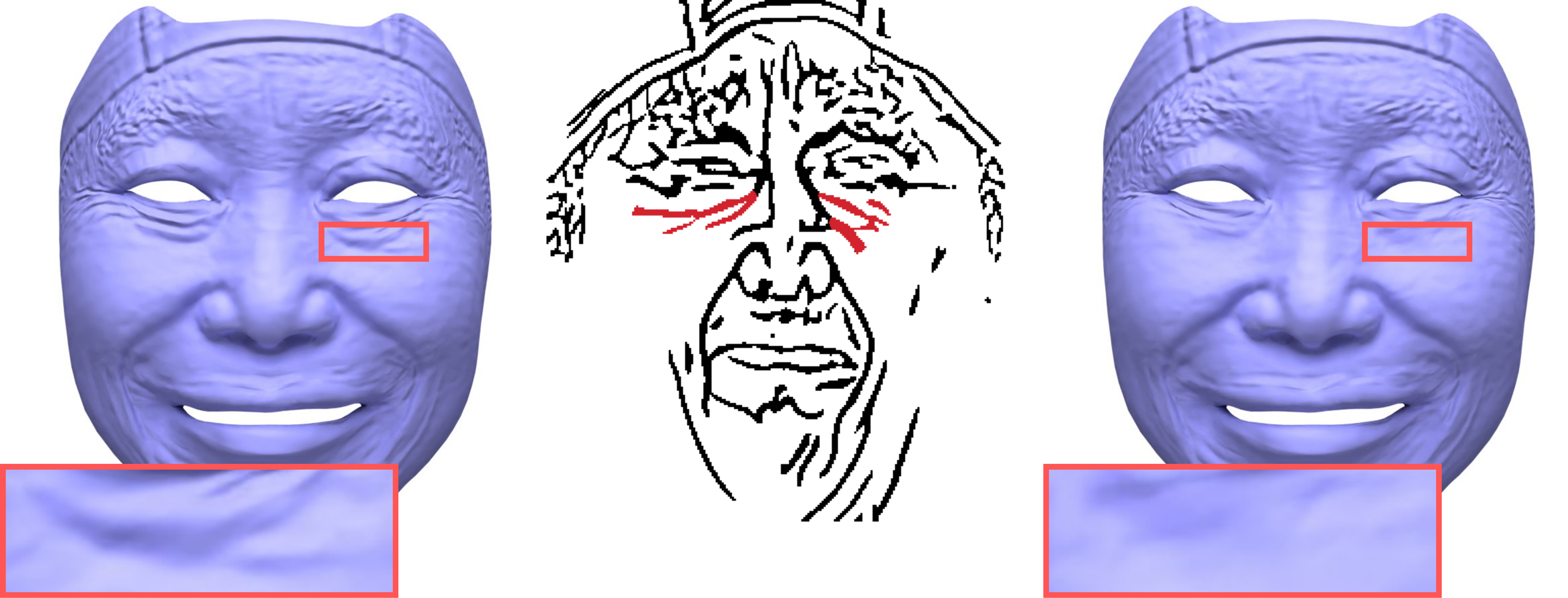}
\end{minipage}
\hspace{.015\textwidth}
\begin{minipage}[t]{.31\textwidth}
  \centering
    \includeinkscape[width=\linewidth]{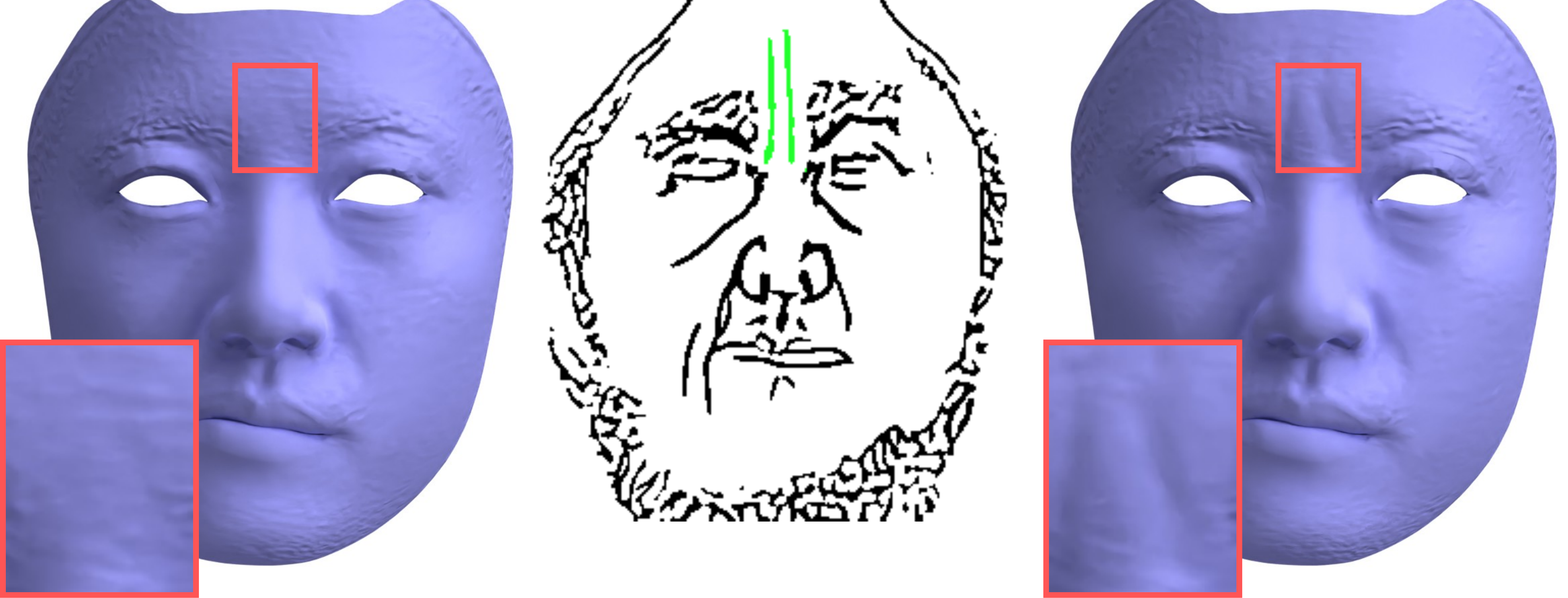}
  \includeinkscape[width=\linewidth]{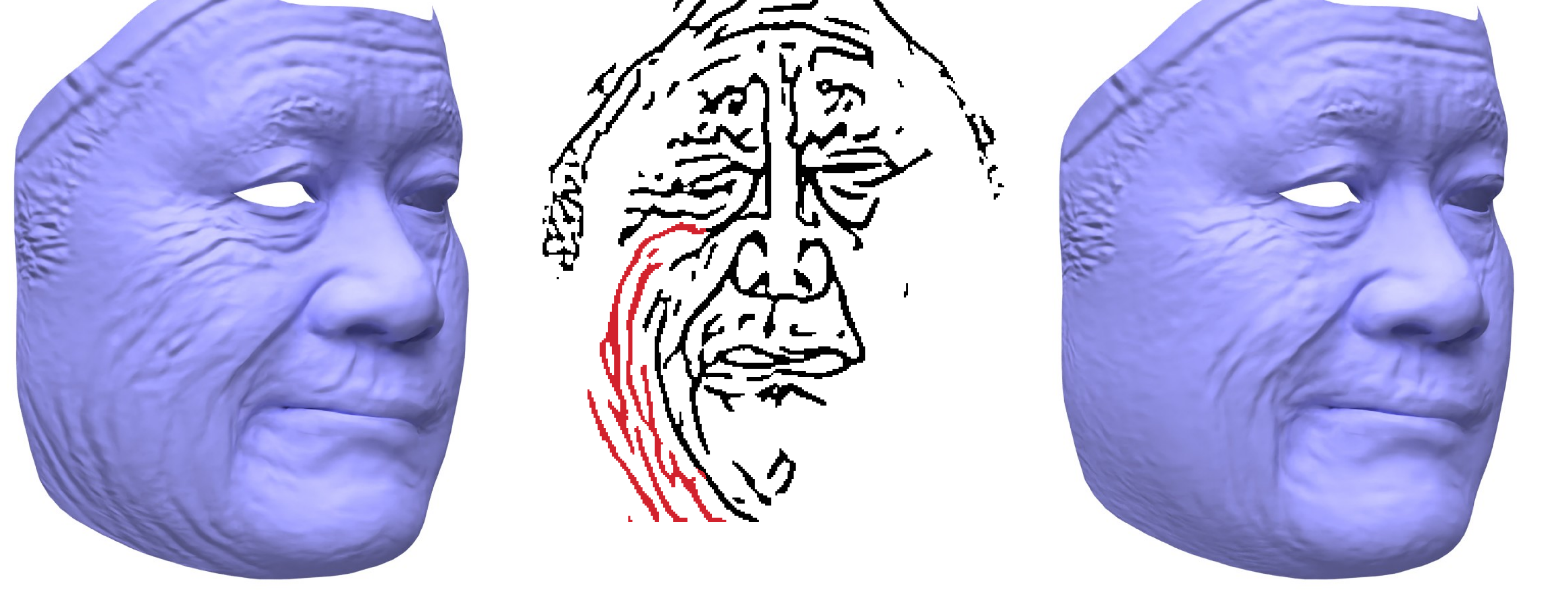}
\end{minipage}
\caption{Users can add or remove details by editing detail lines. Drawn lines are shown in green and erased lines are shown in red.}
\label{fig:sketch_edit}
\end{figure*}

\label{sec:app}
Our model supports meaningful and intuitive ways to animate and manipulate facial details, by changing two semantic factors, expression and age, that often cause facial detail changes.
Besides, by allowing directly editing wrinkle lines, we give users more flexibility to edit facial details. 
More application results are shown in our supplementary video.

{\bf Age progression.}
Our method supports continuous age editing by specifying target age values, as shown in Fig. \ref{fig:age_progression}.
We can both rejuvenate the input (left column) and make the input older (right column).
Diverse types of wrinkles, including forehead wrinkles (top right), eye bags (middle right), and crow's feet (bottom right), can be generated by our model.
Our model works with various non-neutral expressions, and can synthesize and remove details while being compatible with the person's expression and identity.
One interesting phenomenon is that we find our model learns to generate nasolabial folds at an earlier age than other details like forehead wrinkles (top right), which is consistent with the biological aging process\cite{radlanski2012face}.

{\bf Wrinkle line editing.}
Our model allows users to edit facial details intuitively by modifying wrinkle lines.
Specifically, users directly edit the detail lines extracted from the original displacement map, then the edited lines are converted to a distance field to generate edited details together with the original displacement map.
We can effectively synthesize new wrinkles and remove existing ones by drawing and erasing lines, as shown in Fig. \ref{fig:sketch_edit}.
As the edited facial details are still within the representation space of our latent code, they can be further edited using expression and age editing, as shown in Figure \ref{fig:teaser}.

{\bf Blendshape animation.}
Our model supports expression animation using blendshape weights.
The expression editing results shown in Fig. \ref{fig:compare_exp} are also obtained by manipulating the blendshape weights.
It has already indicated that our method can remove the dynamic details of the original expression, generate dynamic details of the new expression and keep the static details. 
For animation results, please refer to our supplementary video.

\section{Limitation}
\begin{figure}[t]
  \centering
  \includeinkscape[width=.5\linewidth]{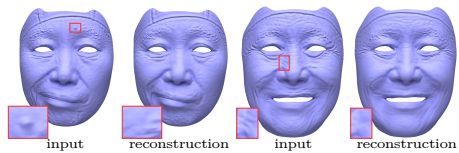}
  \caption{Limitation of our method. Our model cannot reconstruct some rare details like moles and subtle wrinkles.}
  \label{fig:limitation}
\end{figure}

As a data-driven method, our model cannot reconstruct rare details that have not been seen in the training data, like moles and subtle wrinkles shown in Fig. \ref{fig:limitation}.
Pore-level details are still challenging to represent.
However, bumps around eyebrows and hairs, which are from the training data, are represented in our model because of its data-driven nature.

As a detail model, we only modify details during age editing. 
Because there does not exist a method to change the age of a large-scale mesh automatically, we fix the mesh when evaluating the detail aging effects in Fig. \ref{fig:age_progression}. 
This can often generate satisfactory results, but sometimes large nasolabial folds are located in the mesh.
If in need, our model is compatible with an artist who manually edits the mesh to match the specified age value.
Future research can extend our model by simultaneously editing facial details and the mesh.

\section{Conclusion}

In this work, we augment morphable face models in representing detail shape by proposing a detail model.
We propose a detail structure representation based on the distance field of wrinkle lines.
It is then combined with displacement maps in an autoencoder to represent and edit wrinkle structure.
Two transformation modules enable expression and age editing while maintaining identity.
Our model produces detail animation compatible with the large scale, achieves better expression control than previous methods qualitatively and quantitatively, and enables unprecedented age and wrinkle line editing.
These properties make the model useful in the production pipeline of high-fidelity facial animation.

\setlength\parindent{0pt}
{\bf Acknowledgements}
This work was supported by Beijing Natural Science Foundation (JQ19015), the NSFC (No. 62021002, 61727808), the National Key R\&D Program of China 2018YFA0704000, the Key Research and Development Project of Tibet Autonomous Region (XZ202101ZY0019G). This work was supported by THUIBCS, Tsinghua University and BLBCI, Beijing Municipal Education Commission. Feng Xu is the corresponding author.

\bibliographystyle{splncs04}
\bibliography{egbib}

\begin{thebibliography}{10}
\providecommand{\url}[1]{\texttt{#1}}
\providecommand{\urlprefix}{URL }
\providecommand{\doi}[1]{https://doi.org/#1}

\bibitem{abrevaya2019decoupled}
Abrevaya, V.F., Boukhayma, A., Wuhrer, S., Boyer, E.: A decoupled 3d facial
  shape model by adversarial training. In: Proceedings of the IEEE/CVF
  International Conference on Computer Vision. pp. 9419--9428 (2019)

\bibitem{abrevaya2018multilinear}
Abrevaya, V.F., Wuhrer, S., Boyer, E.: Multilinear autoencoder for 3d face
  model learning. In: 2018 IEEE Winter Conference on Applications of Computer
  Vision (WACV). pp.~1--9. IEEE (2018)

\bibitem{amberg2009weight}
Amberg, B., Paysan, P., Vetter, T.: Weight, sex, and facial expressions: On the
  manipulation of attributes in generative 3d face models. In: International
  Symposium on Visual Computing. pp. 875--885. Springer (2009)

\bibitem{bagautdinov2018modeling}
Bagautdinov, T., Wu, C., Saragih, J., Fua, P., Sheikh, Y.: Modeling facial
  geometry using compositional vaes. In: Proceedings of the IEEE Conference on
  Computer Vision and Pattern Recognition. pp. 3877--3886 (2018)

\bibitem{bando2002simple}
Bando, Y., Kuratate, T., Nishita, T.: A simple method for modeling wrinkles on
  human skin. In: 10th Pacific Conference on Computer Graphics and
  Applications, 2002. Proceedings. pp. 166--175. IEEE (2002)

\bibitem{bermano2014facial}
Bermano, A.H., Bradley, D., Beeler, T., Zund, F., Nowrouzezahrai, D., Baran,
  I., Sorkine-Hornung, O., Pfister, H., Sumner, R.W., Bickel, B., et~al.:
  Facial performance enhancement using dynamic shape space analysis. ACM
  Transactions on Graphics (TOG)  \textbf{33}(2),  1--12 (2014)

\bibitem{bickel2008pose}
Bickel, B., Lang, M., Botsch, M., Otaduy, M.A., Gross, M.H.: Pose-space
  animation and transfer of facial details. In: Symposium on Computer
  Animation. pp. 57--66 (2008)

\bibitem{blanz1999morphable}
Blanz, V., Vetter, T.: A morphable model for the synthesis of 3d faces. In:
  Proceedings of the 26th annual conference on Computer graphics and
  interactive techniques. pp. 187--194 (1999)

\bibitem{boissieux2000simulation}
Boissieux, L., Kiss, G., Thalmann, N.M., Kalra, P.: Simulation of skin aging
  and wrinkles with cosmetics insight. In: Computer Animation and Simulation
  2000, pp. 15--27. Springer (2000)

\bibitem{bolkart2015groupwise}
Bolkart, T., Wuhrer, S.: A groupwise multilinear correspondence optimization
  for 3d faces. In: Proceedings of the IEEE international conference on
  computer vision. pp. 3604--3612 (2015)

\bibitem{booth20163d}
Booth, J., Roussos, A., Zafeiriou, S., Ponniah, A., Dunaway, D.: A 3d morphable
  model learnt from 10,000 faces. In: Proceedings of the IEEE Conference on
  Computer Vision and Pattern Recognition. pp. 5543--5552 (2016)

\bibitem{bouritsas2019neural}
Bouritsas, G., Bokhnyak, S., Ploumpis, S., Bronstein, M., Zafeiriou, S.: Neural
  3d morphable models: Spiral convolutional networks for 3d shape
  representation learning and generation. In: Proceedings of the IEEE/CVF
  International Conference on Computer Vision. pp. 7213--7222 (2019)

\bibitem{brunton2014multilinear}
Brunton, A., Bolkart, T., Wuhrer, S.: Multilinear wavelets: A statistical shape
  space for human faces. In: European Conference on Computer Vision. pp.
  297--312. Springer (2014)

\bibitem{cao2015real}
Cao, C., Bradley, D., Zhou, K., Beeler, T.: Real-time high-fidelity facial
  performance capture. ACM Transactions on Graphics (ToG)  \textbf{34}(4),
  ~1--9 (2015)

\bibitem{cao2014displaced}
Cao, C., Hou, Q., Zhou, K.: Displaced dynamic expression regression for
  real-time facial tracking and animation. ACM Transactions on graphics (TOG)
  \textbf{33}(4),  1--10 (2014)

\bibitem{cao2013facewarehouse}
Cao, C., Weng, Y., Zhou, S., Tong, Y., Zhou, K.: Facewarehouse: A 3d facial
  expression database for visual computing. IEEE Transactions on Visualization
  and Computer Graphics  \textbf{20}(3),  413--425 (2013)

\bibitem{chandran2020semantic}
Chandran, P., Bradley, D., Gross, M., Beeler, T.: Semantic deep face models.
  In: 2020 International Conference on 3D Vision (3DV). pp. 345--354. IEEE
  (2020)

\bibitem{chen2019photo}
Chen, A., Chen, Z., Zhang, G., Mitchell, K., Yu, J.: Photo-realistic facial
  details synthesis from single image. In: Proceedings of the IEEE/CVF
  International Conference on Computer Vision. pp. 9429--9439 (2019)

\bibitem{chen2020self}
Chen, Y., Wu, F., Wang, Z., Song, Y., Ling, Y., Bao, L.: Self-supervised
  learning of detailed 3d face reconstruction. IEEE Transactions on Image
  Processing  \textbf{29},  8696--8705 (2020)

\bibitem{cheng2019meshgan}
Cheng, S., Bronstein, M., Zhou, Y., Kotsia, I., Pantic, M., Zafeiriou, S.:
  Meshgan: Non-linear 3d morphable models of faces. arXiv preprint
  arXiv:1903.10384  (2019)

\bibitem{chibane2020ndf}
Chibane, J., Mir, A., Pons-Moll, G.: Neural unsigned distance fields for
  implicit function learning. In: Advances in Neural Information Processing
  Systems ({NeurIPS}) (December 2020)

\bibitem{choi2020stargan}
Choi, Y., Uh, Y., Yoo, J., Ha, J.W.: Stargan v2: Diverse image synthesis for
  multiple domains. In: Proceedings of the IEEE/CVF Conference on Computer
  Vision and Pattern Recognition. pp. 8188--8197 (2020)

\bibitem{deng2021plausible}
Deng, Q., Ma, L., Jin, A., Bi, H., Le, B.H., Deng, Z.: Plausible 3d face
  wrinkle generation using variational autoencoders. IEEE Transactions on
  Visualization \& Computer Graphics (01), ~1--1 (2021)

\bibitem{egger20203d}
Egger, B., Smith, W.A., Tewari, A., Wuhrer, S., Zollhoefer, M., Beeler, T.,
  Bernard, F., Bolkart, T., Kortylewski, A., Romdhani, S., et~al.: 3d morphable
  face models—past, present, and future. ACM Transactions on Graphics (TOG)
  \textbf{39}(5),  1--38 (2020)

\bibitem{feng2021learning}
Feng, Y., Feng, H., Black, M.J., Bolkart, T.: Learning an animatable detailed
  3d face model from in-the-wild images. ACM Transactions on Graphics (TOG)
  \textbf{40}(4),  1--13 (2021)

\bibitem{feng2018evaluation}
Feng, Z.H., Huber, P., Kittler, J., Hancock, P., Wu, X.J., Zhao, Q., Koppen,
  P., R{\"a}tsch, M.: Evaluation of dense 3d reconstruction from 2d face images
  in the wild. In: 2018 13th IEEE International Conference on Automatic Face \&
  Gesture Recognition (FG 2018). pp. 780--786. IEEE (2018)

\bibitem{fyffe2014driving}
Fyffe, G., Jones, A., Alexander, O., Ichikari, R., Debevec, P.: Driving
  high-resolution facial scans with video performance capture. ACM Transactions
  on Graphics (TOG)  \textbf{34}(1),  1--14 (2014)

\bibitem{garrido2013reconstructing}
Garrido, P., Valgaerts, L., Wu, C., Theobalt, C.: Reconstructing detailed
  dynamic face geometry from monocular video. ACM Trans. Graph.
  \textbf{32}(6),  158--1 (2013)

\bibitem{gerig2018morphable}
Gerig, T., Morel-Forster, A., Blumer, C., Egger, B., Luthi, M., Sch{\"o}nborn,
  S., Vetter, T.: Morphable face models-an open framework. In: 2018 13th IEEE
  International Conference on Automatic Face \& Gesture Recognition (FG 2018).
  pp. 75--82. IEEE (2018)

\bibitem{golovinskiy2006statistical}
Golovinskiy, A., Matusik, W., Pfister, H., Rusinkiewicz, S., Funkhouser, T.: A
  statistical model for synthesis of detailed facial geometry. ACM Transactions
  on Graphics (TOG)  \textbf{25}(3),  1025--1034 (2006)

\bibitem{guo2018cnn}
Guo, Y., Cai, J., Jiang, B., Zheng, J., et~al.: Cnn-based real-time dense face
  reconstruction with inverse-rendered photo-realistic face images. IEEE
  transactions on pattern analysis and machine intelligence  \textbf{41}(6),
  1294--1307 (2018)

\bibitem{huynh2018mesoscopic}
Huynh, L., Chen, W., Saito, S., Xing, J., Nagano, K., Jones, A., Debevec, P.,
  Li, H.: Mesoscopic facial geometry inference using deep neural networks. In:
  Proceedings of the IEEE Conference on Computer Vision and Pattern
  Recognition. pp. 8407--8416 (2018)

\bibitem{ichim2017phace}
Ichim, A.E., Kadle{\v{c}}ek, P., Kavan, L., Pauly, M.: Phace: physics-based
  face modeling and animation. ACM Transactions on Graphics (TOG)
  \textbf{36}(4),  1--14 (2017)

\bibitem{igarashi2007appearance}
Igarashi, T., Nishino, K., Nayar, S.K.: The appearance of human skin: A survey.
  Now Publishers Inc (2007)

\bibitem{gansteerability}
Jahanian, A., Chai, L., Isola, P.: On the "steerability" of generative
  adversarial networks. In: International Conference on Learning
  Representations (2020)

\bibitem{jiang20183d}
Jiang, L., Zhang, J., Deng, B., Li, H., Liu, L.: 3d face reconstruction with
  geometry details from a single image. IEEE Transactions on Image Processing
  \textbf{27}(10),  4756--4770 (2018)

\bibitem{jiang2019disentangled}
Jiang, Z.H., Wu, Q., Chen, K., Zhang, J.: Disentangled representation learning
  for 3d face shape. In: Proceedings of the IEEE/CVF Conference on Computer
  Vision and Pattern Recognition. pp. 11957--11966 (2019)

\bibitem{Karras2018ProgressiveGO}
Karras, T., Aila, T., Laine, S., Lehtinen, J.: Progressive growing of gans for
  improved quality, stability, and variation. In: International Conference on
  Learning Representations (2018)

\bibitem{karras2020analyzing}
Karras, T., Laine, S., Aittala, M., Hellsten, J., Lehtinen, J., Aila, T.:
  Analyzing and improving the image quality of stylegan. In: Proceedings of the
  IEEE/CVF Conference on Computer Vision and Pattern Recognition. pp.
  8110--8119 (2020)

\bibitem{kim2015interactive}
Kim, H.J., Oeztireli, A.C., Shin, I.K., Gross, M., Choi, S.M.: Interactive
  generation of realistic facial wrinkles from sketchy drawings. In: Computer
  Graphics Forum. vol.~34, pp. 179--191. Wiley Online Library (2015)

\bibitem{lau2009face}
Lau, M., Chai, J., Xu, Y.Q., Shum, H.Y.: Face poser: Interactive modeling of 3d
  facial expressions using facial priors. ACM Transactions on Graphics (TOG)
  \textbf{29}(1),  1--17 (2009)

\bibitem{lewis2014practice}
Lewis, J.P., Anjyo, K., Rhee, T., Zhang, M., Pighin, F.H., Deng, Z.: Practice
  and theory of blendshape facial models. Eurographics (State of the Art
  Reports)  \textbf{1}(8), ~2 (2014)

\bibitem{li2021semantic}
Li, D., Yang, J., Kreis, K., Torralba, A., Fidler, S.: Semantic segmentation
  with generative models: Semi-supervised learning and strong out-of-domain
  generalization. In: Proceedings of the IEEE/CVF Conference on Computer Vision
  and Pattern Recognition. pp. 8300--8311 (2021)

\bibitem{li2020dynamic}
Li, J., Kuang, Z., Zhao, Y., He, M., Bladin, K., Li, H.: Dynamic facial asset
  and rig generation from a single scan. ACM Trans. Graph.  \textbf{39}(6),
  215--1 (2020)

\bibitem{li2015lightweight}
Li, J., Xu, W., Cheng, Z., Xu, K., Klein, R.: Lightweight wrinkle synthesis for
  3d facial modeling and animation. Computer-Aided Design  \textbf{58},
  117--122 (2015)

\bibitem{li2007modeling}
Li, M., Yin, B., Kong, D., Luo, X.: Modeling expressive wrinkles of face for
  animation. In: Fourth International Conference on Image and Graphics (ICIG
  2007). pp. 874--879. IEEE (2007)

\bibitem{li2020learning}
Li, R., Bladin, K., Zhao, Y., Chinara, C., Ingraham, O., Xiang, P., Ren, X.,
  Prasad, P., Kishore, B., Xing, J., et~al.: Learning formation of
  physically-based face attributes. In: Proceedings of the IEEE/CVF Conference
  on Computer Vision and Pattern Recognition. pp. 3410--3419 (2020)

\bibitem{li2017learning}
Li, T., Bolkart, T., Black, M.J., Li, H., Romero, J.: Learning a model of
  facial shape and expression from 4d scans. ACM Transactions on Graphics
  \textbf{36}(6) (2017)

\bibitem{li2007wrinkle}
Li, Y.b., Xiao, H., Zhang, S.y.: The wrinkle generation method for facial
  reconstruction based on extraction of partition wrinkle line features and
  fractal interpolation. In: Fourth International Conference on Image and
  Graphics (ICIG 2007). pp. 933--937. IEEE (2007)

\bibitem{li2018feature}
Li, Y., Ma, L., Fan, H., Mitchell, K.: Feature-preserving detailed 3d face
  reconstruction from a single image. In: Proceedings of the 15th ACM SIGGRAPH
  European Conference on Visual Media Production. pp.~1--9 (2018)

\bibitem{li2019linestofacephoto}
Li, Y., Chen, X., Wu, F., Zha, Z.J.: Linestofacephoto: Face photo generation
  from lines with conditional self-attention generative adversarial networks.
  In: Proceedings of the 27th ACM International Conference on Multimedia. pp.
  2323--2331 (2019)

\bibitem{liu2019few}
Liu, M.Y., Huang, X., Mallya, A., Karras, T., Aila, T., Lehtinen, J., Kautz,
  J.: Few-shot unsupervised image-to-image translation. In: Proceedings of the
  IEEE/CVF International Conference on Computer Vision. pp. 10551--10560 (2019)

\bibitem{ma2019real}
Ma, L., Deng, Z.: Real-time hierarchical facial performance capture. In:
  Proceedings of the ACM SIGGRAPH Symposium on Interactive 3D Graphics and
  Games. pp. 1--10 (2019)

\bibitem{mescheder2018training}
Mescheder, L., Geiger, A., Nowozin, S.: Which training methods for gans do
  actually converge? In: International conference on machine learning. pp.
  3481--3490. PMLR (2018)

\bibitem{neumann2013sparse}
Neumann, T., Varanasi, K., Wenger, S., Wacker, M., Magnor, M., Theobalt, C.:
  Sparse localized deformation components. ACM Transactions on Graphics (TOG)
  \textbf{32}(6),  1--10 (2013)

\bibitem{Park_2019_CVPR}
Park, J.J., Florence, P., Straub, J., Newcombe, R., Lovegrove, S.: Deepsdf:
  Learning continuous signed distance functions for shape representation. In:
  The IEEE Conference on Computer Vision and Pattern Recognition (CVPR) (June
  2019)

\bibitem{park2020swapping}
Park, T., Zhu, J.Y., Wang, O., Lu, J., Shechtman, E., Efros, A., Zhang, R.:
  Swapping autoencoder for deep image manipulation. Advances in Neural
  Information Processing Systems  \textbf{33},  7198--7211 (2020)

\bibitem{parke1974parametric}
Parke, F.I.: A parametric model for human faces. The University of Utah (1974)

\bibitem{paysan2010statistical}
Paysan, P.: Statistical modeling of facial aging based on 3D scans. Ph.D.
  thesis, University\_of\_Basel (2010)

\bibitem{radlanski2012face}
Radlanski, R.J., Wesker, K.: The face: pictorial atlas of clinical anatomy.
  Quintessence publishing (2012)

\bibitem{ranjan2018generating}
Ranjan, A., Bolkart, T., Sanyal, S., Black, M.J.: Generating 3d faces using
  convolutional mesh autoencoders. In: Proceedings of the European Conference
  on Computer Vision (ECCV). pp. 704--720 (2018)

\bibitem{richardson2017learning}
Richardson, E., Sela, M., Or-El, R., Kimmel, R.: Learning detailed face
  reconstruction from a single image. In: Proceedings of the IEEE conference on
  computer vision and pattern recognition. pp. 1259--1268 (2017)

\bibitem{sanyal2019learning}
Sanyal, S., Bolkart, T., Feng, H., Black, M.J.: Learning to regress 3d face
  shape and expression from an image without 3d supervision. In: Proceedings of
  the IEEE/CVF Conference on Computer Vision and Pattern Recognition. pp.
  7763--7772 (2019)

\bibitem{sela2017unrestricted}
Sela, M., Richardson, E., Kimmel, R.: Unrestricted facial geometry
  reconstruction using image-to-image translation. In: Proceedings of the IEEE
  International Conference on Computer Vision. pp. 1576--1585 (2017)

\bibitem{sengupta2018sfsnet}
Sengupta, S., Kanazawa, A., Castillo, C.D., Jacobs, D.W.: Sfsnet: Learning
  shape, reflectance and illuminance of facesin the wild'. In: Proceedings of
  the IEEE conference on computer vision and pattern recognition. pp.
  6296--6305 (2018)

\bibitem{serup2006handbook}
Serup, J., Jemec, G.B., Grove, G.L.: Handbook of non-invasive methods and the
  skin. CRC press (2006)

\bibitem{shamai2019synthesizing}
Shamai, G., Slossberg, R., Kimmel, R.: Synthesizing facial photometries and
  corresponding geometries using generative adversarial networks. ACM
  Transactions on Multimedia Computing, Communications, and Applications (TOMM)
   \textbf{15}(3s),  1--24 (2019)

\bibitem{shi2014automatic}
Shi, F., Wu, H.T., Tong, X., Chai, J.: Automatic acquisition of high-fidelity
  facial performances using monocular videos. ACM Transactions on Graphics
  (TOG)  \textbf{33}(6),  1--13 (2014)

\bibitem{shin2014extraction}
Shin, I.K., {\"O}ztireli, A.C., Kim, H.J., Beeler, T., Gross, M., Choi, S.M.:
  Extraction and transfer of facial expression wrinkles for facial performance
  enhancement. In: PG (Short Papers) (2014)

\bibitem{simo2016learning}
Simo-Serra, E., Iizuka, S., Sasaki, K., Ishikawa, H.: Learning to simplify:
  fully convolutional networks for rough sketch cleanup. ACM Transactions on
  Graphics (TOG)  \textbf{35}(4),  1--11 (2016)

\bibitem{slossberg2018high}
Slossberg, R., Shamai, G., Kimmel, R.: High quality facial surface and texture
  synthesis via generative adversarial networks. In: Proceedings of the
  European Conference on Computer Vision (ECCV) Workshops. pp.~0--0 (2018)

\bibitem{suwajanakorn2014total}
Suwajanakorn, S., Kemelmacher-Shlizerman, I., Seitz, S.M.: Total moving face
  reconstruction. In: European conference on computer vision. pp. 796--812.
  Springer (2014)

\bibitem{taubin95}
Taubin, G.: A signal processing approach to fair surface design. In:
  Proceedings of the 22nd Annual Conference on Computer Graphics and
  Interactive Techniques. p. 351–358. SIGGRAPH '95, Association for Computing
  Machinery, New York, NY, USA (1995). \doi{10.1145/218380.218473},
  \url{https://doi.org/10.1145/218380.218473}

\bibitem{tran2019towards}
Tran, L., Liu, F., Liu, X.: Towards high-fidelity nonlinear 3d face morphable
  model. In: Proceedings of the IEEE/CVF Conference on Computer Vision and
  Pattern Recognition. pp. 1126--1135 (2019)

\bibitem{vlasic2006face}
Vlasic, D., Brand, M., Pfister, H., Popovic, J.: Face transfer with multilinear
  models. In: ACM SIGGRAPH 2006 Courses, pp. 24--es (2006)

\bibitem{wang2018high}
Wang, T.C., Liu, M.Y., Zhu, J.Y., Tao, A., Kautz, J., Catanzaro, B.:
  High-resolution image synthesis and semantic manipulation with conditional
  gans. In: Proceedings of the IEEE conference on computer vision and pattern
  recognition. pp. 8798--8807 (2018)

\bibitem{wang2020single}
Wang, Z., Yu, X., Lu, M., Wang, Q., Qian, C., Xu, F.: Single image portrait
  relighting via explicit multiple reflectance channel modeling. ACM
  Transactions on Graphics (TOG)  \textbf{39}(6),  1--13 (2020)

\bibitem{tsdf}
Werner, D., Al-Hamadi, A., Werner, P.: Truncated signed distance function:
  Experiments on voxel size. In: Campilho, A., Kamel, M. (eds.) Image Analysis
  and Recognition. pp. 357--364. Springer International Publishing, Cham (2014)

\bibitem{wu2016anatomically}
Wu, C., Bradley, D., Gross, M., Beeler, T.: An anatomically-constrained local
  deformation model for monocular face capture. ACM transactions on graphics
  (TOG)  \textbf{35}(4),  1--12 (2016)

\bibitem{wu1999simulating}
Wu, Y., Kalra, P., Moccozet, L., Magnenat-Thalmann, N.: Simulating wrinkles and
  skin aging. The visual computer  \textbf{15}(4),  183--198 (1999)

\bibitem{xu2014controllable}
Xu, F., Chai, J., Liu, Y., Tong, X.: Controllable high-fidelity facial
  performance transfer. ACM Transactions on Graphics (TOG)  \textbf{33}(4),
  1--11 (2014)

\bibitem{yang2020facescape}
Yang, H., Zhu, H., Wang, Y., Huang, M., Shen, Q., Yang, R., Cao, X.: Facescape:
  a large-scale high quality 3d face dataset and detailed riggable 3d face
  prediction. In: Proceedings of the IEEE/CVF Conference on Computer Vision and
  Pattern Recognition. pp. 601--610 (2020)

\bibitem{yenamandra2021i3dmm}
Yenamandra, T., Tewari, A., Bernard, F., Seidel, H.P., Elgharib, M., Cremers,
  D., Theobalt, C.: i3dmm: Deep implicit 3d morphable model of human heads. In:
  Proceedings of the IEEE/CVF Conference on Computer Vision and Pattern
  Recognition. pp. 12803--12813 (2021)

\bibitem{zeng2019df2net}
Zeng, X., Peng, X., Qiao, Y.: Df2net: A dense-fine-finer network for detailed
  3d face reconstruction. In: Proceedings of the IEEE/CVF International
  Conference on Computer Vision. pp. 2315--2324 (2019)

\bibitem{zhang2018unreasonable}
Zhang, R., Isola, P., Efros, A.A., Shechtman, E., Wang, O.: The unreasonable
  effectiveness of deep features as a perceptual metric. In: Proceedings of the
  IEEE conference on computer vision and pattern recognition. pp. 586--595
  (2018)

\bibitem{zhang2021datasetgan}
Zhang, Y., Ling, H., Gao, J., Yin, K., Lafleche, J.F., Barriuso, A., Torralba,
  A., Fidler, S.: Datasetgan: Efficient labeled data factory with minimal human
  effort. In: Proceedings of the IEEE/CVF Conference on Computer Vision and
  Pattern Recognition. pp. 10145--10155 (2021)

\bibitem{zhuang2020enjoy}
Zhuang, P., Koyejo, O.O., Schwing, A.: Enjoy your editing: Controllable gans
  for image editing via latent space navigation. In: International Conference
  on Learning Representations (2020)

\end{thebibliography}

\appendix

\section{Overview}

In this supplementary material we present:
\begin{itemize}
  \item Effectiveness of LPIPS\cite{zhang2018unreasonable} as a displacement map metric
  \item Quantitative comparison using LPIPS
  \item User study
  \item Additional discussion on wrinkle line editing
  \item More qualitative results
  \item Additional qualitative comparison on in-the-wild images
  \item Examples of extracted detail line maps and distance fields
\end{itemize}

Please also refer to our video for animation results.

\section{Effectiveness of LPIPS\cite{zhang2018unreasonable} as a displacement map metric}

\begin{figure*}[t]
  \centering
  \begin{minipage}[t]{.48\textwidth}
    \centering
    \includeinkscape[width=\linewidth]{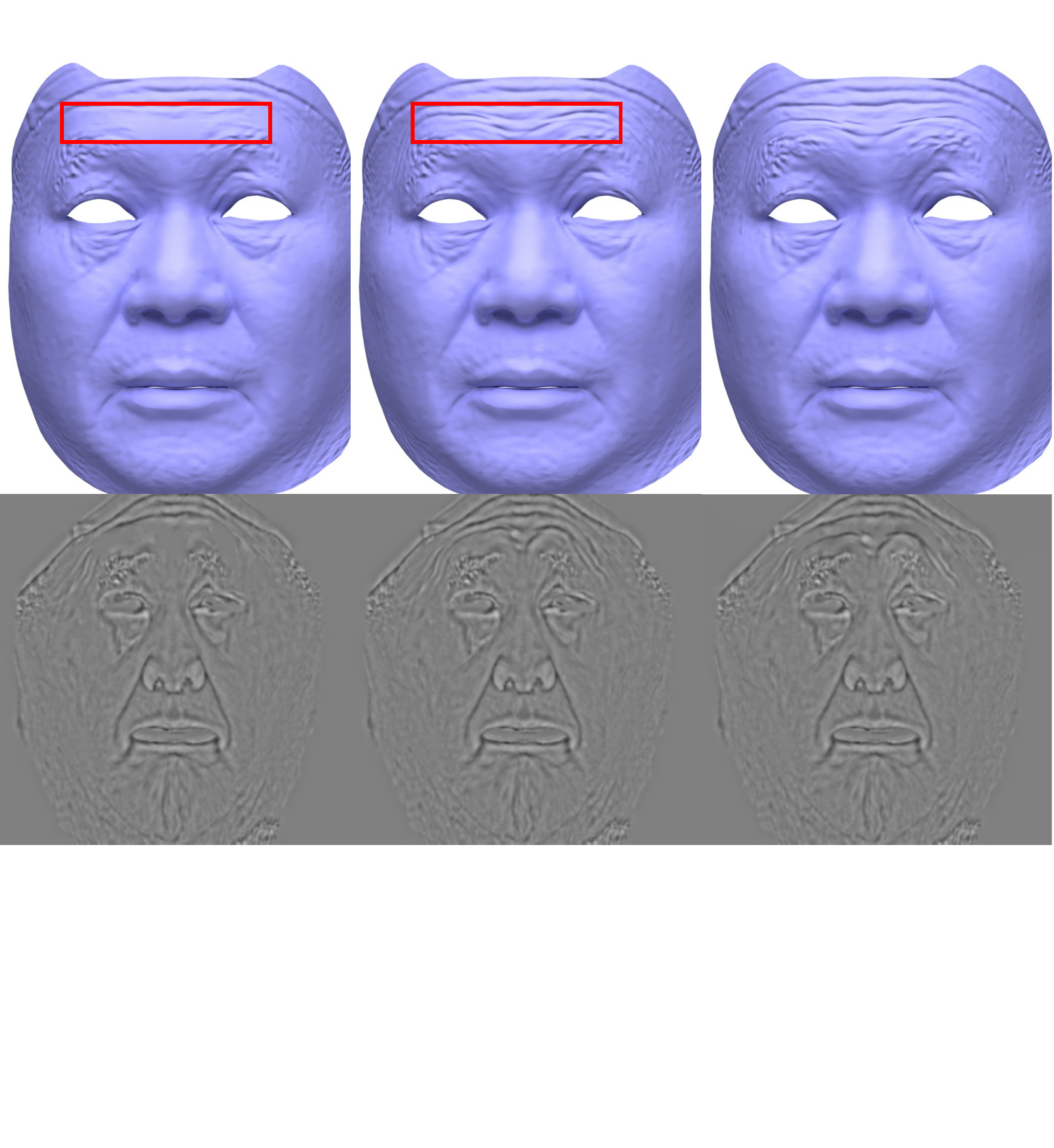}
  \end{minipage}
  \hspace{.02\textwidth}
  \begin{minipage}[t]{.48\textwidth}
    \centering
    \includeinkscape[width=\linewidth]{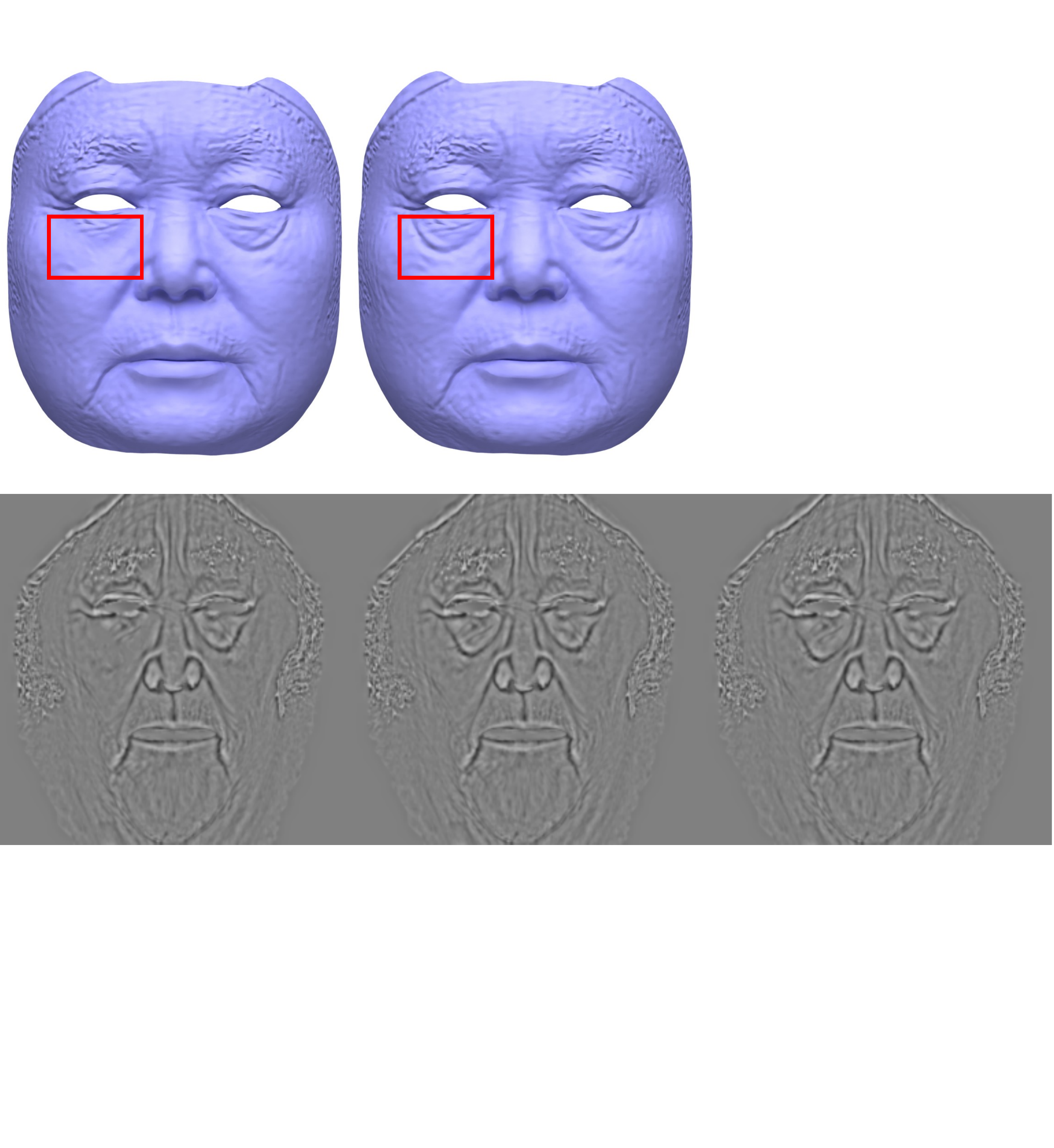}
  \end{minipage}
  \caption{LPIPS agrees with humans on the similarity of facial details, while L1 Loss disagrees with humans.}
  \label{fig:lpips_results}
\end{figure*}

We use LPIPS\cite{zhang2018unreasonable} to evaluate the similarity between displacement maps, because it is more consistent with the visual similarity perceived by humans, compared to traditional metrics such as L1 Loss.
To investigate this, we select a wrinkle on a displacement map, delete it or move it, and generate two modified displacement maps, which we refer to as ``absent wrinkles'' and ``misaligned wrinkles'', respectively.
We evaluate the L1 Loss and LPIPS between the modified displacement maps and the ground truth.
When evaluating LPIPS, we normalize the displacement value to range $[-1, 1]$ and convert grayscale to RGB.
We also invite 23 participants and ask them which is more similar to the ground truth.
The results are shown in Fig. \ref{fig:lpips_results}.
We find that LPIPS considers misaligned wrinkles to be closer to the ground truth, which agrees with human judgements, while L1 Loss disagrees with humans.
Therefore, we believe LPIPS is more suitable to measure displacement map similarity.

\section{Quantitative comparison using LPIPS}

We use LPIPS to quantitatively compare the displacement maps generated by our method, FaceScape\cite{yang2020facescape} and DECA\cite{feng2021learning}.
The comparison is performed on randomly selected 618 test samples from the dataset from \cite{yang2020facescape}, which has samples of the same person with different expressions.
For each sample, all the methods (ours, FaceScape and DECA) first obtain the detail representation from an input image.
Then, original details are generated from the representation and used to evaluate the reconstruction error in LPIPS.
The detail representation is also combined with target expression parameters to generate displacement maps with other target expressions.
We evaluate the editing error between the generated and reference displacement maps in LPIPS.
Because DECA's mesh topology is different from the dataset from \cite{yang2020facescape}, we perform non-rigid registration and then extract the displacement maps in the way described in our paper.
FaceScape is known to work better with neutral expression inputs, so we separately report the errors using neutral expression inputs and using inputs with non-neutral expressions.
The results are shown in Table \ref{tab:comparison_all}.

Our model achieves the lowest error both in reconstruction and editing.
The generated details of DECA are visually plausible, but their quantitative errors are higher than ours, possibly because their method is only trained on 2D image data.
In DECA's paper, a similar phenomenon is reported that after adding details, their mesh reconstruction error increases.
The results indicate that our model is able to both more accurately represent input details and generate dynamic details better matching the input person's identity, with either neutral or non-neutral input expressions.

\begin{table}[t]
  \centering
  \caption{Quantitative comparison with FaceScape and DECA.}
  \begin{tabular}{c c c c}
    \toprule
    &Ours & FaceScape & DECA\\
    \midrule
    neutral recon & {\bf 0.1447} & 0.1784 & 0.4023 \\
    neutral edit & {\bf 0.1719} & 0.1991 & 0.4020 \\
    non-neutral recon & {\bf 0.1511} & 0.1991 & 0.4039 \\
    non-neutral edit &  {\bf 0.1759} & 0.1980 & 0.4021 \\
    \bottomrule
  \end{tabular}
  \label{tab:comparison_all}
\end{table}

\section{User study}

\begin{figure}[t]
  \centering
  \includegraphics[width=\linewidth]{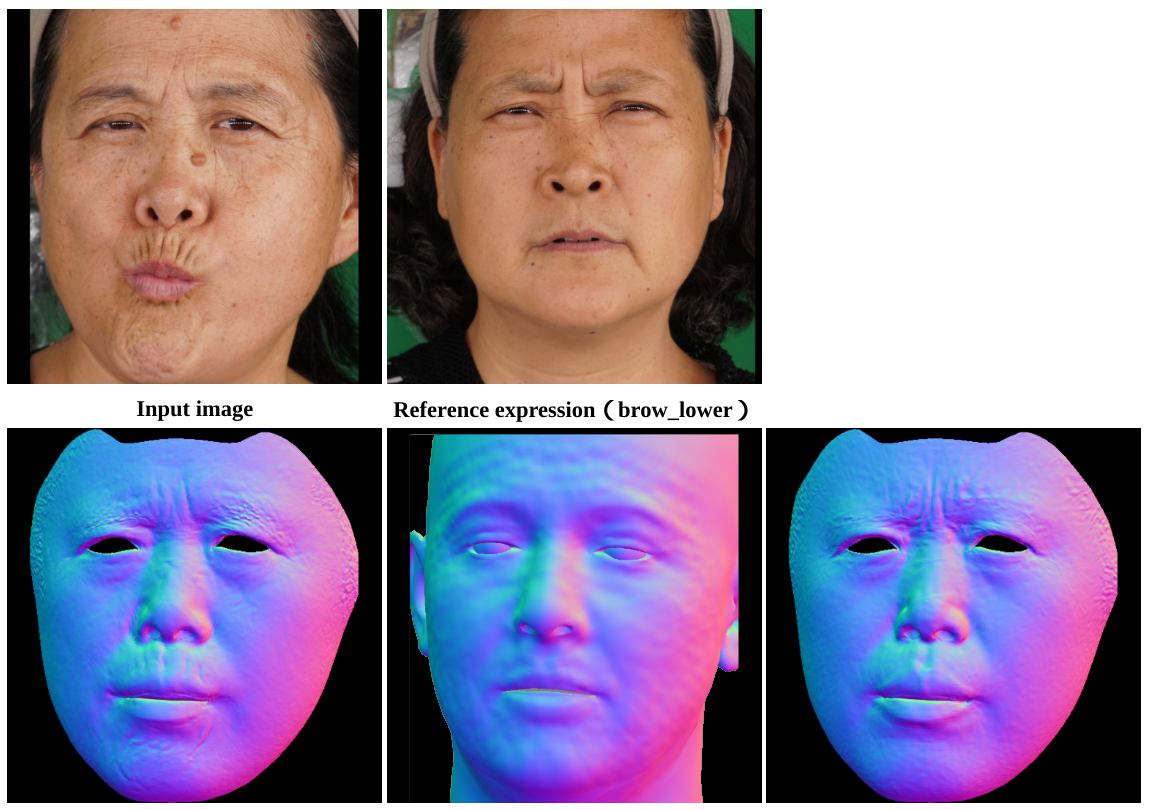}
  \caption{A user study example. We use normal map rendering for each method.}
  \label{fig:user_study}
\end{figure}

To compare with FaceScape and DECA, we conduct a user study to measure: (1) how well each method preserves the input identity, (2) how well it conveys the target expression the user wants to change to, and (3) the overall generation quality.
First, we generated 297 samples from the dataset from \cite{wang2020single}, each containing an input image, a reference image with a different expression, and the editing results generated by different methods. 
In generating these samples, while we use DECA's original renderer to better visualize their results in the qualitative study, we used a normal map rendering shown in Fig. \ref{fig:user_study} for all the methods to render details without bias.
The results are also randomly shuffled.
Then, 20 randomly selected samples were provided to each participant, and they rated in the three aspects mentioned above from 1 to 5 (higher is better). In total, we collected 282 valid responses from 15 participants.
The average ratings are shown in Table \ref{tab:user_study}. Notice that FaceScape treats the dynamic details as static ones and wrongly keeps them for other expressions. Since all the details are kept, it may lead to ``better'' identity preservation as a side effect. Our method is better than FaceScape in conveying target expressions and is considered the best in overall quality.

\begin{table}[t]
  \centering
  \caption{User study results vs. FaceScape\cite{yang2020facescape} and DECA\cite{feng2021learning}. 
  FaceScape is expectedly better at preserving identity, at the cost of not animating details to convey target expression. 
  Our method is considered the best in overall quality.}
  \begin{tabular}{c c c c}
    \toprule
& Ours & FaceScape & DECA \\
    \midrule
    {\small Same identity} & 3.33 & {\bf 3.37} & 1.37 \\
    {\small Convey target expression} &  {\bf 3.40} & 3.01 & 1.87 \\
    {\small Overall better} & {\bf 3.44} & 3.18 & 1.36 \\
    \bottomrule
  \end{tabular}
  \label{tab:user_study}
\end{table}

\section{Additional discussion on wrinkle line editing}

\begin{figure*}[t]
  \centering
  \includeinkscape[width=\linewidth]{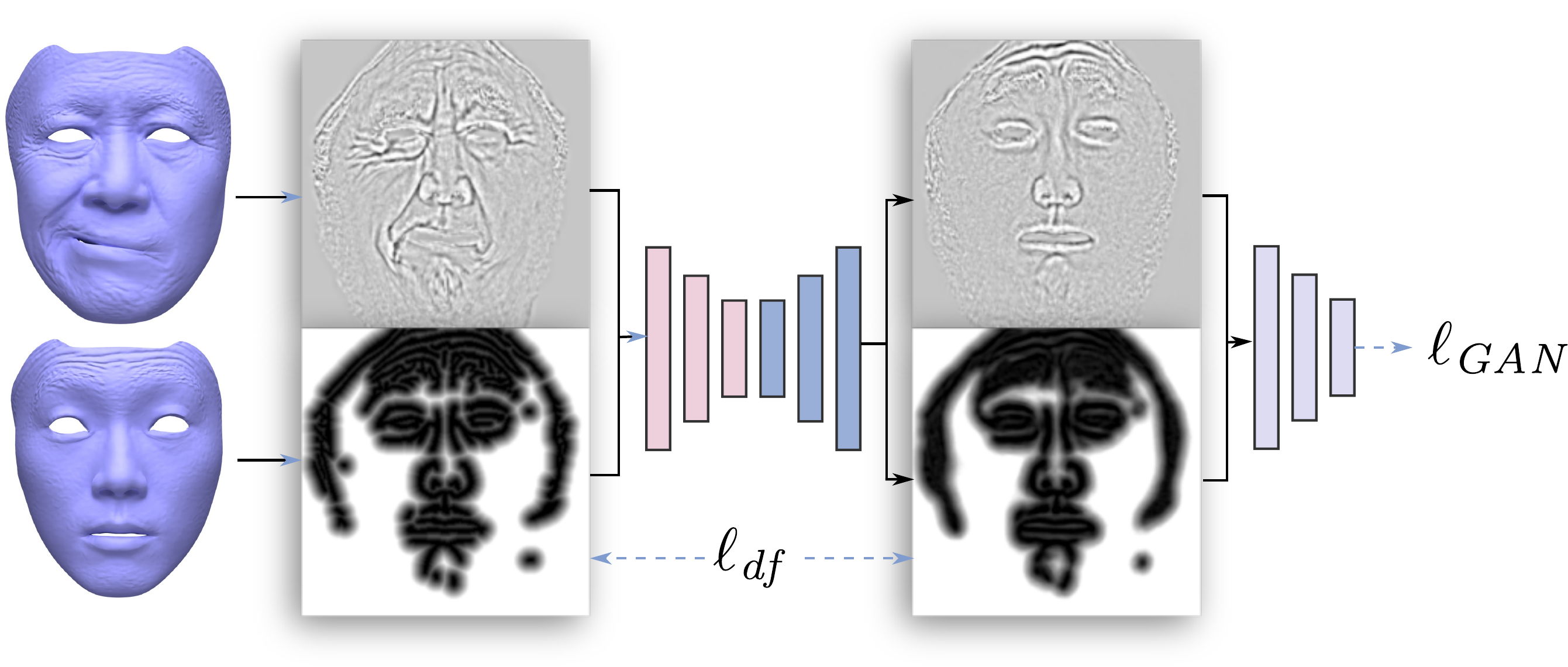}
  \caption{Training pipeline for wrinkle line editing.}
  \label{fig:struct_edit}
\end{figure*}

The key to achieving wrinkle line editing is using mismatched distance field and displacement map in training, as shown in Fig. \ref{fig:struct_edit}.
Specifically, the input displacement map represents the original details.
As the distance field is from another random face, it is used to mimic the user editing.
$\ell_{\mathit{df}}$ supervises the output distance field to be consistent with the input, keeping the output wrinkle structure consistent with the edits.
$\ell_{\mathit{GAN}}$ keeps the output displacement map consistent with the output distance field, thus translating the distance field to the final details represented by the displacement map.

\section{More qualitative results}

\begin{figure}[h]
\begin{subfigure}{\linewidth}
  \centering
  \begin{minipage}[t]{.495\linewidth}
    \centering
    \includeinkscape[width=\linewidth]{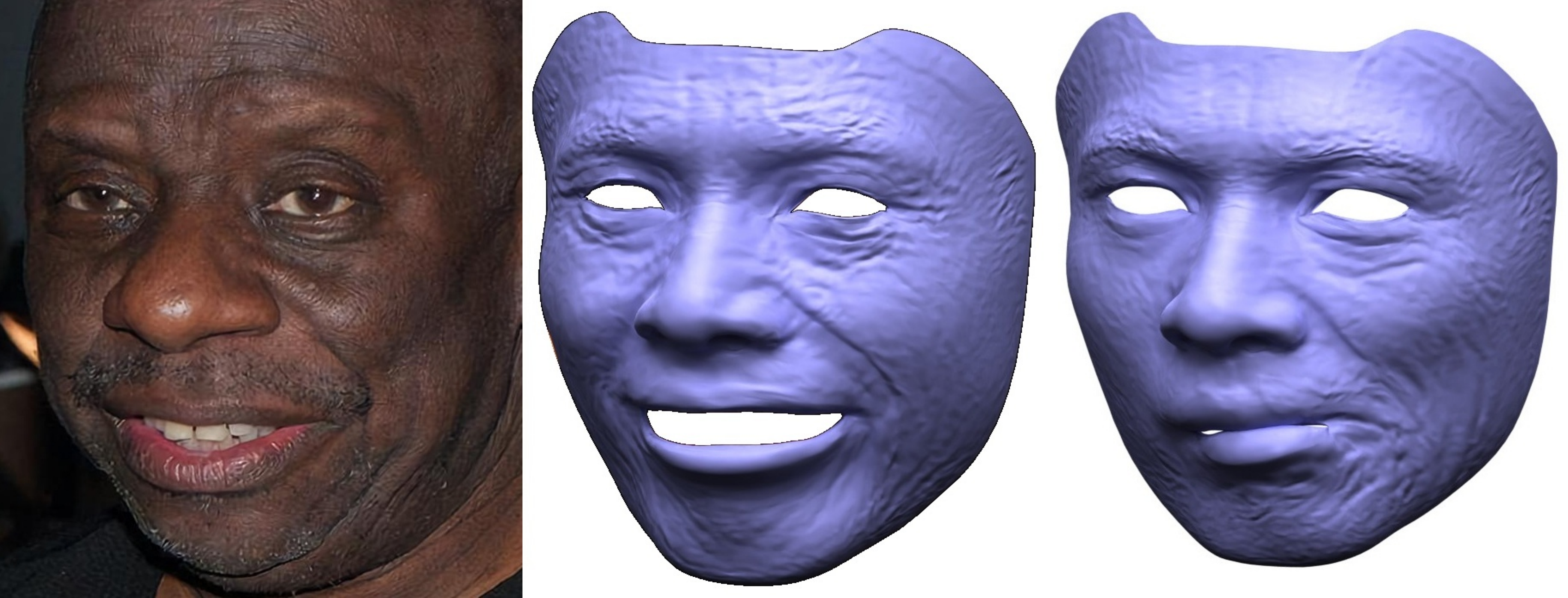}
    \includeinkscape[width=\linewidth]{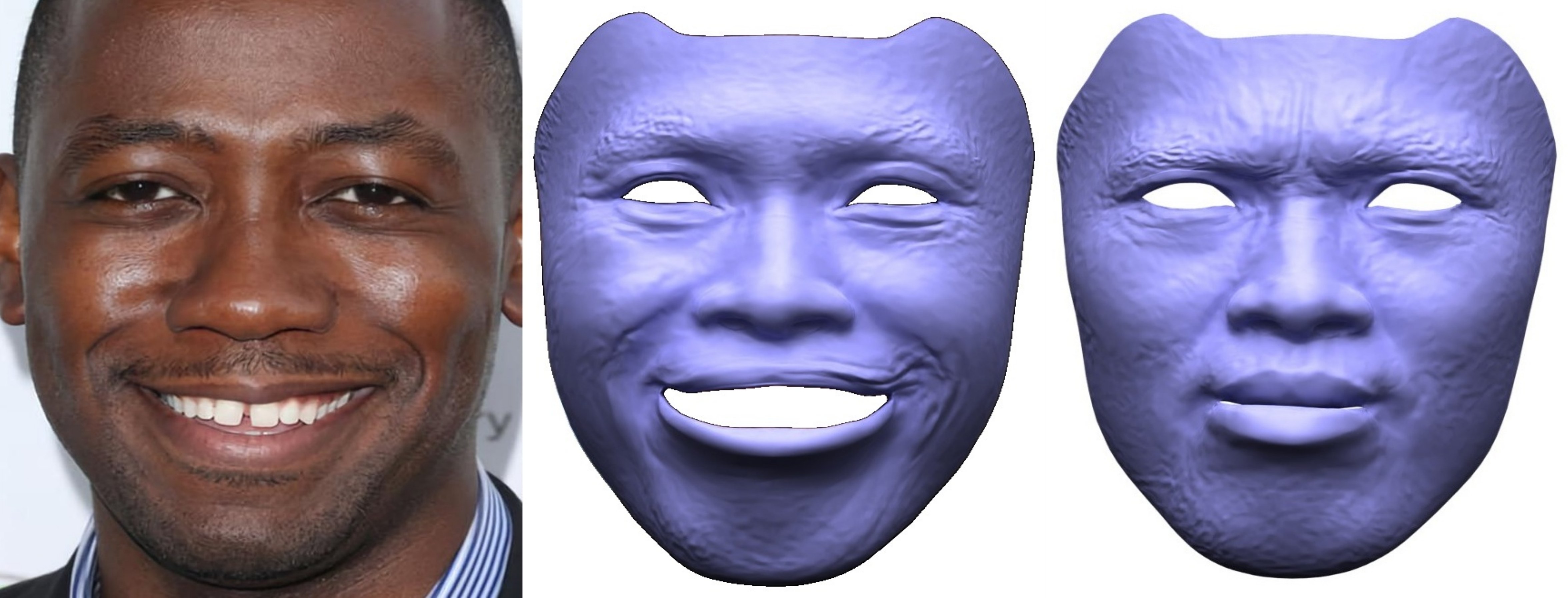}
    \includeinkscape[width=\linewidth]{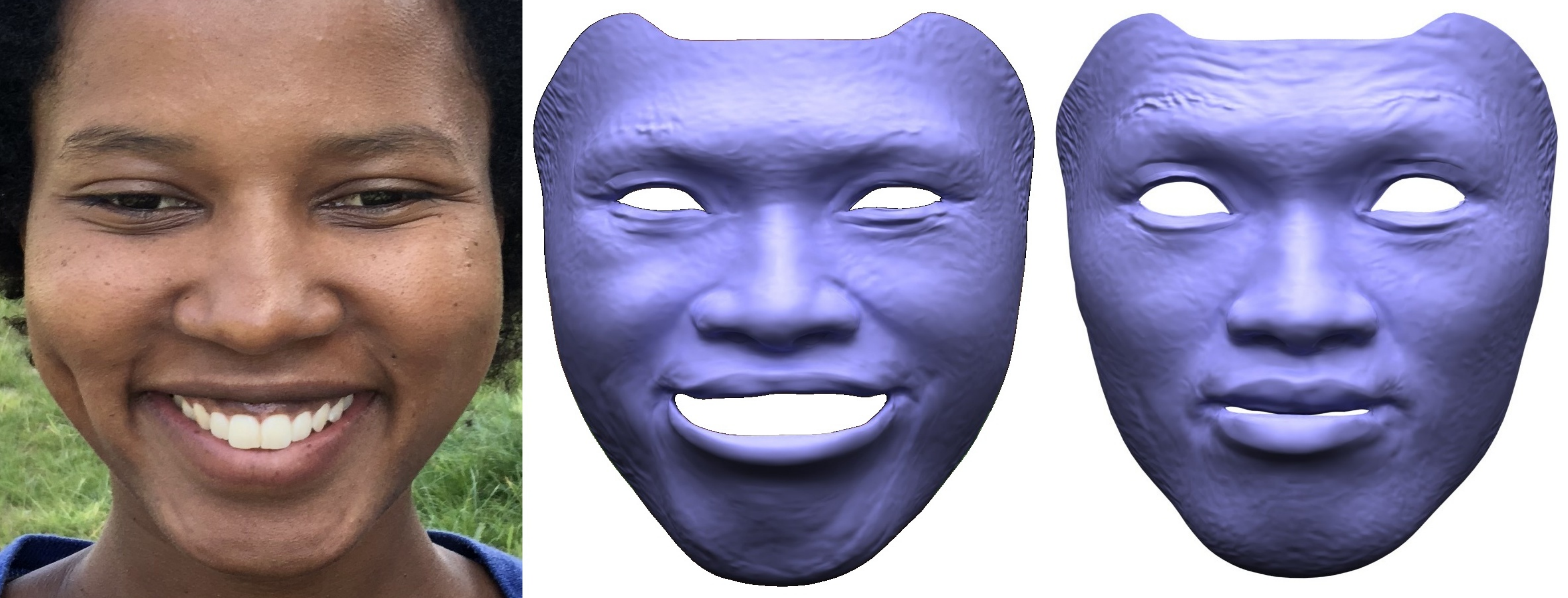}

  \end{minipage}
  \begin{minipage}[t]{.495\linewidth}
    \centering
    \includeinkscape[width=\linewidth]{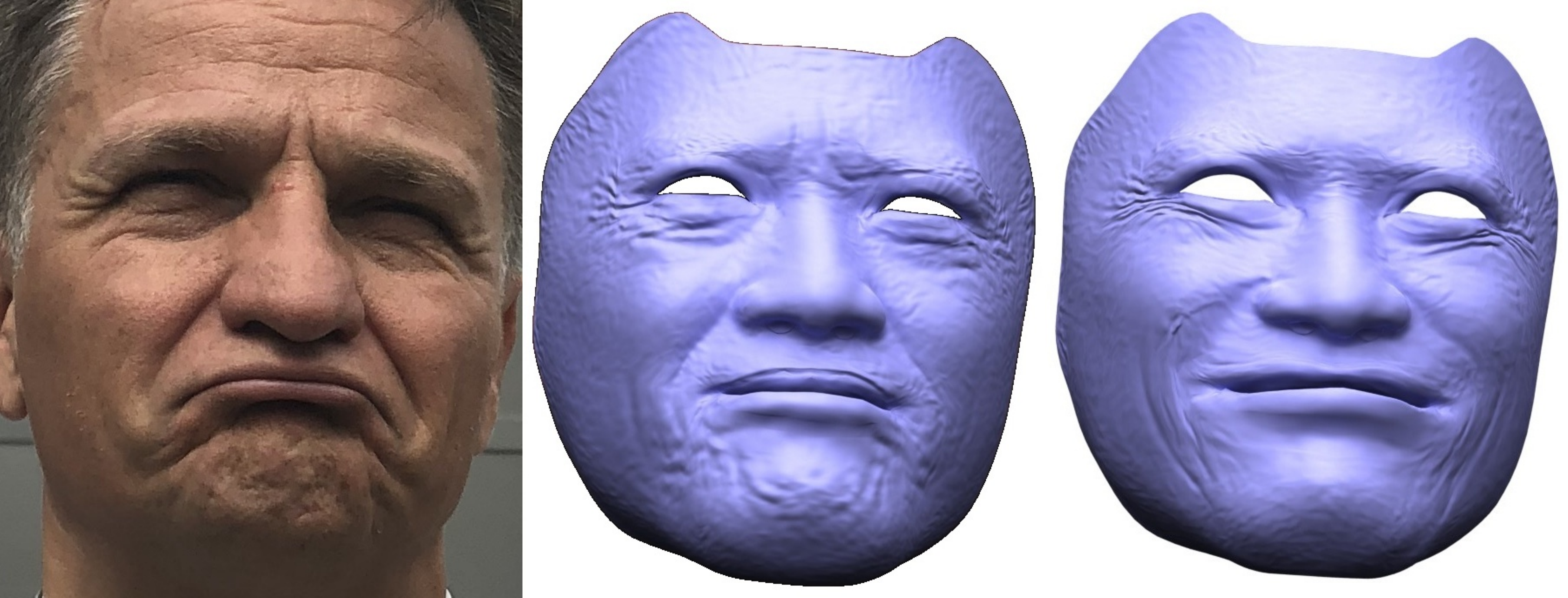}
    \includeinkscape[width=\linewidth]{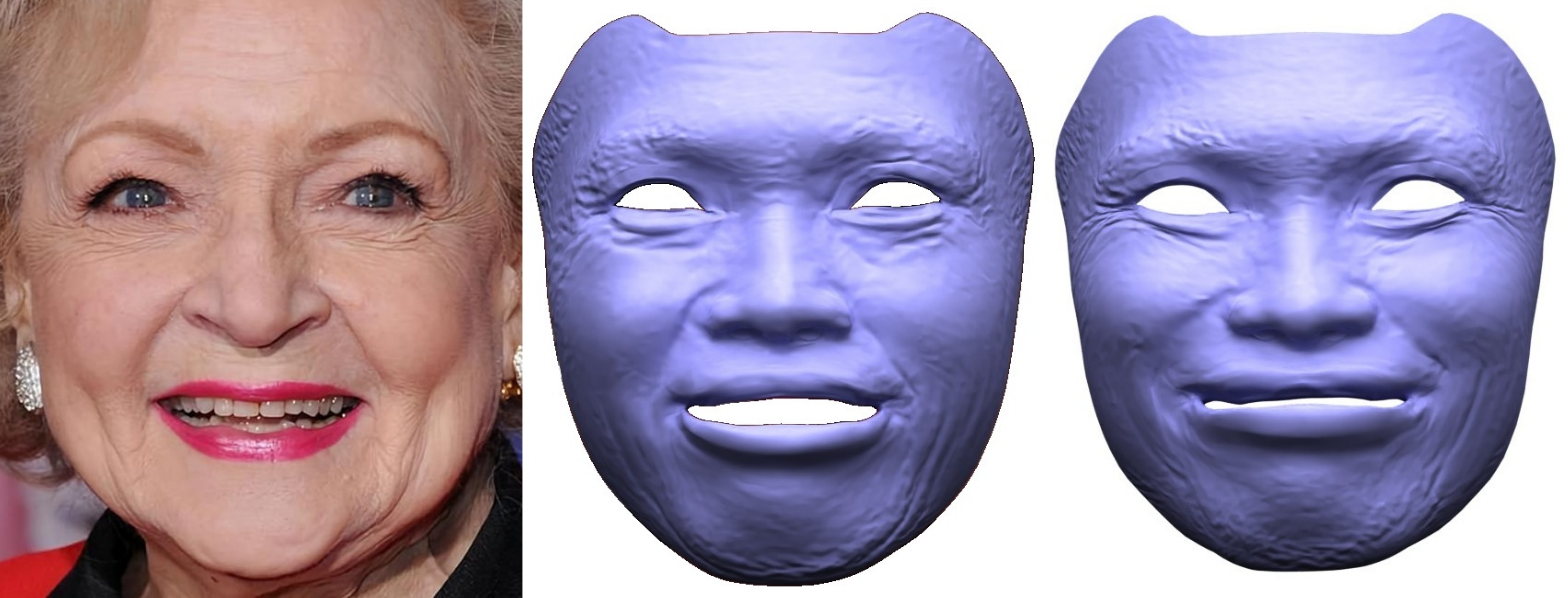}
    \includeinkscape[width=\linewidth]{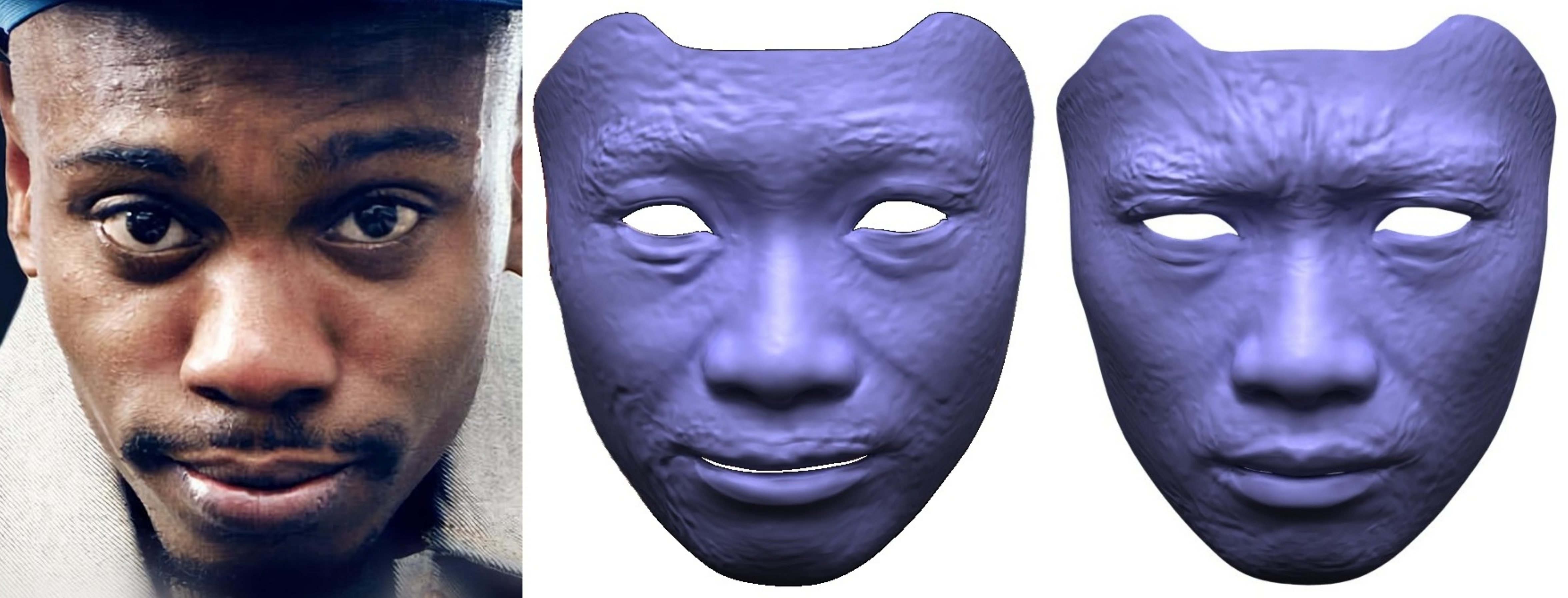}

  \end{minipage}
  \caption{More diverse skin tones.}
  \end{subfigure}
\begin{subfigure}{\linewidth}
  \centering
  \begin{minipage}[t]{.495\linewidth}
    \centering
    \includeinkscape[width=\linewidth]{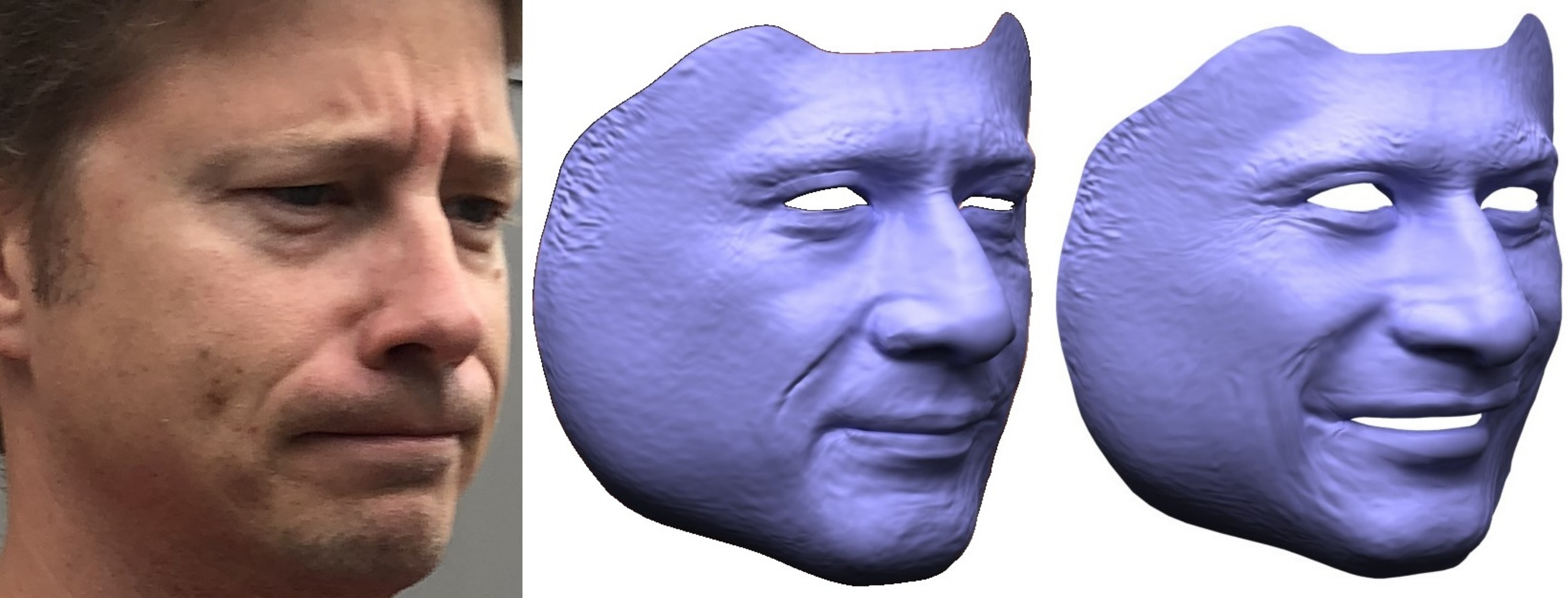}
    \includeinkscape[width=\linewidth]{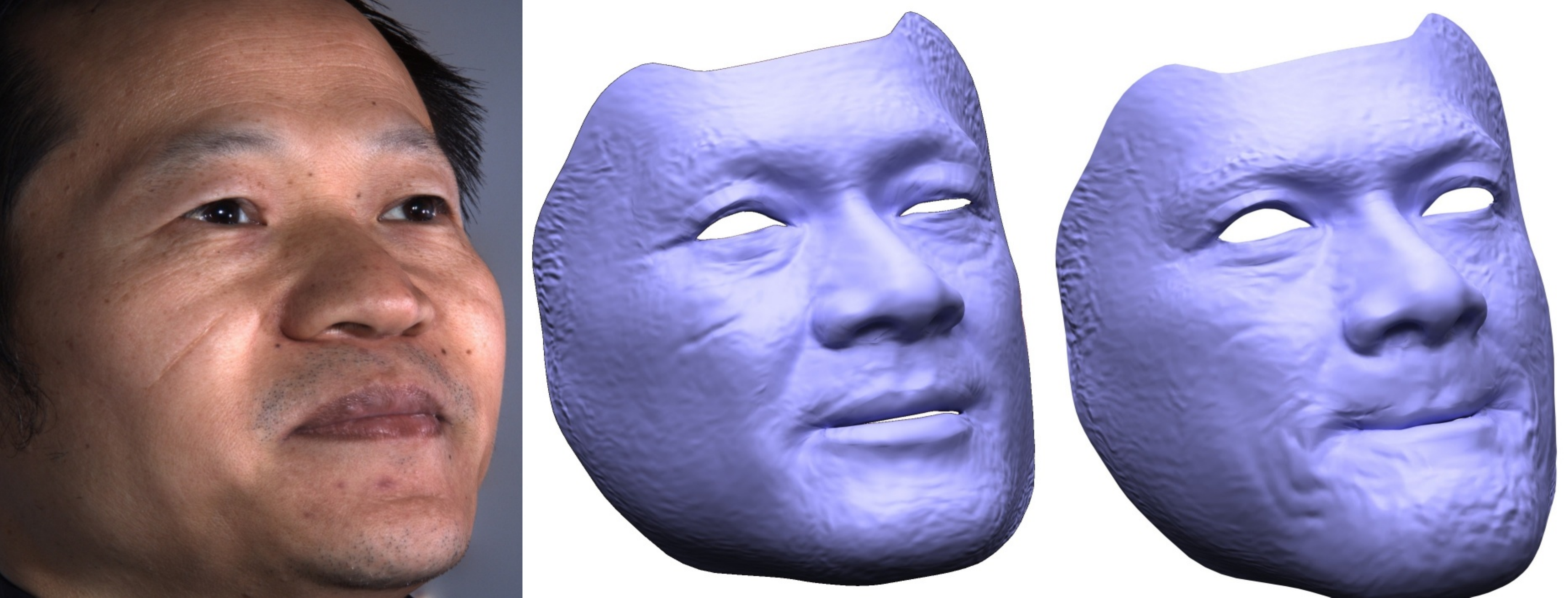}
    \includeinkscape[width=\linewidth]{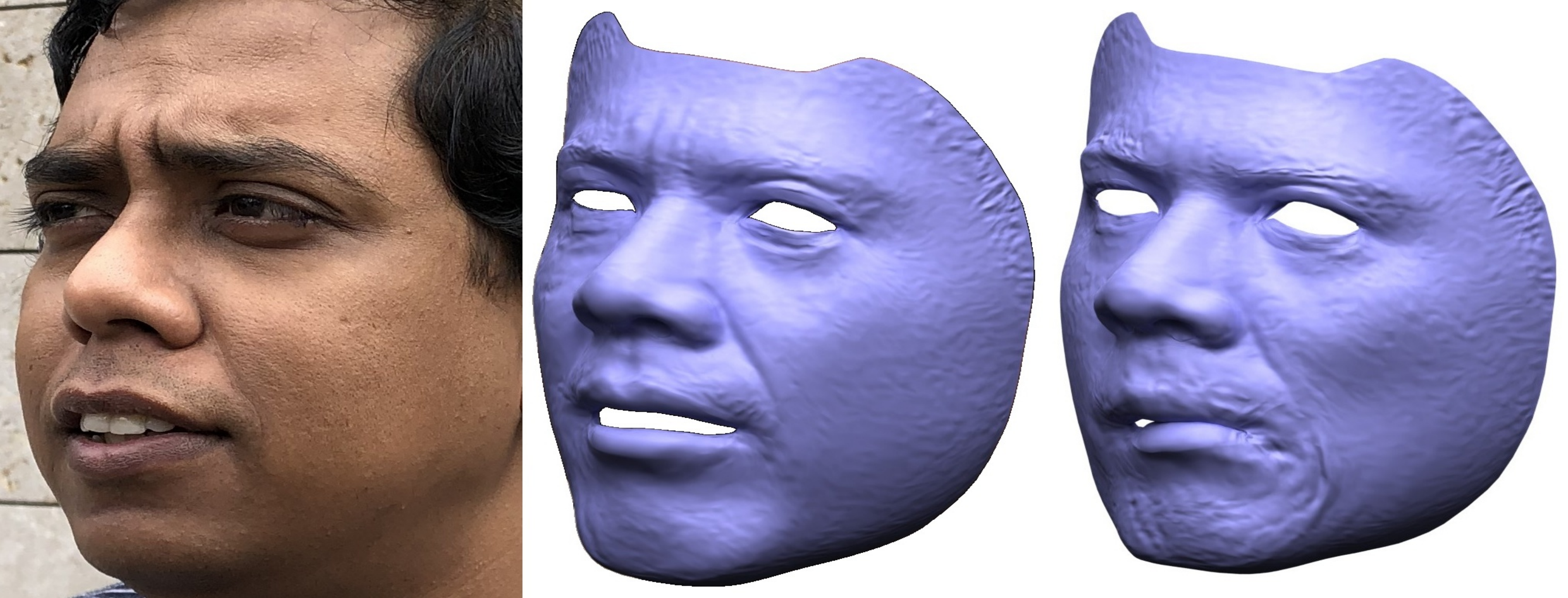}

  \end{minipage}
  \begin{minipage}[t]{.495\linewidth}
    \centering
    \includeinkscape[width=\linewidth]{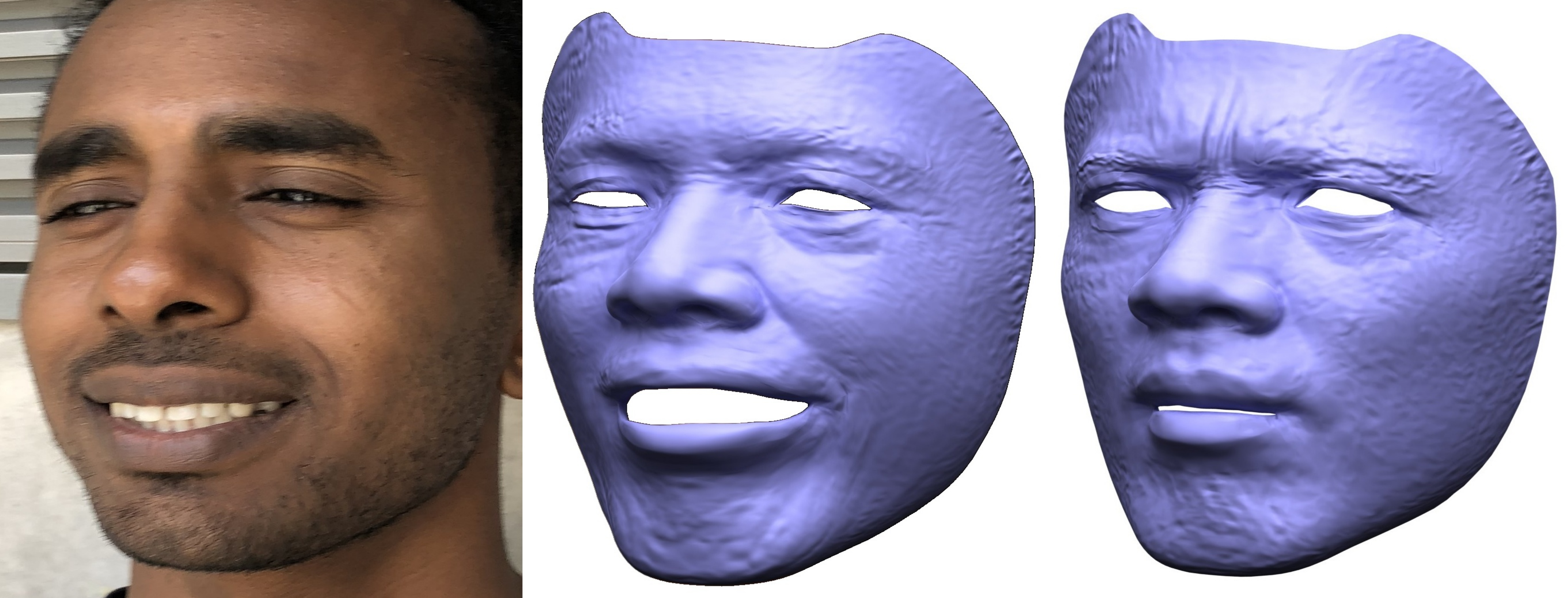}
    \includeinkscape[width=\linewidth]{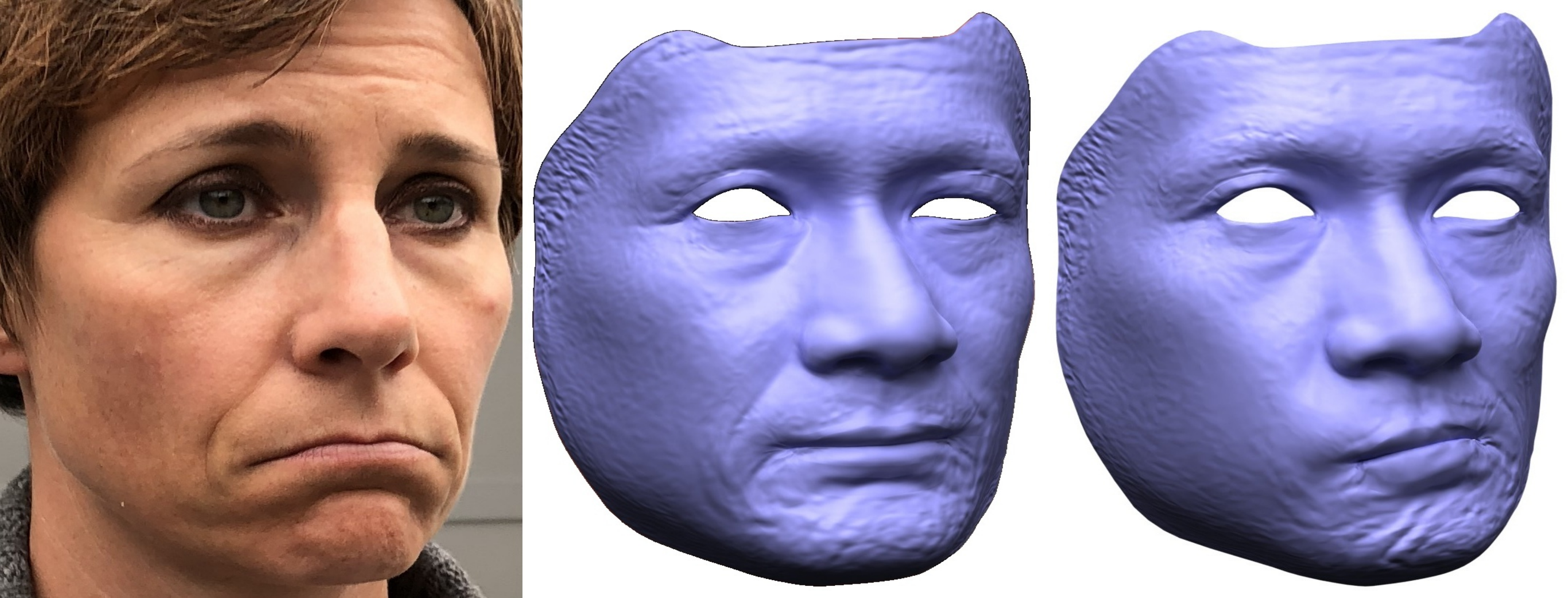}
    \includeinkscape[width=\linewidth]{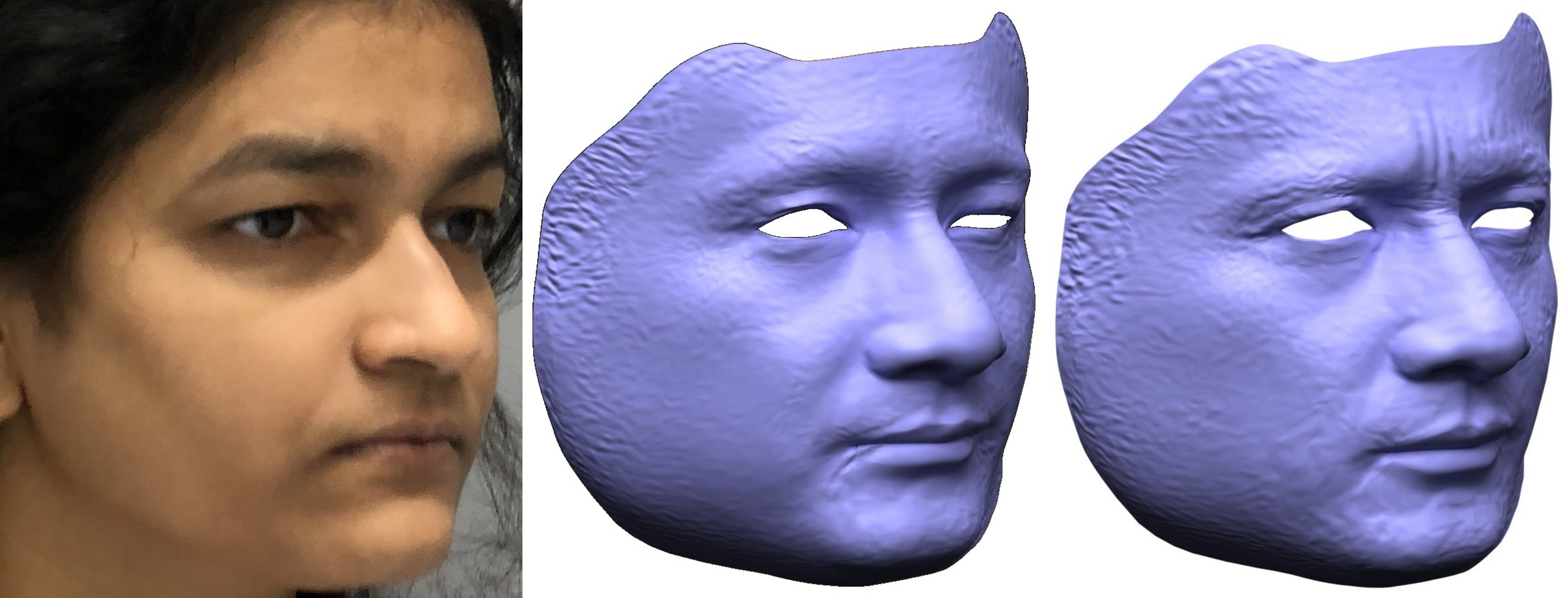}

  \end{minipage}
  \caption{More diverse head poses.}
  \end{subfigure}
  \caption{Input images, our reconstruction and editing results.}
  \label{fig:additional_res}
\end{figure}

Here we present more extreme results from Feng et al.\cite{sanyal2019learning}, NoW\cite{feng2018evaluation}, and CelebA-HQ\cite{Karras2018ProgressiveGO} datasets in Fig. \ref{fig:additional_res}, where more varieties in skin tones and head poses are well handled.
Note that we cannot handle extreme profile poses as the used large-scale 3DMM fitting fails, which is beyond the scope of this paper on detail modeling.

\section{Additional qualitative comparison on in-the-wild images}

\begin{figure*}[ht]
  \centering
  \includeinkscape[width=\linewidth]{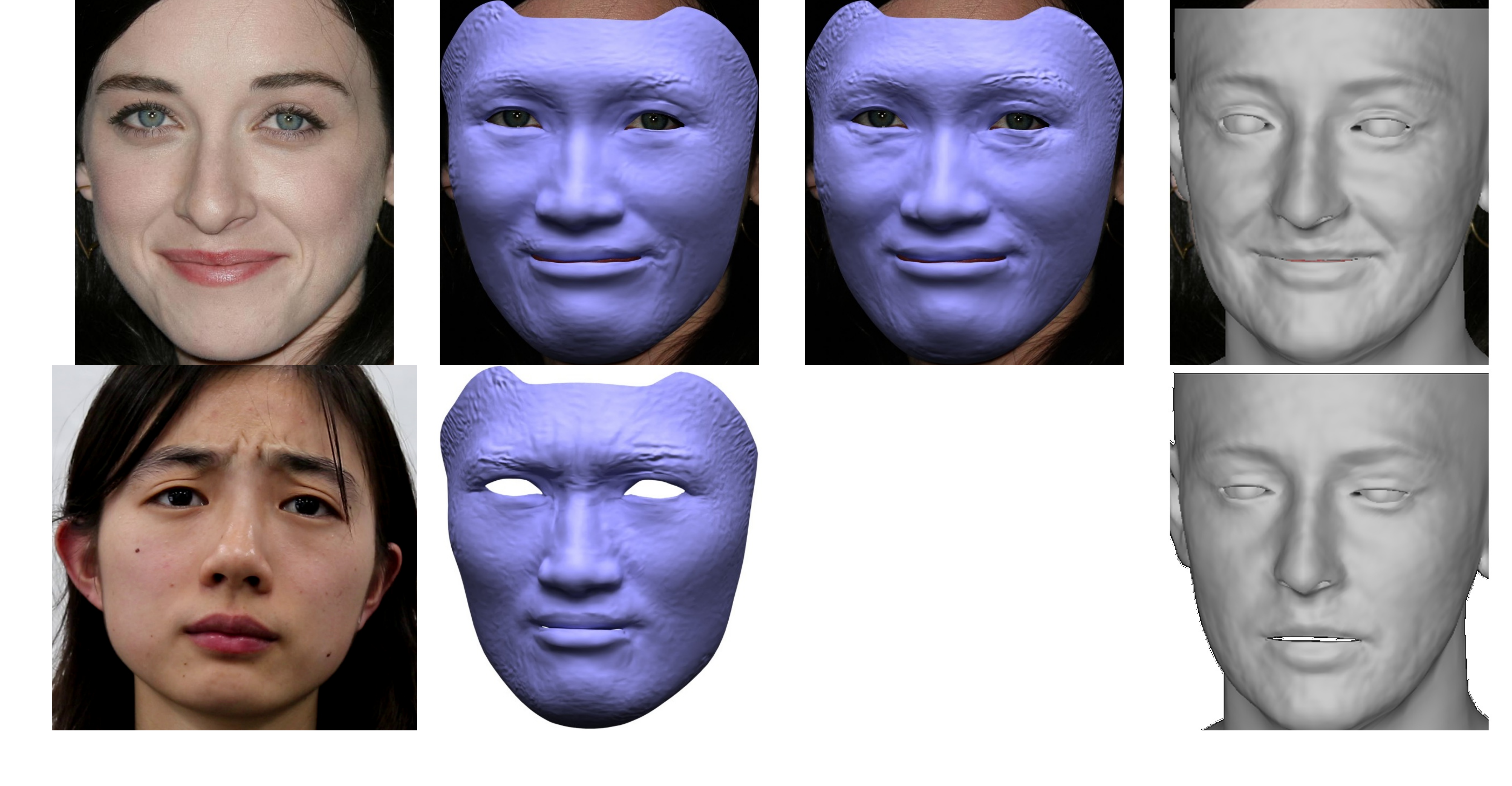}
  \includeinkscape[width=\linewidth]{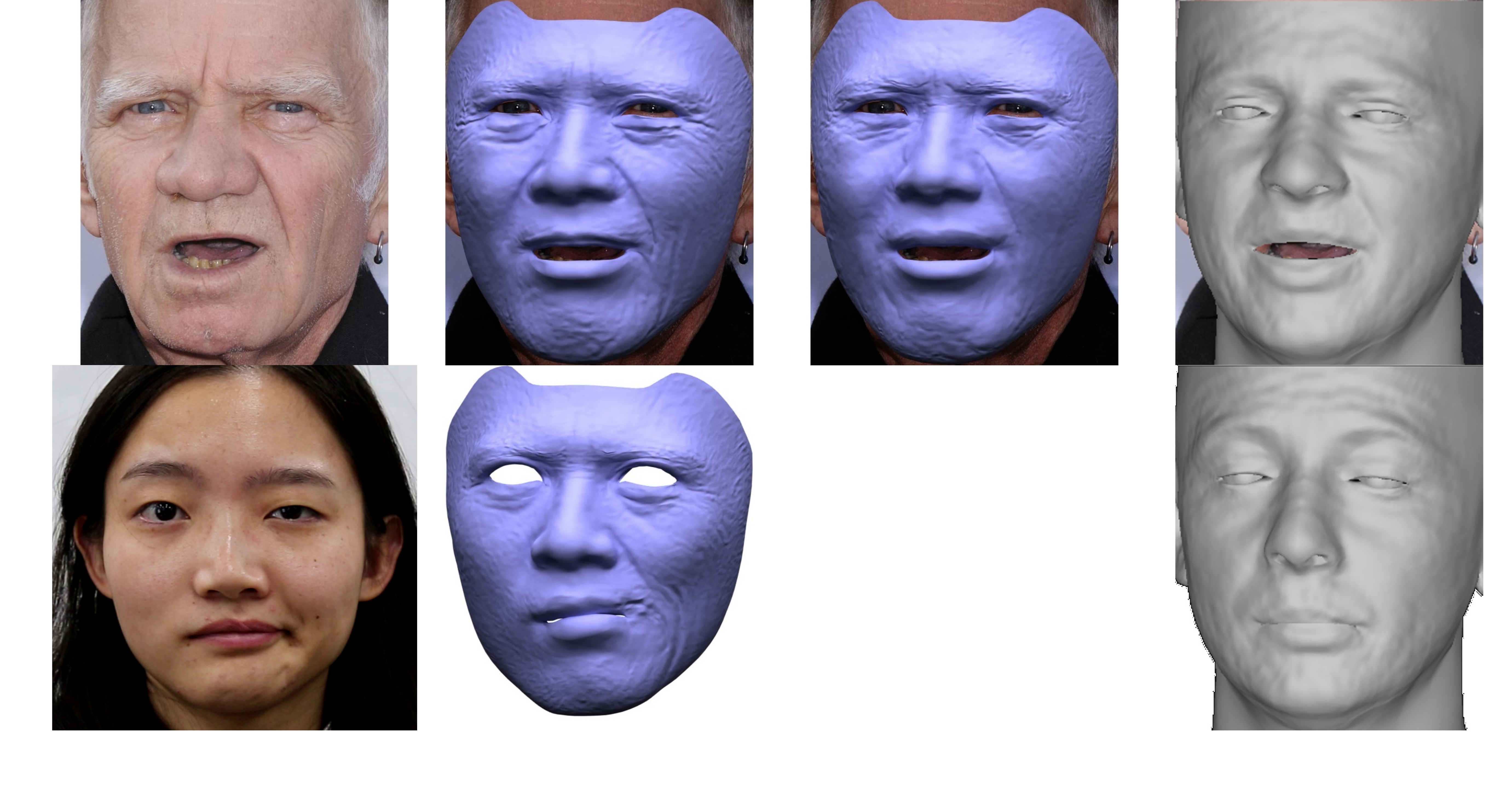}
  \caption{Qualitative results on CelebA-HQ.}
  \label{fig:celeba_results1}
\end{figure*}

\begin{figure*}[ht]
  \centering
  \includeinkscape[width=\linewidth]{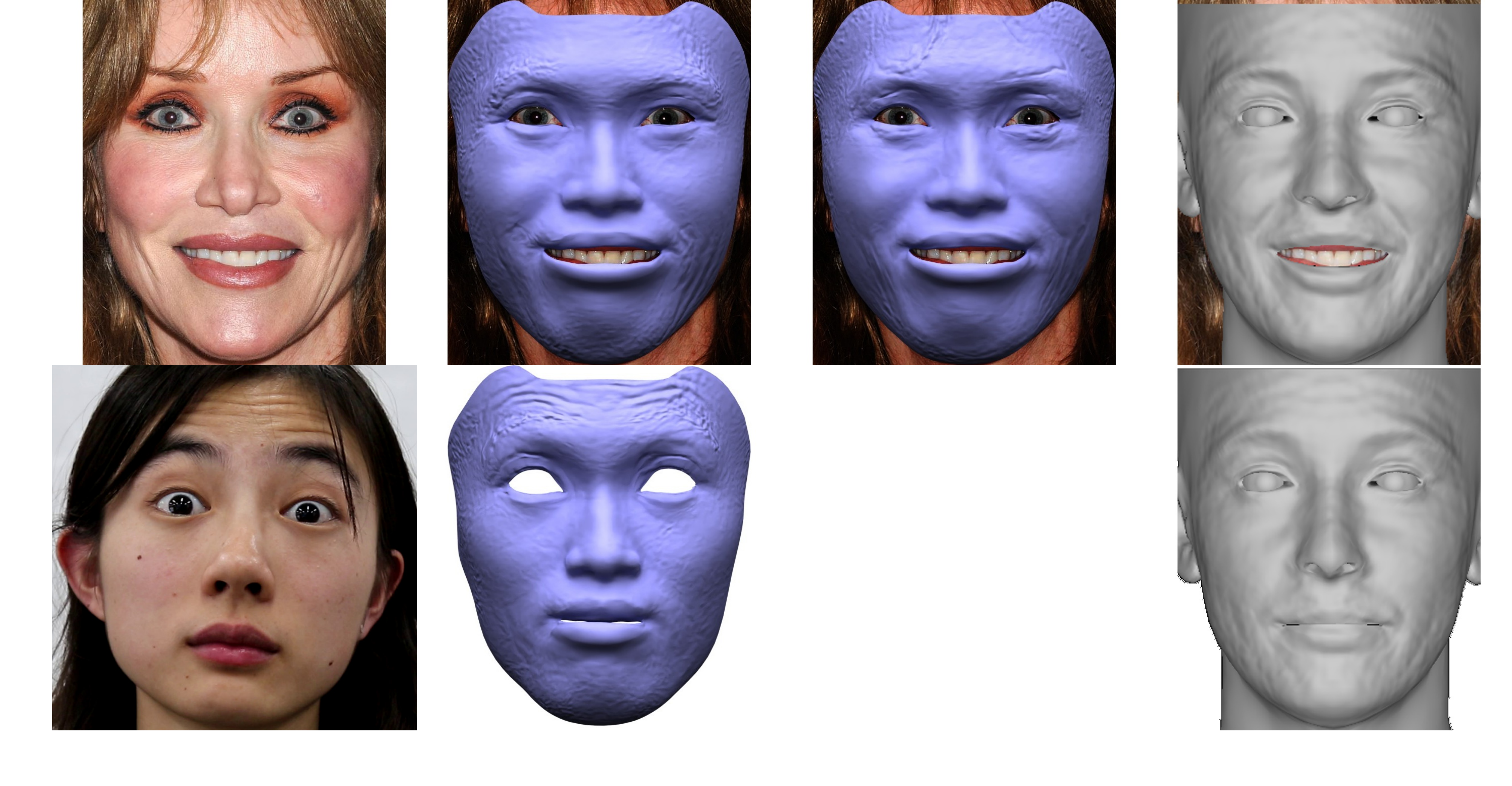}
  \includeinkscape[width=\linewidth]{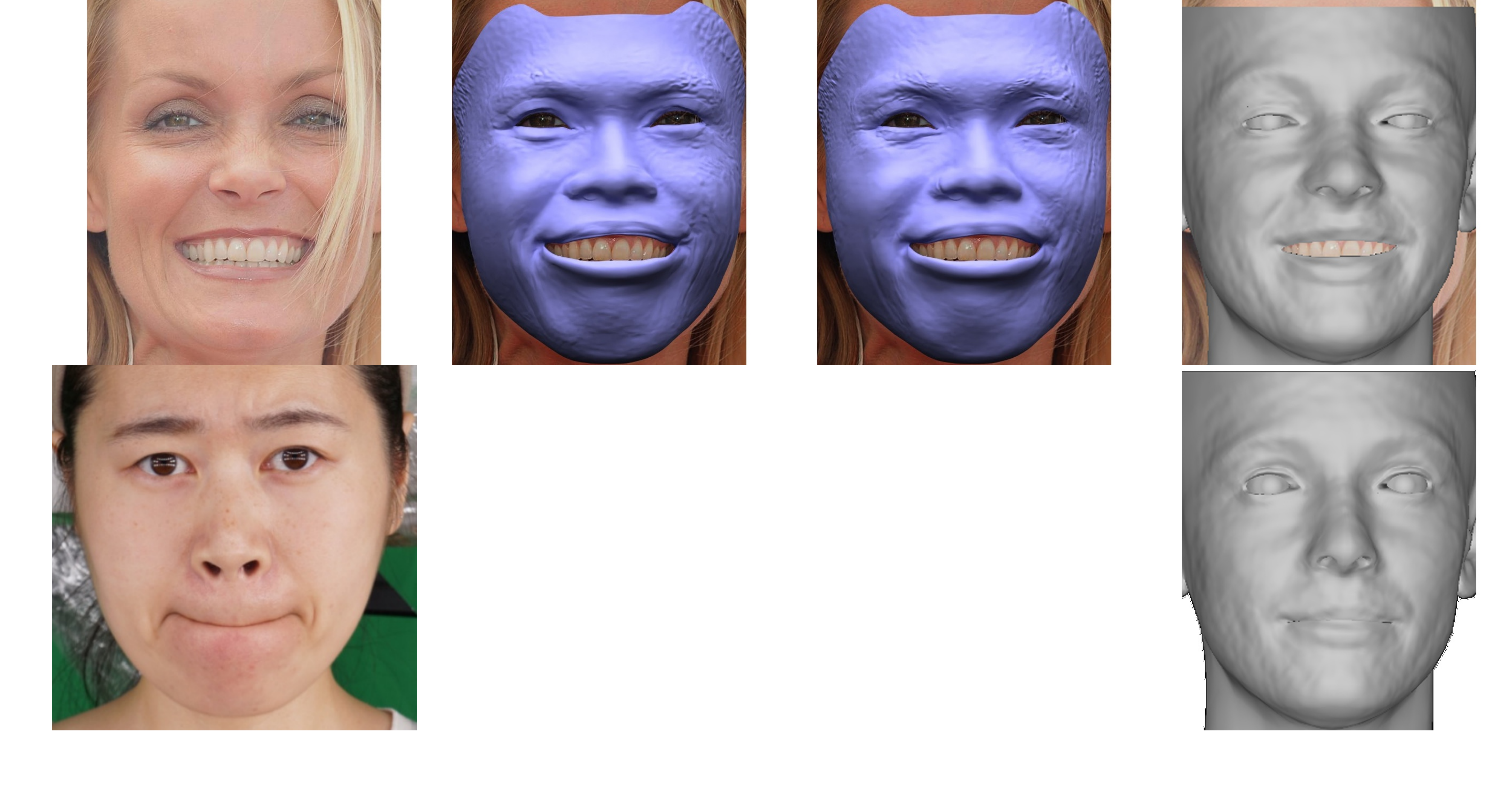}
  \caption{Qualitative results on CelebA-HQ.}
  \label{fig:celeba_results3}
\end{figure*}

\begin{figure*}[ht]
  \centering
  \includeinkscape[width=\linewidth]{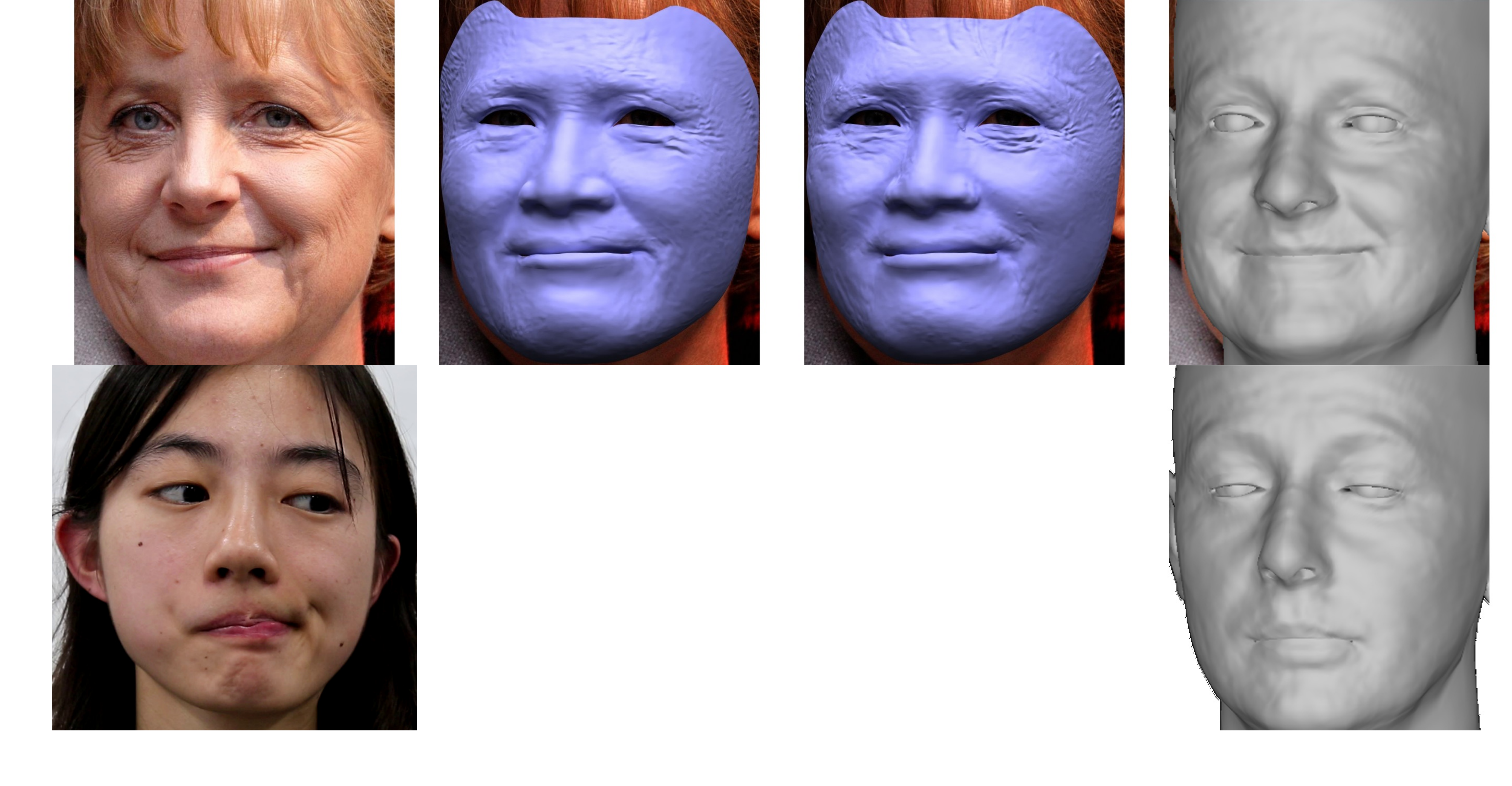}
  \includeinkscape[width=\linewidth]{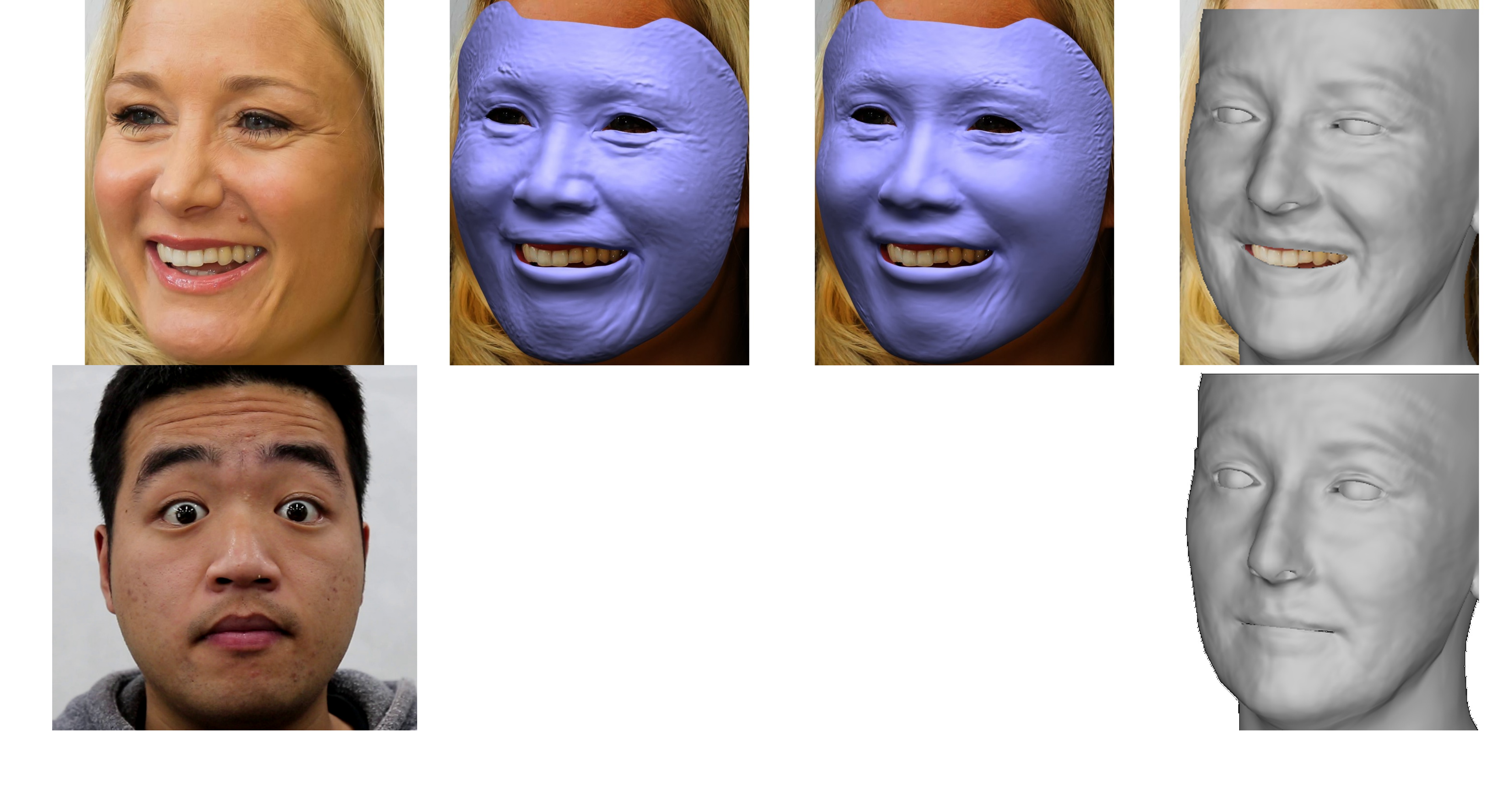}
  \caption{Qualitative results on CelebA-HQ.}
  \label{fig:celeba_results4}
\end{figure*}

Our method can reconstruct and manipulate details from an in-the-wild image.
In Fig. \ref{fig:celeba_results1}, \ref{fig:celeba_results3} and \ref{fig:celeba_results4}, we show more comparison results on the images from CelebA-HQ\cite{Karras2018ProgressiveGO} dataset.
We generate more diverse dynamic details corresponding to the reference expression, and can properly animate the dynamic details in the input image.
As a morphable model-based method, we are also more robust than FaceScape in handling occlusions like facial hair.

\section{Examples of extracted detail line maps and distance fields}

\begin{figure*}[ht]
  \centering
  \includeinkscape[width=\linewidth]{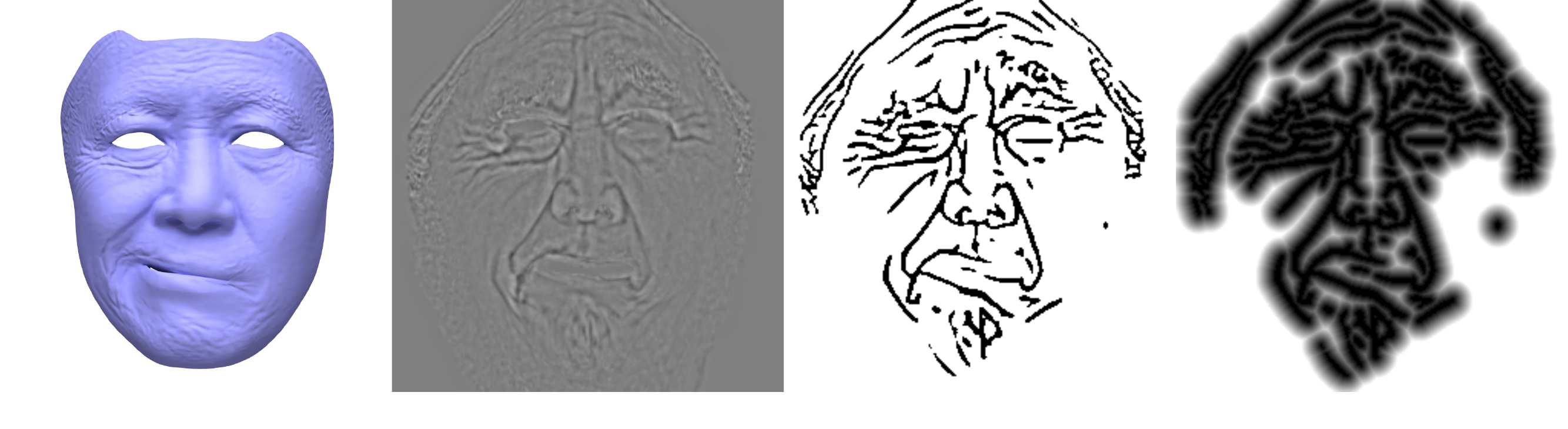}
  \includeinkscape[width=\linewidth]{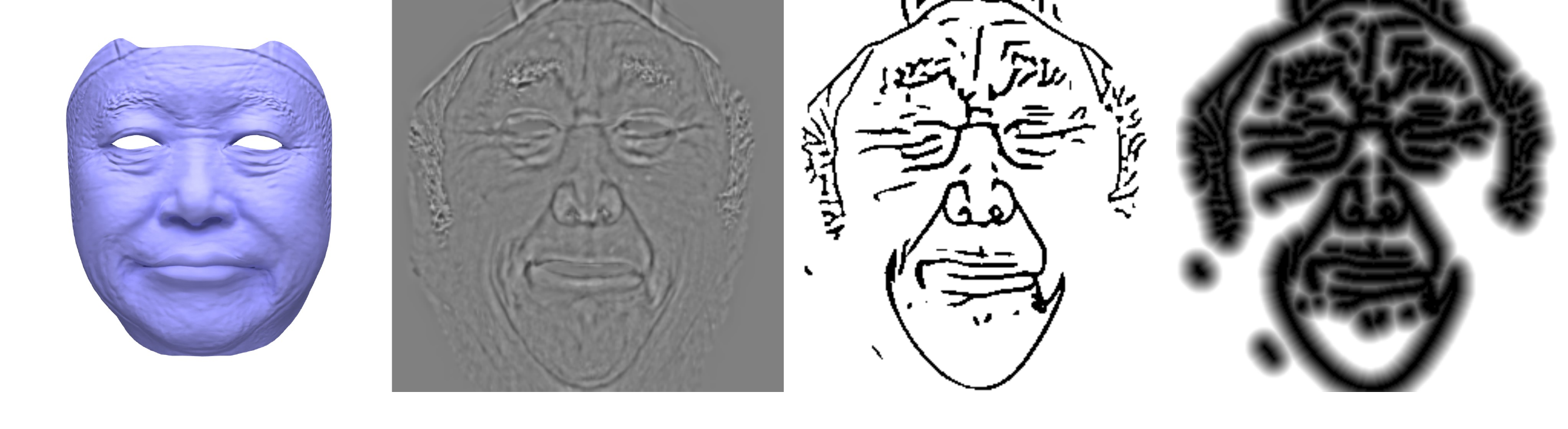}
  \includeinkscape[width=\linewidth]{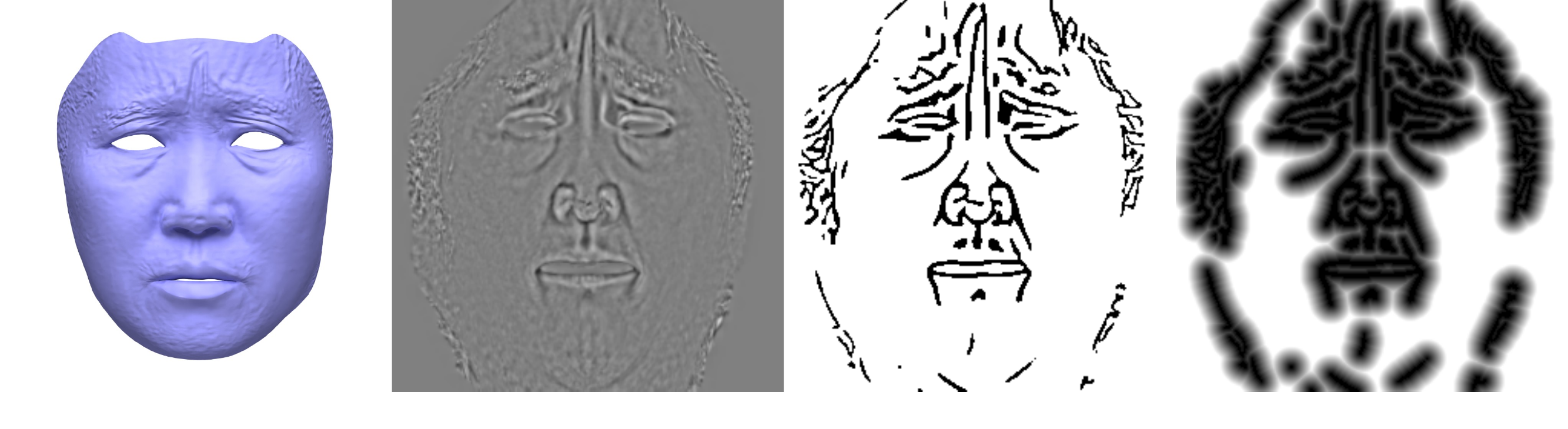}
  \includeinkscape[width=\linewidth]{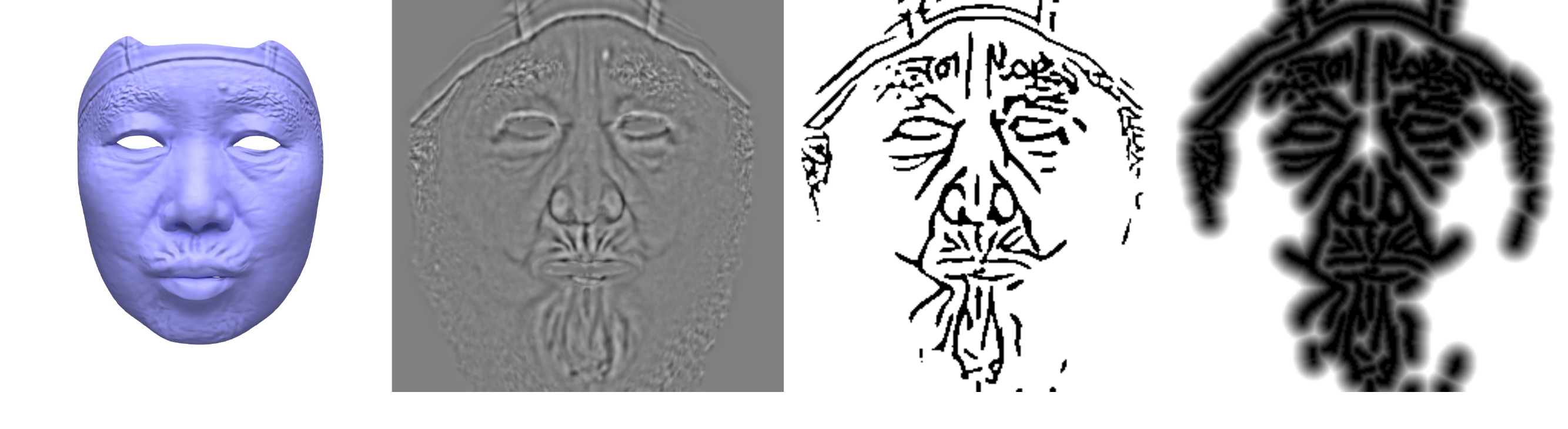}
  \caption{Scans, displacement maps, extracted detail lines and distance fields.}
  \label{fig:data_results1}
\end{figure*}

We use a distance field as the detail structure representation in our model.
We obtain the distance field by extracting lines from the displacement map, and converting the lines to a distance field.
More examples of the input scans, displacement maps, extracted line maps and distance fields are shown in Fig. \ref{fig:data_results1}.

\end{document}